\documentclass[english]{article}
\usepackage[T1]{fontenc}
\usepackage[latin9]{inputenc}
\usepackage{geometry}
\usepackage{color}
\usepackage{babel}
\usepackage{enumitem}
\usepackage{amsmath, amsthm, amssymb, amsfonts}
\usepackage{graphicx}
\usepackage{tikz}
\usetikzlibrary{shapes,backgrounds}

\usepackage{algorithmic}
\usepackage{algorithm} 
\usepackage{booktabs}

\usepackage{hyperref}
\hypersetup{colorlinks}
\usepackage{authblk}

\usepackage{scalerel,stackengine}
\stackMath
\newcommand\widecheck[1]{%
\savestack{\tmpbox}{\stretchto{%
  \scaleto{%
    \scalerel*[\widthof{\ensuremath{#1}}]{\kern-.6pt\bigwedge\kern-.6pt}%
    {\rule[-\textheight/2]{1ex}{\textheight}}
  }{\textheight}%
}{0.5ex}}%
\stackon[1pt]{#1}{\scalebox{-1}{\tmpbox}}%
}

\makeatother

\usepackage{bm}
\usepackage{mathtools}
\usepackage{caption}
\usepackage{subcaption}

\global\long\def\var{\operatorname{Var}}%
\global\long\def\spn{\operatorname*{span}}%
\global\long\def\real{\mathbb{R}}%
\global\long\def\E{\mathbb{E}}%
\global\long\def\indic{\mathds{1}}%
\global\long\def\P{\mathbb{P}}%
\global\long\def\intr{\mathrm{int}\,}%
\global\long\def\ddup{\textup{d}}%
\global\long\def\b{\bm{b}}%
\global\long\def\c{\bm{c}}%
\global\long\def\e{\bm{e}}%
\global\long\def\u{\bm{u}}%
\global\long\def\v{\bm{v}}%
\global\long\def\w{\bm{w}}%
\global\long\def\x{\bm{x}}%
\global\long\def\y{\bm{y}}%
\global\long\def\bbeta{\bm{\beta}}%
\global\long\def\bzero{\bm{0}}%
\global\long\def\cB{\mathcal{B}}%
\global\long\def\cV{\mathcal{V}}%
\global\long\def\bmu{\bm{\mu}}%
\global\long\def\btheta{\bm{\theta}}%
\global\long\def\bTheta{\bm{\Theta}}%
\global\long\def\B{\bm{B}}%
\global\long\def\U{\bm{U}}%
\global\long\def\V{\bm{V}}%
\global\long\def\X{\bm{X}}%
\global\long\def\Id{\bm{I}}%
\global\long\def\cL{\mathcal{L}}%
\global\long\def\vor{\cV}%
\global\long\def\softvor{\widetilde{\cV}}%
\global\long\def\bdr{\partial}%
\global\long\def\softbdr{\widetilde{\partial}}%
\global\long\def\goodset{\mathcal{G}}%
\global\long\def\deltamax{\Delta_{\max}}%
\global\long\def\deltamin{\Delta_{\min}}
\global\long\def\deltacell{D_{\mathrm{cell}}}%
\global\long\def\tdeltacell{\widetilde{\Delta}_{\mathrm{cell}}}%
\global\long\def\std{\sigma}%
\global\long\def\stdstar{\std}%
\global\long\def\bthetastar{\btheta^{*}}%
\global\long\def\bThetastar{\bTheta^{*}}%
\global\long\def\bbetabar{\bar{\bbeta}}%
\global\long\def\ktrue{k_{*}}%
\global\long\def\kfit{k}%
\global\long\def\asso{\psi}%
\DeclareSymbolFont{cmbrightop}{OT1}{cmbr}{m}{n}
\DeclareMathSymbol{\sfPsi}{\mathalpha}{cmbrightop}{9}
\global\long\def\Asso{\sfPsi}%

\newcommand{\NN}{\mathbb{N}}
\newcommand{\bOne}{\boldsymbol{1}}
\newcommand{\GMM}{\texttt{GMM}}
\DeclareSymbolFont{sfletters}{OML}{cmbrm}{m}{it}
\DeclareMathSymbol{\sphi}{\mathord}{sfletters}{"1E}

\newcommand{\setA}{\mathcal{A}}
\newcommand{\setB}{\mathcal{B}}
\newcommand{\setC}{\mathcal{C}}

\newcommand{\cA}{\mathcal{A}}
\newcommand{\cE}{\mathcal{E}}
\newcommand{\cI}{\mathcal{I}}
\newcommand{\cJ}{\mathcal{J}}
\newcommand{\cN}{\mathcal{N}}

\newcommand{\cR}{\mathcal{R}}
\newcommand{\cS}{\mathcal{T}} 
\newcommand{\cT}{\mathcal{S}} 
\newcommand{\cX}{\mathcal{X}}
\newcommand{\bbS}{\mathbb{T}} 
\newcommand{\bbT}{\mathbb{S}} 
\newcommand{\deff}{d_\textnormal{eff}}
\newcommand{\sfx}{\textsf{x}}
\newcommand{\sfz}{\textsf{z}}
\newcommand{\assop}{\asso^{\mathrm{prox}}}

\newcommand{\Assop}{\Asso^{\mathrm{prox}}}
\newcommand{\bAssop}{\overline{\Asso}^{\mathrm{prox}}}

\newcommand{\psiab}{\psi_{\alpha,\beta}}

\newcommand{\setprob}{\cA^{\mathrm{prob}}}
\newcommand{\setconf}{\cB^{\mathrm{conf}}}
\newcommand{\Alg}{\texttt{Alg}}

\newcommand{\hw}{\widehat{w}}

\newcommand{\bv}{\boldsymbol{v}}
\newcommand{\hbv}{\widehat{\bv}}
\newcommand{\hbbeta}{\widehat{\bbeta}}

\newcommand{\bdw}{\varepsilon_{\delta}}

\theoremstyle{plain}
\newtheorem{definition}{Definition}

\newtheorem{theorem}{Theorem}
\newtheorem{corollary}{Corollary}
\newtheorem{lemma}{Lemma}
\newtheorem{proposition}{Proposition}

\theoremstyle{remark}
\newtheorem{remark}{Remark}

\usepackage{babel}
\usepackage{dsfont}

\title{
    Local Minima Structures in Gaussian Mixture Models
    }
\author[1]{Yudong Chen}
\author[2]{Dogyoon Song}
\author[3]{Xumei Xi}
\author[4]{Yuqian Zhang}
\affil[1]{\small Department of Computer Sciences, University of Wisconsin-Madison}
\affil[2]{\small Department of Electrical Engineering and Computer Science, University of Michigan}
\affil[3]{\small School of Operations Research and Information Engineering, Cornell University}
\affil[4]{\small Department of Electrical and Computer Engineering, Rutgers University}

\begin{document}
\maketitle

\begin{abstract}
We investigate the landscape of the negative log-likelihood function of Gaussian Mixture Models (GMMs) with a general number of components in the population limit. As the objective function is non-convex, there can exist multiple spurious local minima that are not globally optimal, even for well-separated mixture models. Our study reveals that all local minima share a common structure that partially identifies the cluster centers (i.e., means of the Gaussian components) of the true location mixture. Specifically, each local minimum can be represented as a non-overlapping combination of two types of sub-configurations: (1) fitting a single mean estimate to multiple Gaussian components or (2) fitting multiple estimates to a single true component. These results apply to settings where the true mixture components satisfy a certain separation condition, and are valid even when the number of components is over- or under-specified. We also present a more fine-grained analysis for the setting of one-dimensional GMMs with three components, which provide sharper approximation error bounds with improved dependence on the separation parameter.
\end{abstract}
\tableofcontents

\section{Introduction\label{sec:intro}}

Mixture models, such as the Gaussian mixture model (GMM), are a class of latent variable models that offer a flexible approach to approximate complex multi-modal distributions. 
These models have been widely used in statistical inference with heterogeneous data. 
A standard method for estimating the parameters of GMM is maximum likelihood estimation, which seeks the global minimum of the negative log-likelihood function of the model. 
The statistical properties of the maximum likelihood estimators, including its asymptotic consistency \cite{vanderVart1998asymptotic} and finite-sample error rates \cite{chen1995mixture,nguyen2013convergence,heinrich2018strong}, have been well studied. 

However, estimating GMMs poses significant computational challenges, and the extent of these challenges is not fully understood. 
The negative log-likelihood function of GMM is non-convex and generally has multiple local minima. Consequently, standard iterative algorithms, such as the Expectation-Maximization (EM) \cite{dempster1977maximum}, are only guaranteed to converge to a local minimum \cite{wu1983convergence,kumar2017rate_EM}. 
A recent study \cite{jin2016local} shows that for GMMs with three or more well-separated components, there provably exist spurious local minima that may be arbitrarily far from the global minimum in both Euclidean distance and in likelihood values; moreover, randomly initialized EM algorithms converge to a spurious local minimum with high probability. 
For certain special cases of GMMs, such as those with two equally weighted components, it has been shown that the negative log-likelihood function in fact has no spurious local minimum, and the EM converges to the global minimum from arbitrary initialization \cite{daskalakis2016ten,xu2016em}. These global convergence results, however, are the exceptions rather than the norm as demonstrated in \cite{jin2016local}. Despite the negative theoretical results, iterative methods such as the EM and its variants are routinely applied to GMMs in practice. 

Motivated by the above challenges and ubiquity of iterative methods, in this paper we perform a more fine-grained investigation of  the likelihood landscape and local minima structures of GMMs. 
Specifically, we seek to answer the following question:
\begin{quote}
    \textit{
    ``Do spurious local minima possess informative structures related to the global minimum?''
    }
\end{quote}

\subsection{Our Contributions}
\label{sec:contributions}

We consider the problem of estimating the component means of a GMM with a general number of equally weighted components. 
Suppose that one fits a mixture of $\kfit$ Gaussians with means $\B=(\bbeta_{1},\ldots,\bbeta_{\kfit})$ to data generated by a true mixture of $\ktrue$ Gaussians with true means $\bThetastar=(\bthetastar_{1},\ldots,\bthetastar_{\ktrue})$. We investigate the structures of the local minima of the negative log-likelihood function in the population limit (i.e., the sample size $n \to \infty$). 

Our main contribution is a proof that \emph{all} local minima $\B$ of the negative likelihood share a similar structure that partially identifies the means $\bThetastar$ of the true mixture model. 
Specifically, each local minimum only involves two types of sub-configurations: either a single estimated center $\bbeta_{i}$ is close to the average of several true component means $\{\bthetastar_{s}\}$, or several estimated centers $\{\bbeta_{i}\}$ are close to a single true center $\bthetastar_{s}$. Moreover, these sub-configurations involve disjoint sets of estimates and component means. 
Notably, this result holds even when the number of components in the fitting model, $\kfit$, is different from the number of components in the true model, $\ktrue$.

\begin{figure}[h]
    \noindent \begin{centering}
        \includegraphics[width=0.5\linewidth]{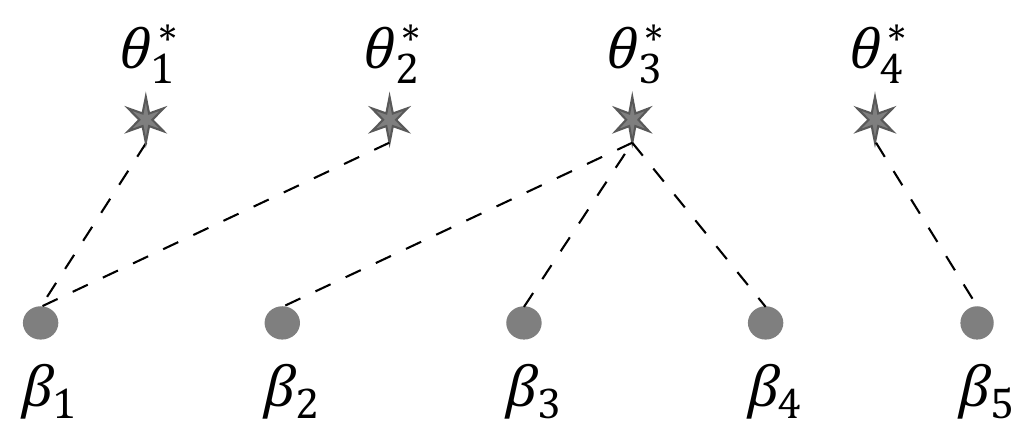}
    \par\end{centering}
\caption{Association between mean estimates $\{\protect\bbeta_{i}\}$ and true component means $\{\protect\bthetastar_{s}\}$ in a local minimum.\label{fig:bipartite}}
\end{figure}

To illustrate the above structure, let us consider an example scenario where $\kfit=5$ and $\ktrue=4$. 
A local minimum $\B$ of the negative log-likelihood has the form
\begin{equation}\label{eq:example}
    \bbeta_{1}\approx\frac{1}{2}\left(\bthetastar_{1}+\bthetastar_{2}\right),
    \bbeta_{2}\approx\bbeta_{3}\approx\bbeta_{4}\approx\bthetastar_{3}
    \text{ and }
    \bbeta_{5}\approx\bthetastar_{4}.
\end{equation}
The relationship between $\B$ and $\bThetastar$ can be summarized by the association graph given in Figure~\ref{fig:bipartite}. 
This association graph can be decomposed into three disjoint subgraphs, each of which is a star graph containing only one mean estimate $\bbeta_i$, or only one true component mean $\bthetastar_s$. We refer to the first type of subgraphs as ``one-fits-many'' (one $\bbeta_i$ fits many $\bthetastar_s$'s), and the second type ``many-fit-one''.
The main theorem in this paper, Theorem \ref{thm:main}, states that \emph{all} local minima of the population negative log-likelihood exhibit a similar combinatorial structure (potentially plus some unassociated estimates). 
In particular, each local minimum $\B$ corresponds to a \emph{disjoint} union of complete bipartite graphs $(\cS_{a},\cT_{a})_{a=1,2,\ldots}$ (potentially plus a degenerate pair $(\cS_0, \cT_0)$ with $\cT_0 =\emptyset$); each bipartite graph is between a subset of estimated centers $\cS_{a} \subseteq [\kfit]$ and a subset of true component means $\cT_{a} \subseteq [\ktrue]$, such that at least one of $\cS_{a}$ and $\cT_{a}$ is a singleton, as shown in Figure~\ref{fig:bipartite}. 
Moreover, Theorem \ref{thm:main} provides upper bounds on the approximation errors in equation~(\ref{eq:example}), characterizing how the errors depend on the separation between the true mixture components. 

The above results apply to GMMs with any values of $\kfit$ and $\ktrue$. Specializing to the exemplar setting of one-dimensional GMM with $\ktrue =3$ components, we show that sharper results can be obtained via a more fine-grained analysis.
In particular, we show that the approximation errors diminish at a quadratic exponential rate with respect to the separation between true component means (Theorem \ref{thm:boost}).

The above findings complement the negative results in~\cite{jin2016local} on the existence of spurious local minima arbitrarily far from the global optimum $\bThetastar$ \emph{in Euclidean distance}. Our results show that all local minima are close to $\bThetastar$ in a different sense, as they partially recover the structure of $\bThetastar$. 
Consequently, standard iterative algorithms such as the EM algorithm and gradient methods, starting from an arbitrary initial solution (except from a set of measure zero \cite{wu1983convergence,kumar2017rate_EM,lee2016gradient}), converge to a solution that is informative of  $\bThetastar$. 

While there has been a long line of research on algorithms and theoretical guarantees for learning GMMs, we emphasize that our focus in this paper is not on obtaining quantitative improvements on these results. Rather, we aim to understand the structural properties of the non-convex likelihood function and iterative methods for optimizing the likelihood. Nevertheless, we expect that our structural results will have important implications for developing better algorithms, especially in the overparameterized regime where $\kfit \gg \ktrue$; we elaborate on this point in Section~\ref{sec:Discussion}. 



\subsection{Related Work}\label{sec:related}

In 2006, Srebro posed the question of whether the population negative log-likelihood function of GMMs has spurious local minima \cite{srebro2007question}. 
In 2016, Jin \textit{et al.} answered this question in the negative for the general case with $\ktrue\ge3$ \cite{jin2016local}.
Motivated by the computational considerations in estimating GMMs, recent works seek to understand the finer properties of the likelihood function as well as those of the EM algorithm---arguably the most popular algorithm for GMMs.

One line of work investigates the \emph{local} behaviors of the likelihood in a neighborhood around the global optimum, which is relevant to the EM starting from a sufficiently good initial solution. 
The work in \cite{balakrishnan2014EM} proposes a general framework for establishing the local geometric convergence of the EM; implicit in their results is that the negative log-likelihood function of a two-component GMM has no other local minima near the global minimum. 
The extension to multiple-component settings is considered in the work \cite{yan2017convergence}. 
Further work in this line studies GMMs with additional structures \cite{yi2015regularized,wang2014HighDimEM,hao2017simultaneous}, the EM with unknown mixture weights and covariances \cite{cai2019chime}, confidence intervals constructed using the EM \cite{chen2022statistical}, and the setting where the number of components is under-specified
\cite{dwivedi2018misspecified}.

Another line of work studies \emph{global} properties of GMMs and the EM in certain restricted settings, mostly that with $\ktrue=2$ (typically equally weighted) components. 
In this setting, the work in \cite{daskalakis2016ten,xu2016em} proves that the EM initialized at a random solution converges to the global minimum, showing that the negative log-likelihood function has no spurious local minimum that is not globally optimal in this case. 
This fact is further investigated in the work \cite{mei2016landscape}, which proposes a general framework for transferring the properties of the population likelihood function to its empirical counterpart. 
Extensions to mixtures of two log-concave distributions \cite{qian2019logconcave} or two linear regressions \cite{kwon2018global} have also been considered. 
A more recent set of papers study the delicate behaviors of the EM when the two components have small or no separation, or when the number of components is mis-specified \cite{dwivedi2020singularity,dwivedi2019challenges,kwon2021minimax,wu2022randomly}; in these settings, the EM may exhibit a slower, non-parametric statistical error rate. 
The above global results for $\ktrue=2$ usually do not generalize to $\ktrue\ge3$, as spurious local minima provably exist in the latter \cite{jin2016local}. 
Moreover, additional spurious local minima may arise when the mixture weights are not equal \cite{xu2018benefits}.

Our recent work~\cite{qian2021structures} studies the related problem of optimizing the (non-smooth) k-means objective function, which can be viewed as a limit of the log-likelihood function of GMMs when the posited variance of Gaussian components goes to zero. 
The results in~\cite{qian2021structures} and this paper are similar in spirit: spurious local minima provably exist but possess additional hidden structures. 
However, the theoretical results in this paper are substantially sharper and apply to the over/under-parameterized regime $\kfit \neq \ktrue$. 
The proofs in this paper rely on quite different techniques and are also considerably simpler than~\cite{qian2021structures} in several aspects, taking advantage of the rich structures in the smooth log-likelihood objective function for GMMs.

We remark that several recent works~\cite{li2023multi,dicks2022elucidating} have developed new algorithms for k-means and GMMs that explicitly leverage the one-fits-many/many-fit-one structures studied in~\cite{qian2021structures} and this paper. 


\subsection{Organization}
This paper is organized as follows. 
In Section \ref{sec:setup}, we provide formal definitions and notation that are used throughout the paper. 
In Section \ref{sec:prelim}, we present preliminary analytical results for the likelihood landscape of GMMs. 
Section \ref{sec:main} contains the main technical results of this work. 
Specifically, in Section \ref{sec:main_theorem}, we state the structures of local minima, and in Section \ref{sec:implications}, we discuss their implications. 
Additionally, in Section \ref{sec:boost}, we explore how we can improve the approximation error bounds in the main theorem in the example setting of a one-dimensional GMM. 
Section \ref{sec:Discussion} provides discussion on the our results and future directions. 

The remaining sections are dedicated to technical details and proofs. In particular, 
Sections \ref{sec:proof_stationary}, \ref{sec:proof_main_theorem} and \ref{sec:proof_main_boost} contain the proofs for the theoretical results in Sections~\ref{sec:prelim}, \ref{sec:main_theorem} and~\ref{sec:boost}, respectively. 
We defer subsidiary items, such as proofs of technical lemmas, to the Appendix.

\section{Problem setup}\label{sec:setup}

In this section, after introducing the basic notation, we describe the problem setup of Gaussian mixture models and the associated maximum likelihood estimation approach.


\subsection{Notation}
We let $\NN_+$ denote the set of positive integers, and $\NN := \NN_+ \cup \{0\}$. 
For each $m \in \NN_+$, let $[m]:=\{1,2,\ldots,m\}$ and $[m]_0 := [m] \cup \{0\}$. 
Given a set $\cI \subseteq \NN_+$, let $|\cI|$ denote the cardinality of $\cI$ and $\cI(i)$ denote the $i$-th smallest integer in $\cI$, so $\cI(1) < \dots < \cI(|\cI|)$. 
Let $\real$ denote the real numbers and $\real_+ := \{ x \in \real: x > 0 \}$. 
We generally use curly letters (e.g., $\cS$) to denote sets. For a set $\cS \subset \real^d$, let $\intr \cS$ denote its interior. 

We use boldface lowercase letters, e.g., $\x$, to denote deterministic (column) vectors, of which $x_{i}$ is the $i$-th element. 
In particular, $\e_{i}=(0,\ldots,0,1,0,\ldots,0)^{\top}$ is the $i$-th standard basis vector in $\real^{d}$, 
and $\bOne_n=(1,1,\dots,1)\in\real^d$  is the vector of all ones. 
We let $\left\Vert \x \right\Vert $ denote the Euclidean $\ell_2$ norm of $\x$.

We use boldface capital letters, e.g., $\X$, to denote matrices.  In particular,  $\Id_{d}$ is the $d$-by-$d$ identity matrix. We identify a matrix with an ordered set of column vectors, and  let $\X_j \in \real^d$ denote the $j$-th column vector of $\X$. 
For any $\X \in \real^{d \times m}$ and any $\cI \subseteq [m]$, we let $\X(\cI) \in \real^{d \times |\cI|}$ denote a matrix such that 
$\X(\cI)_i = \X_{\cI(i)}$. 
We write $\X \succeq 0$ when $\X$ is symmetric positive semidefinite.

We use sans-serif letters, e.g., $\sf x$ and $\sf X$, to denote random variables and vectors. 
For a probability measure $\nu$, we write $\sf x \sim \nu$ to denote that $\sf x$ is distributed per $\nu$. 
For any $\u \in \real^d$ and $\mathbf{\Sigma} \in \real^{d \times d}$ such that $\mathbf{\Sigma} \succeq 0$, we denote by $\mathcal{N}(\u, \mathbf{\Sigma})$ the Gaussian distribution with mean $\u$ and covariance $\mathbf{\Sigma}$; moreover, we let $\phi(\cdot \mid \u, \mathbf{\Sigma})$ denote the probability density function of $\mathcal{N}(\u, \mathbf{\Sigma})$. 
With an underlying probability measure specified, we use $\E$ and $\P$ to denote the expectation and probability. 
Given a sample space $\Omega$ and an event $A \subseteq \Omega$, the indicator function $\indic_A: \Omega \to \{0,1\}$ is defined such that $\indic_A(x) = 1$ if and only if $x \in A$. 
We sometimes use $\indic(x \in A)$ to denote $\indic_A(x)$. 


\subsection{Gaussian Mixture Models}\label{sec:gmm}

Consider a mixture of $\ktrue$ equally weighted Gaussian distributions, denoted by $\nu^*$, which has the density
\begin{equation}\label{eq:true_mixture}
    f^{*}(\cdot)=\frac{1}{\ktrue}\sum_{s\in[\ktrue]}f_{s}^{*}(\cdot),
\end{equation}
where $f_{s}^{*}(\cdot)$ denotes the density of the $s$-th mixture component, which is a Gaussian distribution. 
Specifically, we assume that $f_{s}^{*}(\cdot):=\phi(\,\cdot\mid\bthetastar_{s},\stdstar^{2} \Id_d)$ for some $\bthetastar_{s} \in \real^d$ and $\sigma \in \real_+$. 
Let $\X = \{ \x_1, \dots, \x_n \} \subset \real^d$ be a collection of $n$ i.i.d. samples generated from $\nu^*$.

The goal of Gaussian mixture modeling is to fit a $\kfit$-component mixture of Gaussians, denoted by $\nu$, to the data $\X$.
Assuming $\sigma$ is known a priori, we may write the density of $\nu$ as 
\begin{equation}
    f(\cdot)=\frac{1}{\kfit}\sum_{i\in[\kfit]}f_{i}(\cdot),\label{eq:fitted_mixture}
\end{equation}
where $f_{i}(\cdot):=\phi(\,\cdot\mid\bbeta_{i},\std^{2} \Id_d)$ for some $\bbeta_i \in \real^d, i \in [\kfit]$. 
We refer to $\{\bbeta_{i},~ i\in[\kfit]\}$ as the \emph{fitted centers}, or \emph{mean estimates}. 
Note that the number of fitted components $\kfit$ may differ from the number of true components $\ktrue$. This covers the exact-parametrization ($\kfit=\ktrue$), over-parametrization
($\kfit>\ktrue$) and under-parametrization ($\kfit<\ktrue$) settings.

To avoid cluttered notation, we have suppressed the dependence of $f^{*}$ and $\{f_{s}^{*}\}$ on $\{\bthetastar_{s}\}$, and likewise for the dependence of $f$ and $\{f_{i}\}$ on $\{\bbeta_{i}\}$. 
We use $\E_{*}$ and $\P_{*}$ to denote the expectation and probability, respectively, under the true mixture model $\nu^{*}$. 
Similarly, for each $s\in[\ktrue]$, we use $\E_{s}$ and $\P_{s}$ to denote the expectation and probability, respectively, under the $s$-th true Gaussian component with density $f_{s}^{*}$. 
Note that $\E_{*}=\frac{1}{\ktrue}\sum_{s\in[\ktrue]}\E_{s}$ and $\P_{*}=\frac{1}{\ktrue}\sum_{s\in[\ktrue]}\P_{s}$ by definition  in \eqref{eq:true_mixture}. 
For clarity, we generally use $s,s'$ to index true mixture components (e.g., $\bthetastar_{s},f_{s}^{*}$), whereas $i,j$ are used to index the components in the fitted model $\nu$ (e.g., $\bbeta_{i},f_{i}$).

For the true mixture model $\nu^*$, we define the maximum and minimum component separations as 
\begin{equation} \label{eq:separation}
\begin{aligned}
    \deltamax &\coloneqq \max_{s,s'\in[\ktrue]}\left\Vert \bthetastar_{s}-\bthetastar_{s'}\right\Vert,\\
    \deltamin &\coloneqq \min_{\substack{s,s'\in[\ktrue]\\s\neq s} }\left\Vert \bthetastar_{s}-\bthetastar_{s'}\right\Vert.
\end{aligned}
\end{equation}
We refer to $\deltamin/\sigma$ as the \emph{Signal-to-Noise Ratio} (SNR). 
Lastly, we define the condition number of $\nu^*$ as
\begin{equation}\label{eqn:cond_num}
    \rho \coloneqq \frac{\deltamax}{\deltamin}.
\end{equation}

\subsection{Maximum Likelihood Estimation}\label{sec:likelihood}

Given the data $\X$ from the true model $\nu^{*}$, the maximum likelihood principle is a standard approach for fitting a model $\nu$ to $\X$.
Let $\bThetastar=(\bthetastar_{1},\ldots,\bthetastar_{\ktrue})\in\real^{d\times\ktrue}$ and $\B=(\bbeta_{1},\ldots,\bbeta_{\kfit})\in\real^{d\times\kfit}$ denote the component mean parameters of the true and fitted models, respectively. 
The population negative log-likelihood function~$L$---the infinite sample limit (i.e., $n \to \infty$) of the sample negative log-likelihood---is given by 
\begin{align}
    L(\B)
        &\equiv L(\B\mid\bThetastar) \nonumber\\
        & =-\E_{*}\left[\log f(\sf x)\right]   \nonumber\\
        & =D_{\mathrm{KL}}\left(f^{*}\Vert f\right)-\E_{*}\left[\log f^{*}(\sf x)\right],\label{eq:likelihood_kl}
\end{align}
where $\sf x$ is a random variable distributed per $\nu^*$, $D_{\mathrm{KL}}\left(f^{*}\Vert f\right):=\E_{*}\left[\log\frac{f^{*}(\sf X)}{f(\sf X)}\right]$ is the Kullback-Leibler (KL) divergence between $\nu^*$ and $\nu$, and the entropy term $-\E_{*}\left[\log f^{*}(\sf x)\right]$ does not depend on $\B$. The maximum likelihood approach involves 
finding $\widehat{\B}$ that minimizes the negative log-likelihood, i.e.,  
\begin{equation}
\label{eqn:miniNLL}
    \widehat{\B} \in \arg\min_{\B \in\real^{d\times\kfit}}L(\B).
\end{equation}

When $\kfit=\ktrue$, it is clear from \eqref{eq:likelihood_kl} and the non-negativity of KL divergence that $\bThetastar$ is a global minimizer of $L$. 
However, $L$ is generally not convex and has local minima other than $\bThetastar$ \cite{jin2016local}. 
This fact poses a significant computational challenge for solving \eqref{eqn:miniNLL} because standard algorithms (such as the EM and gradient methods) are only guaranteed to find a local minimum, which can be substantially suboptimal.

\begin{remark}[Euclidean invariance of $L$]\label{rem:coordinate}
    It is easy to verify that the function $L$ is invariant under a Euclidean transformation, i.e., rotations, translations, reflections, and a sequence thereof. 
    That is, for any orthonormal matrix $\U\in\real^{d\times d}$ and vector $\v\in\real^{d}$, the following holds: 
    \begin{align*}
     & L\left( \U\B+\v\otimes\bOne_{\ktrue} \mid \U\bThetastar+\v\otimes\bOne_{\ktrue} \right)
        =  L(\B \mid \bThetastar)
    \end{align*}
    where $\v\otimes\bOne_{\ktrue} = \begin{bmatrix} \v & \cdots & \v \end{bmatrix} \in \real^{d \times \ktrue}$. 
    In the analysis we frequently make use of this invariance property to choose a convenient coordinate system. 
\end{remark}

\section{Preliminary Analysis}\label{sec:prelim}

In this section, we present a preliminary analysis for the landscape of the likelihood function of Gaussian mixture models. 
In Section~\ref{sec:coeff_assoc}, we characterize the stationary points of the population negative log-likelihood function $L$. 
In Section \ref{sec:stationary}, we derive an equivalent form of the stationary condition (Theorem \ref{thm:equiv}), which immediately implies several useful properties of the stationary points of $L$. 
These results will be used in our analysis in subsequent sections.

\subsection{Coefficients of Association and Optimality Conditions}\label{sec:coeff_assoc}

Here we define the coefficients of association, which are then used to characterize the first-order and the second-order optimality conditions for $L$ as well as the fixed points of the EM and gradient methods. 
Most of the materials in this subsection are not new, at least in the exact parameterization setting ($\kfit = \ktrue$). 
Nevertheless, we collect these results here to set the stage for subsequent analysis.

\subsubsection{Coefficients of Association}

The coefficient of association, defined below, measures the relative strength of association between a data point $\x$ and each of the $k$ estimated components $\{f_i: i \in [\kfit]\}$ in the fitted mixture model.

\begin{definition}\label{defn:coeff_asso}
    Let $\x \in \real^d$, $\B = (\bbeta_1, \dots, \bbeta_k) \in \real^{d \times k}$ and $\sigma \in \real_+$ be given.
    For each $i \in [\kfit]$, the \emph{coefficient of association} between $\x$ and $\bbeta_i$ at level $\sigma$ 
    is defined as
    \begin{equation}
    \label{eq:asso}
    \begin{aligned}
        \asso_{i}(\x) 
            &\equiv \asso_{i}(\x; \B, \sigma)\\
            &\coloneqq \frac{\frac{1}{\kfit}f_{i}(\x)}{f(\x)}=\frac{\exp\left(-\frac{\left\Vert \x-\bbeta_{i}\right\Vert ^{2}}{2\std^{2}}\right)}{\sum_{j\in[\kfit]}\exp\left(-\frac{\left\Vert \x-\bbeta_{j}\right\Vert ^{2}}{2\std^{2}}\right)}.
    \end{aligned}
    \end{equation}
\end{definition}

The association coefficient $\asso_{i}(\x)$ takes the form of a soft (arg)min function and can be viewed as an approximation of the indicator function of $\bbeta_{i}$ being the closest to $\x$ among all the fitted centers $\{\bbeta_j: j \in [\kfit]\}$; that is, 
\[
    \asso_i(\x) \approx \indic\Big\{ i=\arg\min_{j\in[\kfit]}\left\Vert \x-\bbeta_{j}\right\Vert ^{2}\Big\}.
\]
The analysis in this paper makes use of this intuitive interpretation. 
From a Bayesian perspective, one may also interpret $\asso_{i}(\x)$ as the posterior probability of a data point $\x$ belonging to the $i$-th fitted component, given the current estimate $\B$ of the component means. 
As such, the quantity $\asso_{i}(\x)$ appears in the E-step of the EM algorithm, as we shall see momentarily. 
With $\sf x$ denoting a random data point generated from the true distribution $\nu^{*}$, the induced scalar random variables 
\begin{equation}\label{eq:asso_rv}
    \Asso_{i}:=\asso_{i}({\sf x}), \quad i \in [\kfit],
\end{equation}
correspond to the association coefficients between the random vector $\sf x$ and $\B = \left( \bbeta_i: i \in [\kfit] \right)$.

\subsubsection{First-order and Second-order Characterizations of Optimality for \texorpdfstring{$L$}{L}}\label{sec:characterization}

Here we discuss the optimality conditions that characterize the local minimizers of the population negative log-likelihood function $L$ defined \eqref{eq:likelihood_kl}.

\paragraph{First-order necessary condition}

The gradient of $L$ can be expressed in terms of the association coefficients, as stated in the following lemma.
\begin{lemma}\label{lem:gradient_NLL}
    Let $\B = (\bbeta_1, \dots, \bbeta_{\kfit}) \in \real^{d \times \kfit}$ and $\sigma > 0$. 
    The partial derivatives of $L$ at $\B$ admit the expression
    \begin{equation}\label{eq:gradient}
        \frac{\partial}{\partial\bbeta_{i}}L(\B)
            = \frac{1}{\sigma^2} \cdot \E_{*}\left[\Asso_{i} \cdot (\bbeta_{i}-\sf x)\right],\quad i\in[\kfit].
\end{equation}
\end{lemma}

It follows from Lemma \ref{lem:gradient_NLL} that $\B$ is a stationary point of $L$ (i.e., $\nabla L(\B) = 0$) if and only if 
\begin{equation}\label{eqn:stationary.1}
    \E_{*}\left[\Asso_{i}(\bbeta_{i}-\sf x)\right]= 0,\quad\forall i\in[\kfit],
\end{equation}
which is equivalent to 
\begin{equation}\label{eq:station_cond}
    \bbeta_{i}=\frac{\E_{*}\left[\Asso_{i}\sf x\right]}{\E_{*}\left[\Asso_{i}\right]},\qquad\forall i\in[\kfit].
\end{equation}
Equations~\eqref{eqn:stationary.1} and~\eqref{eq:station_cond} are the first-order necessary condition for $\B$ being a local minimizer of for $L$.

\paragraph{Second-order necessary condition} 

We can also write the Hessian of $L$ using the coefficients of association, as done in the next lemma.
\begin{lemma}\label{lem:hessian_NLL}
    Let $\B = (\bbeta_1, \dots, \bbeta_\kfit) \in \real^{d \times \kfit}$ and $\sigma > 0$. 
    The second-order partial derivatives of $L$ at $\B$ admit the expression
    \begin{equation}\label{eq:hessian}
    \begin{aligned}
        \frac{\partial^2}{\partial\bbeta_{i} \partial\bbeta_j}L(\B)
            &= \frac{1}{\sigma^4} \cdot \E_{*}\left[\Asso_{i} \Asso_{j} \cdot (\bbeta_{j}-\sf x) (\bbeta_{i}-\sf x)^{\top} \right]\\
            &\quad + \delta_{ij} \cdot \frac{1}{\sigma^4} \cdot \E_{*}\Big[\Asso_{i} \cdot \left\{ \sigma^2 \cdot \Id_{d} - (\bbeta_{i}-\sf x) (\bbeta_{i}-\sf x)^{\top} \right\} \Big]
    \end{aligned}
    \end{equation}
    for all $i, j \in [\kfit]$, where $\delta_{ij}$ is the Kronecker delta (i.e., $\delta_{ij} = 1$ if $i = j$, and $\delta_{ij} = 0$ otherwise).
\end{lemma}

The Hessian $\nabla^2 L(\B)$ can be expressed as a $\kfit d \times \kfit d$ matrix whose $(i,j)$-th block (of size $d$-by-$d$) is $\frac{\partial^2}{\partial\bbeta_{i} \partial\bbeta_j}L(\B)$ for $i, j \in [\kfit]$. Any point $\B$ that satisfies 
\begin{equation}
    \label{eq:2nd_order_statoinary}
    \nabla L(\B)=0 \quad\text{and}\quad \nabla^2 L(\B) \succeq 0
\end{equation} 
is called a second-order stationary point. Equation~\eqref{eq:2nd_order_statoinary} is the second-order necessary condition for $\B$  being a local minimizer of $L$.

In summary, we have the following inclusion relation:
\begin{align*}
    \{ \B \in \real^{d \times \kfit}: \B \text{ is a local minimizer of }L \}
    &\subseteq
    \{ \B \in \real^{d \times \kfit}: \nabla L(\B)=0 \text{ and } \nabla^2 L(\B) \succeq 0 \}\\
    &\subseteq
    \{ \B \in \real^{d \times \kfit}: \nabla L(\B)=0 \}.
\end{align*}

\subsubsection{Connection between the Stationarity Condition and the EM Algorithm}

The EM algorithm is a popular iterative method for optimizing the likelihood function. In the population setting, the EM update takes the form
\begin{equation}
    \bbeta_{i}\;\leftarrow\;\frac{\E_{*}\left[\Asso_{i}\sf x\right]}{\E_{*}\left[\Asso_{i}\right]}=\bbeta_{i}-\frac{1}{\E_{*}\left[\Asso_{i}\right]}\cdot\frac{\partial}{\partial\bbeta_{i}}L(\B),\quad \forall i\in[\kfit],\label{eq:EM}
\end{equation}
where we have combined the E-step (computing $\Asso_{i}$) and the M-step (computing $\bbeta_{i}$) into one update. The EM update~\eqref{eq:EM} can be viewed as a fixed point iteration for solving the stationary condition~(\ref{eq:station_cond}), or as 
a gradient descent-like (or quasi-Newton) algorithm with a coordinate-dependent step size $1/\E_{*}\left[\Asso_{i}\right]$ \cite{xu1996convergence}.
Therefore, the fixed points of the EM algorithm correspond to the stationary points of $L$, and the \emph{stable fixed points} of the EM correspond to the local minimizers.

Our results in Section \ref{sec:stationary} provide further characterizations of the stationary points and local minimizers of $L$. 
In light of the above discussion, these results immediately apply to the solution returned by the EM algorithm and other local algorithms including gradient descent and Newton methods.

\subsection{Properties of the Stationary Points of \texorpdfstring{$L$}{L}}\label{sec:stationary}

Our first theorem provides a necessary and sufficient condition for the (first-order) stationary points of $L$.  Recall that $\bTheta^* = \left( \btheta^*_1, \btheta^*_2, \dots, \btheta^*_{\ktrue} \right)$ is the true parameters for the Gaussian mixture model whose density is given in \eqref{eq:true_mixture}. 

\begin{theorem}[Equivalent stationary condition]\label{thm:equiv}  
    A point $\B= (\bbeta_1, \dots, \bbeta_\kfit)\in\real^{d\times\kfit}$ is a stationary point of $L$ if and only if 
    \begin{align}
        \sum_{j\in[\kfit]}\bbeta_{j}\sum_{s\in[\ktrue]}\E_{s}\left[\Asso_{i}\Asso_{j}\right] & =\sum_{s\in[\ktrue]}\bthetastar_{s}\E_{s}\left[\Asso_{i}\right],\quad\forall i\in[\kfit].\label{eq:equiv}
    \end{align}
\end{theorem}
\noindent We prove Theorem \ref{thm:equiv} in Section~\ref{sec:proof_equiv} using the Stein's identity.

Equation \eqref{eq:equiv} is equivalent to the original stationary condition \eqref{eq:station_cond}.
However, the expression in~\eqref{eq:equiv} is often more useful as it exposes the relationship between the fitted centers $\{\bbeta_{j}\}$ and the true centers $\{\bthetastar_{s}\}$. 
This result plays a key role in establishing the main results of this paper in Section \ref{sec:main}. 

Equation \eqref{eq:equiv} links the estimates $\B$ to the true component centers $\bThetastar$ by establishing a system of equations: for each $i \in [\kfit]$, a convex combination of the estimates $\bbeta_j$ weighted by $w_j \coloneqq \sum_{s\in[\ktrue]}\E_{s}\left[\Asso_{i}\Asso_{j}\right]$ is equated to a convex combination of true component centers $\bthetastar_s$ weighted by $w'_s \coloneqq \E_{s}\left[\Asso_{i}\right]$. 
This system of equations implicitly characterizes possible configurations of (first-order) stationary points. 
The four corollaries in Section \ref{sec:geometric_corollaries} derive various useful properties of the stationary points and local minimizers of $L$ from equation \eqref{eq:equiv}. Although these corollaries might seem intuitive, proving some of them could be challenging using other methods, particularly when $\kfit \neq \ktrue$.

\subsubsection{General Geometric Properties of Stationary Points}\label{sec:geometric_corollaries}

The result in the first corollary is probably well known. It states that for any stationary point of $L$, the weighted average of the fitted centers must equal the mean of the true centers.

\begin{corollary}[Mean consistency]\label{cor:mean}
    If $\B\in\real^{d\times\kfit}$ is a stationary point of $L$, then
    \[
        \sum_{j\in[\kfit]} \E_{*}\left[\Asso_{j}\right] \cdot \bbeta_{j} 
            = \frac{1}{\ktrue}\sum_{s\in[\ktrue]}\bthetastar_{s}.
    \]
\end{corollary}

\begin{proof}[Proof of Corollary \ref{cor:mean}]
    Adding up the equation \eqref{eq:equiv} over $i\in[\kfit]$ and using the fact that $\sum_{i\in[\kfit]}\Asso_{i}=1$ surely, 
    we obtain that $\sum_{j\in[\kfit]}\bbeta_{j}\sum_{s\in[\ktrue]}\E_{s}\left[\Asso_{j}\right]=\sum_{s\in[\ktrue]}\bthetastar_{s}.$
    Since $\sum_{s\in[\ktrue]}\E_{s}\left[\Asso_{j}\right]=\ktrue\E_{*}\left[\Asso_{j}\right]$, the proof is complete. 
\end{proof}

The next corollary states that any stationary point of $L$ must lie in the linear subspace spanned by the true component centers. 

\begin{corollary}[Linear span]\label{cor:span}
    If $\B\in\real^{d\times\kfit}$ is a stationary point of $L$, then we have
    \[
    \bbeta_{i}\in\spn\left\{ \bthetastar_{s},s\in[\ktrue]\right\} ,\quad i\in[\kfit].
    \]
\end{corollary}
\noindent 
With this property, we can restrict ourselves in subsequent analysis to $\spn\left\{ \bthetastar_{s}\right\}$, a $\ktrue$-dimensional subspace of $\real^d$. 
This property is particularly useful when $\ktrue$ is much smaller than $d$. 
The proof of Corollary \ref{cor:span} is deferred to Section~\ref{sec:proof_span}.

\subsubsection{Two Extreme Cases of Fitting Gaussian Mixtures}\label{sec:two_extreme}

We consider two extreme cases of the problem of fitting a mixture model of $\kfit$ Gaussians to data generated by a mixture of $\ktrue$ Gaussians: the case with $\ktrue = 1$, and the case with $\kfit = 1$. 
Theorem \ref{thm:equiv} readily identifies the stationary points in each of these cases.
Specifically, the population log-likelihood $L$ has a unique stationary point that is the global minimizer, when we fit multiple Gaussians to a single one (Corollary \ref{cor:1Gaussian}), or a single Gaussian to multiple ones (Corollary \ref{cor:1_fit_many}). 

As we  show shortly in Section \ref{sec:main}, these two settings are  the atomic cases of the general setting with arbitrary $\kfit$ and $\ktrue$. 
In particular, any local minimizer of $L$ can be decomposed into a non-overlapping collection of the global minimizers of sub-problems in these two settings (plus a collection of vectors that are almost irrelevant, if any); see Theorem~\ref{thm:main} and Section \ref{sec:implications}.

\paragraph{Case 1 ($\ktrue=1$)}

If the true model has only one component with mean $ \bthetastar_1$, then $L$ has a unique stationary point corresponding to the true center, regardless of the number $\kfit$ of fitted centers. There are no other local or global minimizers.

\begin{corollary}[Fitting $\kfit$ Gaussians to one Gaussian]\label{cor:1Gaussian}
    If $\ktrue=1$, then $L$ has a unique stationary point $\B\in\real^{d\times\kfit}$ such that 
    \[
        \bbeta_{i}=\bthetastar_{1},\quad\forall i\in[\kfit].
    \]
\end{corollary}

\begin{proof}[Proof of Corollary \ref{cor:1Gaussian}]
Without loss of generality we may assume that $\bthetastar_{1}=\bzero$
(see Remark~\ref{rem:coordinate}). If $\B$ is a stationary point of $L$, then Corollary~\ref{cor:span} implies that $\bbeta_{i}\in\spn\left\{ \bthetastar_{1}\right\} =\{\bzero\},\forall i\in[\kfit]$.
Conversely, if $\bbeta_{i}=\bzero$ for all $i$, then $\asso_{i}(\x)=\frac{1}{\kfit}$ for all $\x\in\real^{d}$ and the stationary condition \eqref{eq:station_cond} is satisfied by $\B$.
\end{proof}

Corollary~\ref{cor:1Gaussian} is related to a recent line of work
in \cite{dwivedi2020singularity,dwivedi2019challenges,wu2022randomly}
on the setting where the number of components in the mixture is over-specified (i.e., $\kfit > \ktrue$).
In the canonical over-specified setting where one fits
a mixture of $\kfit=2$ Gaussians to data from a single Gaussian,
the work above showed that the EM algorithm converges towards the true center from random initialization
(albeit with a slower convergence rate and a larger statistical error
$(d/n)^{1/4}$ than in the exact-specified setting). At the population
level, Corollary~\ref{cor:1Gaussian} provides a more general result,
applicable to any number $\kfit\ge 1$ of specified components and any descent algorithms beyond the EM.

\paragraph{Case 2 ($\kfit=1$)}

As a sanity check, we consider an under-specified setting with $\kfit=1$ and $\ktrue\ge1$, that is, fitting a single Gaussian to a mixture of multiple Gaussians. 
In this case, we have $\asso_{1}(\x)=1,~\forall\x\in\real^{d}$, hence equation \eqref{eq:equiv} immediately implies the following result (its proof is trivial):

\begin{corollary}[Fitting One Gaussian to $\ktrue$ Gaussians]\label{cor:1_fit_many}
    If $\kfit=1$, then $L$ has a unique stationary point $\B=(\bbeta_{1})\in\real^{d\times1}$ satisfying 
    \[
        \bbeta_{1}=\frac{1}{\ktrue}\sum_{s\in[\ktrue]}\bthetastar_{s}=\E_{*}[\sf x].
    \]
\end{corollary}
\noindent We thus recover the elementary fact that the Maximum Likelihood Estimator of fitting a single Gaussian to a dataset is given by the sample mean.

\section{Main Results: Combinatorial Structures of Local Minima}\label{sec:main}

In this section, we present the main results of this work. Section~\ref{sec:main_theorem} gives the main theorem (Theorem~\ref{thm:main}), which characterizes the common structures shared by all local minima of $L$ with an arbitrary number of mixture components, $\ktrue, \kfit \geq 1$. 
In Section~\ref{sec:implications}, we discuss the implications of this result by connecting it to the decomposability of mixture learning problem instances. 
In Section~\ref{sec:boost}, we show that the approximation error bounds in Theorem \ref{thm:main} can be exponentially improved in the example setting of a one-dimensional three-component mixture of Gaussian distributions.

\subsection{Main Theorem Statement}\label{sec:main_theorem}

To state the main theorem, we define the Voronoi cells induced by a solution $\B$ and the associated index function.

\begin{definition}[Voronoi cells]\label{defn:voronoi}
    Let $\B = (\bbeta_i)_{i=1}^{\kfit} \in \real^{d \times \kfit}$ and $i\in[\kfit]$. 
    The $i$-th \emph{Voronoi cell} of $\B$ is defined as 
    \begin{equation}    \label{eq:vor_def}
        \begin{aligned}
            \vor_{i} 
                = \vor_{i}(\B) 
                & \coloneqq\left\{ \x\in\real^{d}:\left\Vert \x-\bbeta_{i}\right\Vert \le\left\Vert \x-\bbeta_{j}\right\Vert ,\forall j\in[\kfit]\right\} \\
                & =\left\{ \x \in \real^d:\asso_{i}(\x)\ge\asso_{j}(\x),\forall j\in[\kfit]\right\}.
        \end{aligned}
    \end{equation}
    The \emph{index function} induced by $\B$ is a function $\iota_{\B}: \real^d \to [\kfit]$ such that
    \[
        \iota_{\B}(\x) \coloneqq \min \left\{ i \in [\kfit]: \x \in \vor_i(\B) \right\}.
    \]
    The maximum width of the Voronoi cells relative to $\B$ is defined as 
    \[
    \deltacell(\B) \coloneqq \max_{j \in [\kfit]} \max_{s \in [\ktrue]: \bthetastar_s \in \vor_j} \| \bthetastar_s - \bbeta_j \|.
    \]
\end{definition}

Note that when $\{\bbeta_i\}$ are distinct, the Voronoi cells partition the ambient subspace $\real^d$ (up to a set of measure zero). Also let $\deff\coloneqq \dim \spn \left\{ \bthetastar_s: ~ s \in [\ktrue] \right\}$ denote the effective dimension of the true mixture $\nu^*$ (cf.\ Corollary~\ref{cor:span}); note that $\deff \le \min\{d,\ktrue\}$. Finally, recall the signal-to-noise ratio $\deltamin/\sigma$ and the condition number $\rho = \deltamax/\deltamin$ defined in equations~\eqref{eq:separation} and~\eqref{eqn:cond_num}. 

We are ready to present the main theorem of this paper.

\begin{theorem}[Main theorem]
\label{thm:main}
    Let $\B \in \real^{d \times k}$ be a local minimizer of $L(\,\cdot \,|\, \bThetastar)$. 
    If
    \begin{equation}\label{eqn:SNR_requirement}
        \frac{\deltamin}{\sigma} > 72 (\sqrt{2\pi} + 1 ) \cdot \ktrue \cdot \kfit^4,
    \end{equation}
    and\footnote{
    Note that the SNR condition in \eqref{eqn:SNR_requirement.2} includes a solution-dependent quantity, $\deltacell(\B)$, tied to the local minimizer $\B$. 
    In \ref{sec:universal_snr}, we discuss replacing this condition using a coarse universal upper bound, $\deltacell(\B) \le \deltamax + \std \ktrue \sqrt{\deff}$, which eliminates the solution dependence but leads to a more restrictive SNR condition. 
    We speculate a tighter universal upper bound for $\deltacell(\B)$ is attainable and leave it as an open question.}
    \begin{equation}\label{eqn:SNR_requirement.2}
        \frac{\deltamin}{\std}  > 
        \sqrt{ 72 \big(\sqrt{2\pi} + 1 \big) \cdot \ktrue^2 \cdot \kfit^3 \cdot \left( 5\ktrue + 2 \kfit \right) \cdot \frac{\deltacell(\B)}{\std} },
    \end{equation}
    then there exist $q \in \NN$, and two collections of sets $\bbS \coloneqq \left\{ \cS_a \subseteq [\kfit]: a \in [q] \cup \{0\} \right\}$ and $\bbT \coloneqq \left\{ \cT_a \subseteq [\ktrue]: a \in [q] \right\}$, for which the following properties hold.
    \begin{enumerate}
        \item (Simple partitions) 
        There exist $q_0$ with $0 \leq q_0 \leq q$ such that the following properties hold:
        \begin{enumerate}
            \item
            $\bbS$ is a partition of $[\kfit]$.
            \item
            $\bbT$ is a partition of $[\ktrue]$.
            \item
            $|\cS_a| = 1$ for all $a \in \{1,2,\ldots, q_0\}$.
            \item
            $|\cS_a| \geq 2$ and $|\cT_a|=1$ for all $a \in \{q_0+1,\ldots, q\}$.
        \end{enumerate}

        \item (Mutual exclusiveness)
        Suppose that $i \in \cS_a$ and $j \in \cS_b$ for $a, b \in [q] \cup \{0\}$ with $a \neq b$. If $i \neq j$, then $\bbeta_i \neq \bbeta_j$.
        
        \item (Approximation error) 
        \begin{enumerate}
            \item (one-fits-many)
            For each $a \in \{1, \cdots, q_0\}$,
            \begin{equation}\label{eqn:approx_bound.main.1}
                \frac{1}{\sigma} \left\|\, \bbeta_{i_a} - \frac{1}{\left| \cT^{\delta}_a \right| } \sum_{s \in \cT^{\delta}_a} \bthetastar_s \,\right\|
                    \lesssim   \left( \frac{\deltacell(\B)}{\std} \cdot \ktrue^2 \cdot \kfit^3 \cdot (\ktrue + \kfit) \right)^{1/2}
                    + \ktrue \cdot \deff^{1/2},
            \end{equation}
            where $i_a$ denotes the unique element in $\cS^{\delta}_a$.
            
            \item (many-fit-one)
            For each $a  \in \{q_0+1, \cdots, q\}$, 
            \begin{equation}\label{eqn:approx_bound.main.2}
                \frac{1}{\sigma} \left\| \bbeta_i - \bthetastar_{s_a} \right\|  
                    \lesssim   \left( \frac{\deltacell(\B)}{\std} \cdot \ktrue^2 \cdot \kfit^3 \cdot (\ktrue + \kfit) \right)^{1/2},   \quad \forall i \in \cS^{\delta}_a, 
            \end{equation}
            where $s_a$ denotes the unique element in $\cT^{\delta}_a$.
        \end{enumerate}

        \item (Non-association)
        Let $a \in \{1,\ldots, q\}$. If $s \in \cT_a$ and $i \in \cS_0$, then
        \begin{equation}\label{eqn:weak_association.main}
            \begin{aligned}
                \E_s \left[ \Asso_i \right] &\lesssim \left( \frac{\kfit^5 \cdot \ktrue}{\ktrue + \kfit} \cdot \frac{\std}{\deltamin} \right)^{1/2},
                \quad\text{and}\\
                \P_s \left( \vor_i \right)  &\lesssim \left( \frac{\kfit^5 \cdot \ktrue}{\ktrue + \kfit} \cdot \frac{\std}{\deltamin} \right)^{1/2}.
            \end{aligned}
        \end{equation}
    \end{enumerate}
\end{theorem}

Theorem~\ref{thm:main} formalizes the structural results discussed in Section~\ref{sec:contributions} and Figure~\ref{fig:bipartite}: all local minimizers of~$L$ involves disjoint one-fit-many and many-fit-one associations between $\{\bbeta_i\}$ and $\{\bthetastar_s\}$ (plus potentially some non-associations). In Section~\ref{sec:implications} below, we provide detailed discussions on the interpretation and implication of Theorem~\ref{thm:main}, as well the quantitative aspects and proof ideas of the theorem.

\subsection{Implications and Discussions of Theorem \ref{thm:main}}\label{sec:implications}

The main message of Theorem~\ref{thm:main} is the following: as far as the local minimizers of the negative log-likelihood is concerned, fitting a GMM can be decomposed into multiple smaller subproblems that involve simple GMMs. To further explain this result, let us elaborate.

\paragraph{Mixture fitting problems} 
Let $\kfit, \ktrue \in \NN$ and $\bThetastar = \big( \bthetastar_1, \dots, \bthetastar_{\ktrue} \big)$. 
We let $\GMM(\kfit, \bThetastar)$ denote the problem instance of fitting $\kfit$ mean estimates to a true GMM with means $\bThetastar$. 
We denote a \emph{global} minimizer of $\GMM(\kfit, \bThetastar)$ by
\begin{equation}\label{eqn:defn_objective}
\begin{aligned}
    \widehat{\B}^{\kfit, \bThetastar} &\coloneqq \arg\min_{\B \in \real^{d \times k}} L(\B \mid \bThetastar)
\end{aligned}
\end{equation}
where the population negative log-likelihood $L$ is defined in \eqref{eq:likelihood_kl}. In Section \ref{sec:two_extreme} we discussed two cases where the global minimizer is unique and simple:
\begin{itemize}
    \item 
    If $\ktrue = 1$, namely, $\bThetastar = (\bthetastar_1)$, then $\widehat{\B}^{\kfit, \bThetastar}_i =  \bthetastar_1 $ for all $i \in [\kfit]$.
    \item
    If $\kfit = 1$, then $\widehat{\B}^{\kfit, \bThetastar} = \big( \hat{\bbeta} \big)$ where $\hat{\bbeta} = \frac{1}{\ktrue} \sum_{s \in [\ktrue]} \bthetastar_s$.
\end{itemize}

\paragraph{Implications of Theorem \ref{thm:main}}
For general values of $\kfit$ and $\ktrue$, Theorem \ref{thm:main} states that  \emph{all} local minimizers $\B = ( \bbeta_1, \dots, \bbeta_{\kfit})$ of $L$ possess a common combinatorial structure. 
Specifically, each of the fitted centers $\bbeta_i$ ($i \in [\kfit]$) must correspond to one of the three possibilities below, depending on which $\cS_a$ the index $i$ belongs to. 
Recall that for any $\X \in \real^{d \times m}$ and any $\cI \subseteq [m]$, we let $\X(\cI) \in \real^{d \times |\cI|}$ denote a submatrix of $X$ such that 
$\X(\cI)_i = \X_{\cI(i)}$.
\begin{itemize}
    \item 
    \emph{(One-fits-many)}
    First, suppose that $i \in \cS_a$ for some $a \in [q_0]$, in which case $\B(\cS_a) = \bbeta_i$ is a singleton. With the notation defined in~\eqref{eqn:defn_objective}, the approximation bound~\eqref{eqn:approx_bound.main.1} states that 
    \begin{equation}\label{eq:one-fit-many_informal}
        \B(\cS_a) = \bbeta_i \approx \frac{1}{|\bThetastar(\cT_{a})|}\sum_{s\in\cT_{a}}\bthetastar_{s} =\widehat{\B}^{|\cS_{a}|, \bThetastar(\cT_{a})},
    \end{equation}
    where the last equality follows from Corollary \ref{cor:1_fit_many}. In other words, $\bbeta_i$ is close to the mean of true centers in $\bThetastar(\cT_{a})$, which in turn is the global minimizer of the one-fits-many subproblem  $\GMM(1, \bThetastar(\cT_{a}))$.
    \item 
    \emph{(Many-fit-one)}
    Second, suppose that $i \in \cS_a$ for some $a \in [q] \setminus [q_0]$, in which case  $\bThetastar(\cT_a) = (\bthetastar_{s_a})$ is a singleton. The approximation bound~\eqref{eqn:approx_bound.main.2} states that 
    \begin{equation}\label{eq:many-fit-one_informal}
        \B(\cS_a) = \big(\bbeta_i\big)_{i\in \cS_a} \approx \big(\bthetastar_{s_a}, \ldots, \bthetastar_{s_a}\big) = \widehat{\B}_i^{|\cS_{a}|, \bThetastar(\cT_{a})},
    \end{equation}
    where the last equality follows from Corollary \ref{cor:1Gaussian}. In other words, the fitted centers $\{\bbeta_i\}_{i\in\cS_a}$ are all close to the true center $\bthetastar_{s_a}$, which in turn is the global minimizer of the many-fit-one subproblem $\GMM(\kfit, \bthetastar_{s_a})$. 
    Here we remark ``one-fits-one'' is subsumed under ``many-fit-one''  configurations.  
    \item 
    \emph{(Non-association)}
    Third, if $i \in \cS_0$, then the inequalities in \eqref{eqn:weak_association.main} state that $\bbeta_i$ is not strongly associated with any of the true component means $\bthetastar_s$, $s \in [\ktrue]$. 
    This implies that most of the data points generated from the true mixture model $\nu^{*}$ are far away from $\bbeta_{i}$ in comparison with other fitted centers $\bbeta_{j}, ~\forall j \in [\ktrue] \setminus \cS_0$. 
    In other words, $\bbeta_{i}$ is virtually \emph{not} used at all to fit any of the $\ktrue$ components in the true mixture $\nu^*$.
\end{itemize}
Moreover, the simple partition property (Theorem \ref{thm:main}, Claim 1) ensures that every $i \in [\kfit]$ belong to one and only one of $\cS_a$, $a \in [q]_0$. 
As a result, the three configurations, namely, one-fits-many, many-fit-one, and non-associations, indeed exhaust all possibilities.

We highlight the symmetry between equations~\eqref{eq:one-fit-many_informal} and~\eqref{eq:many-fit-one_informal}. Therefore,  one can informally summarize the implications of Theorem \ref{thm:main} in terms of the decomposability of a given mixture problem instance. 
If $\B$ is any local minimizer of $L$, then there exist partitions $\bbS = \left\{ \cS_a \subseteq [\kfit]: a \in [q]_0 \right\}$ and $\bbT = \left\{ \cT_a \subseteq [\ktrue]: a \in [q] \right\}$ such that
\[
    \B( \cS_a ) \approx \widehat{\B}^{|\cS_a|, \bThetastar(\cT_a)},\qquad\forall a \in [q]
\]
where for all $a \in [q]$, either $|\cS_a| = 1$ or $|\cT_a|=1$ holds.
That is, any local minimizer of $L$---identified as a set of vectors---can be decomposed into the global solutions of simple sub-problems in the `one-fits-many' and `many-fit-one' settings (up to some approximation error and plus a collection of non-associated estimates). 
While not mathematically precise, this decomposability property can be written schematically as 
\[
    \GMM\left(k, \bThetastar \right) \approx \bigoplus_{a=1}^q \GMM \left( |\cS_a|, \bThetastar(\cT_a) \right),
\]
with the understanding that the decomposition is with respect to a local minimizer of the negative log-likelihood function, and different minimizers lead to different decompositions.

\begin{figure}[h]
    \centering
    \includegraphics[width=0.65\linewidth]{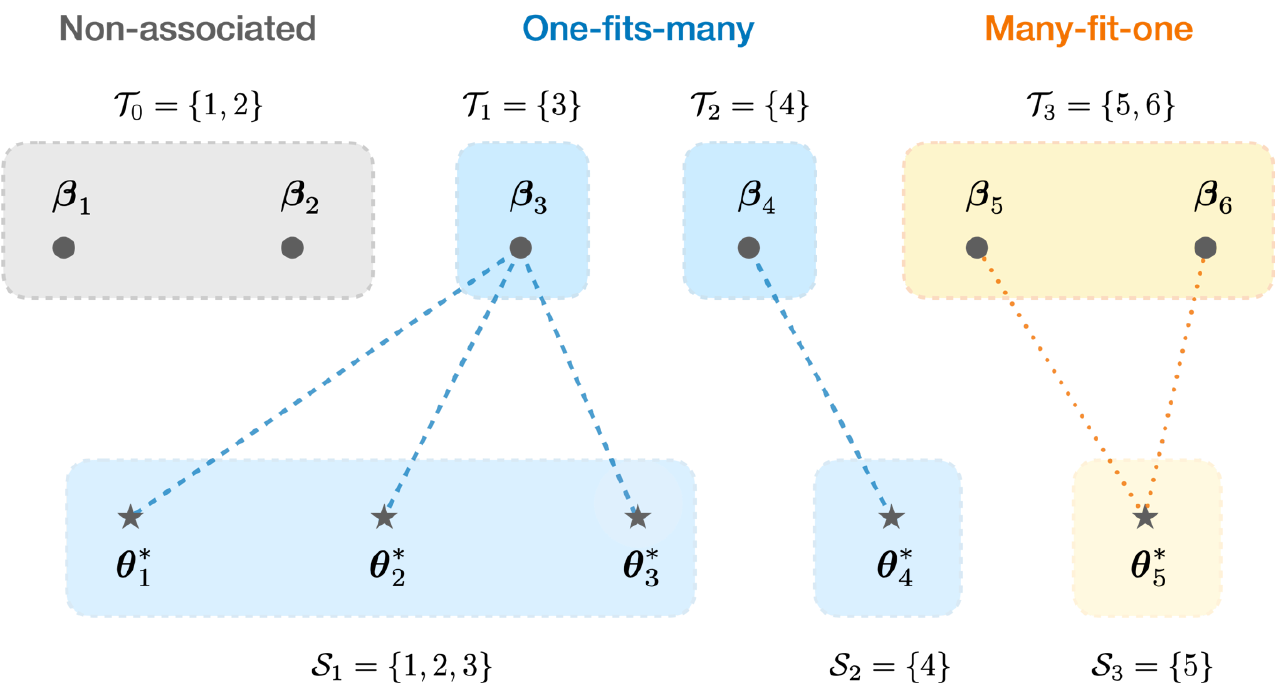}
    \caption{Illustration of the association between fitted centers $\{\protect\bbeta_{i}\}$ and true centers $\{\protect\bthetastar_{s}\}$ in a local minimum.}
    \label{fig:illustration}
\end{figure}

We illustrate these results from Theorem~\ref{thm:main} with an example.
Suppose that we are fitting $\kfit = 6$ estimated centers to data generated by a mixture of $\ktrue=5$ Gaussians with true component means $\bThetastar = \big( \bthetastar_1, \dots, \bthetastar_{5}\big)$. 
A minimizer $\B = \big( \bbeta_1, \dots, \bbeta_{6} \big) \in \real^{d \times 6}$ of $L$ correspond to $q_0 = 2$, $q = 3$ and the partitions $\bbS = \{ \cS_a: a \in [q]_0 \}$ and $\bbT = \{ \cT_a: a \in [q] \}$ with
\begin{align*}
    &\cS_0 = \{ 1, 2 \},  &&\cS_1 = \{3 \},  && \cS_2 = \{4\},  && \cS_3 = \{5, 6\},  \\
    &                       && \cT_1 = \{1, 2, 3 \},       && \cT_2 = \{4\},     && \cT_3 = \{5 \}.
\end{align*}
The association between $\{\bbeta_{i}\}$ and $\{\bthetastar_{s}\}$ is depicted by the bipartite graph in Figure \ref{fig:illustration}. 
The graph can be partitioned into three types of star graphs indexed by: (i) $\cS_0$ (non-associated centers); (ii) $(\cS_1,\cT_1)$ and $(\cS_2,\cT_2)$ (one-fits-many); (iii) $(\cS_3, \cT_3)$ (many-fit-one).

We highlight that our paper focuses on characterizing local minima structure, distinguishing sub-GMM components, and providing upper bounds on approximation errors and cross-association. 
We note that our results do not specify the quantity or sizes of these local minima or different GMM problems, i.e., values like $q_0$, $q$, $|\cS_a|$, etc., are not addressed. 
Additionally, we conjecture that the non-association set could potentially be empty, but leave its proof as an open question for future exploration.

\subsubsection{Quantitative Remarks on Theorem \ref{thm:main}} 
We discuss several quantitative aspects of Theorem \ref{thm:main}, assuming $\kfit = \ktrue$ for the convenience of discussion. 

    \paragraph{Universal upper bound on $\deltacell(\B)$} \label{sec:universal_snr}
    Theorem \ref{thm:main} and in particular the SNR requirement~\eqref{eqn:SNR_requirement.2} involve a solution-specific quantity $\deltacell(\B)$ that depends on the local minimizer $\B$ under consideration. If one desires, the following universal upper bound for $\deltacell(\B)$ can be used instead. 
    By utilizing the first-order optimality condition \eqref{eq:station_cond}, we note that 
    \begin{align*}
        \bbeta_i 
            &= \frac{\sum_{s=1}^{\ktrue} \E_s\big[ \Asso_i \sfx \big] }{\sum_{s=1}^{\ktrue} \E_s\big[ \Asso_i \big]}\\
            &= \frac{\sum_{s=1}^{\ktrue} \E_s\big[ \Asso_i \big] \cdot \bthetastar_s + \E_s \big[ \Asso_i \cdot ( \sfx - \bthetastar_s) \big] }{\sum_{s=1}^{\ktrue} \E_s\big[ \Asso_i \big]}.
    \end{align*}
    Observe that $\frac{\sum_{s=1}^{\ktrue} \E_s\big[ \Asso_i \big] \cdot \bthetastar_s }{\sum_{s=1}^{\ktrue} \E_s\big[ \Asso_i \big]}$ is in the convex hull of $\{ \bthetastar_s: s \in [\ktrue]\}$. 
   For each $s \in [\ktrue]$, we have
    \begin{align*}
        \left\| \E_s \big[ \Asso_i \cdot ( \sfx - \bthetastar_s)  \big] \right\|
            &\leq \E_s \big[ \big\| \Asso_i \cdot ( \sfx - \bthetastar_s ) \big\| \big]\\
            &\leq  \E_s \big[ \big\| \sfx - \bthetastar_s \big\| \big]\\
            &\leq \std \sqrt{\deff}.
    \end{align*}
    Hence, we have $\deltacell(\B) \leq \deltamax + \std \ktrue \sqrt{\deff} \le \deltamax + \std \ktrue \sqrt{\ktrue}$ for all local minimizers $\B$. Using this upper bound, the SNR requirement~\eqref{eqn:SNR_requirement.2} can be simplied to $\deltamin/\sigma \ge C \rho \cdot \ktrue^6$ for some universal constant $C$.
    We note that this is only a crude upper bound, and we conjecture that a tighter universal upper bound for $\deltacell(\B)$ is attainable.

    \paragraph{SNR requirements}
    By the preceding remark, Theorem \ref{thm:main} requires the SNR to satisfy $\deltamin/\sigma \geq C\rho \cdot \ktrue^6$.
    The dependence on $\ktrue^6$ is likely an artifact of our analysis due to the use of the Voronoi-cell-based arguments and union bounds; see Propositions \ref{prop:small_bdr1}--\ref{prop:proposition1}. While this paper does not focus on algorithmic guarantees, we remark that this SNR condition is sufficient for, e.g., spectral methods to learn the true model~\cite{vempala2004spectral}.
    We conjecture the conclusions of Theorem \ref{thm:main} would hold as long as $\deltamin/\sigma$ is on the order of $\ktrue^{1/2}$ or above, which corresponds to the SNR range where clustering-based mixture-learning methods succeed \cite{dasgupta1999learning, vempala2004spectral}.


    \begin{figure*}[t]
        \centering
        \begin{subfigure}[t]{0.48\textwidth}
            \centering
            \includegraphics[width=0.95\textwidth]{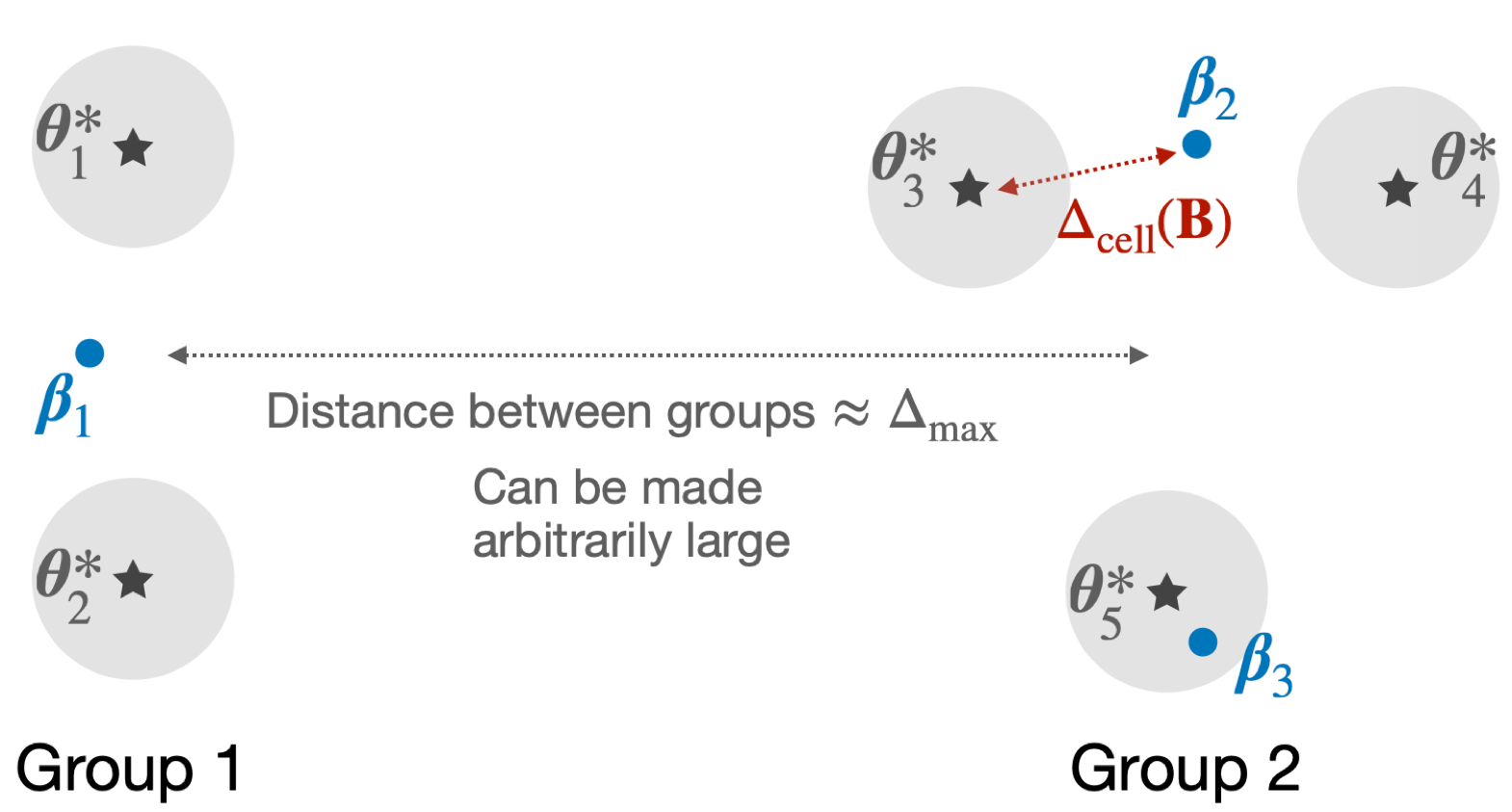}
            \caption{$\deltamax$ has little effects on the difficulty of mixture learning problem and our error bounds.}
            \label{fig:two_clusters}
        \end{subfigure}
        \hfill
        \begin{subfigure}[t]{0.48\textwidth}
            \centering
            \includegraphics[width=0.99\textwidth]{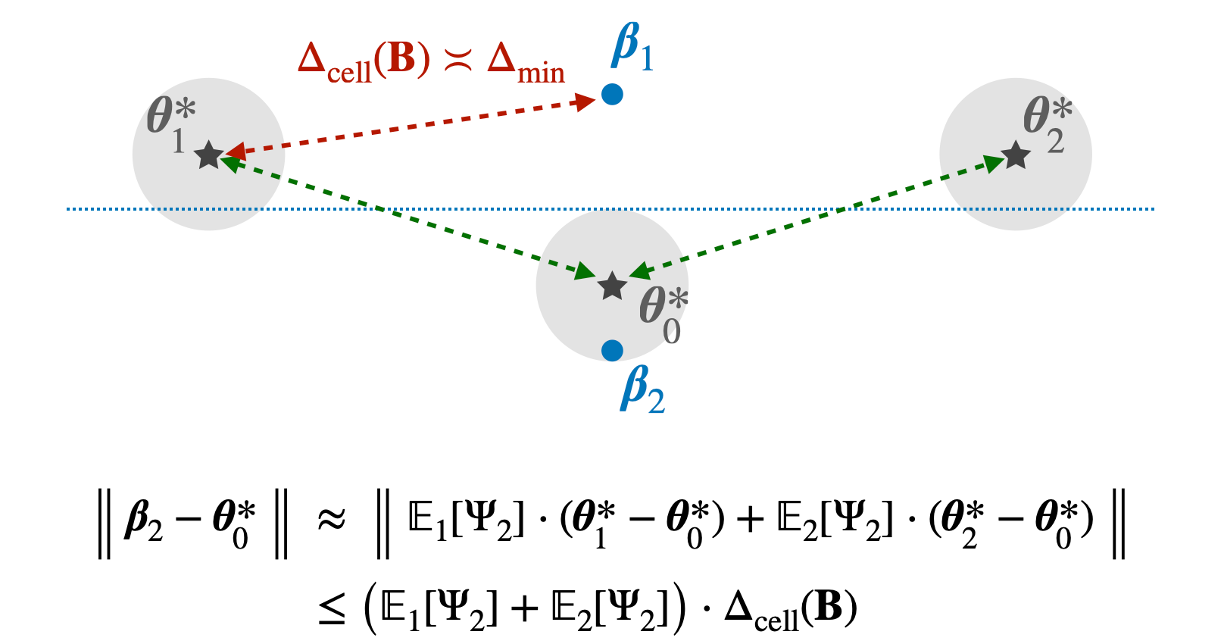}
            \caption{Our analysis of the approximation error yields an upper bound that relies on the ``cellular width'' $\deltacell(\B)$.}
            \label{fig:illustration_cell}
        \end{subfigure}
        \caption{Illustrative examples that support our remarks on Theorem \ref{thm:main}.}
        \label{fig:illustrations}
    \end{figure*}

    \paragraph{Approximation error bounds}
    Observe that the approximation error bounds \eqref{eqn:approx_bound.main.1} and \eqref{eqn:approx_bound.main.2} are of order $\big( \frac{\deltacell(\B)}{\std} \big)^{1/2} \cdot \ktrue^3$. 
    Here, we make three points interpreting this result. 
    \begin{itemize}
        \item
        Firstly, since $\deltacell(\B) \lesssim \deltamax$ (excluding the extra $\std \ktrue \sqrt{\deff}$ term), the approximation error is upper bounded by $\big( \frac{\deltamax}{\std} \big)^{1/2} \cdot \ktrue^3$. 
        \item
        Secondly, this approximation error bound is no greater than $\frac{\deltamin}{\std}$ due to the condition \eqref{eqn:SNR_requirement.2}. 
        In particular, when $\deltamin$ is sufficiently large so that $\deltacell(\B) \cdot \ktrue^6 \ll \deltamin^2$, this becomes significantly smaller that $\deltamin$. 
        \item
        Lastly, multiplying $\std$ to both sides of \eqref{eqn:approx_bound.main.1} and \eqref{eqn:approx_bound.main.2} reveals that the approximation becomes exact in the noiseless limit $\sigma \to 0$. 
    \end{itemize}
    Similarly to the SNR requirements, we believe these error bounds admit room for improvement in terms of their specific scaling with respect to the problem parameters, as showcased in Section~\ref{sec:boost} to follow.

    \paragraph{Cross-association bounds}
    We remark that the non-association property in Claim 4 of Theorem \ref{thm:main} actually holds beyond $i \in \cS_0$. 
    In Section \ref{sec:proof_main_theorem}, we present a more general Theorem~\ref{thm:master}, which establishes this property for $i \not \in \cS_0$. 
    We did not include this result in the statement of Theorem \ref{thm:main} because Theorem~\ref{thm:main} is derived from Theorem~\ref{thm:master} by rearranging the (quasi-)partitions $(\bbS,\bbT)$ via an iterative procedure described in Section \ref{sec:distillation} (Algorithm \ref{alg:distillation}), and some of the $(i,s)$ pairs could be affected by this procedure. In fact, the set $\cS_0$ (corresponding to non-associated nodes in Figure~\ref{fig:illustration}) is likely an artifact of our analysis, and we conjecture that Theorem~\ref{thm:main} in fact holds with $\cS_0 = \emptyset$.

    \paragraph{(In)dependence on problem parameters $\deltamin$ and $\deltamax$} 
    Observe that the cross-association bounds \eqref{eqn:weak_association.main} decreases as the SNR  $\frac{\deltamin}{\std}$ increases. 
    This is expected because the SNR determines the difficulty of separating one component in the mixture from others, and higher SNR makes separating components in the mixture easier. 
    On the other hand, the maximum separation $\deltamax$ has minimal impact on the problem's difficulty and our error bounds.  
    To illustrate, consider the scenario in Figure \ref{fig:two_clusters} with $\kfit = 3$ and $\ktrue = 5$, where the mixture components $\{\bthetastar_1,\ldots,\bthetastar_5\}$ can be divided into two groups that are roughly $\deltamax$ away, and a local minimizer $\B$ satisfies $\bbeta_1 \approx (\bthetastar_1 + \bthetastar_2)/2$,  $\bbeta_2 \approx (\bthetastar_3 + \bthetastar_4)/2$ and $\bbeta_3 \approx \bthetastar_5$. 
    Even if the distance $\approx \deltamax$ between the groups becomes significantly large, it barely affects the difficulty of estimating the component centers within each group.
    Specifically, the approximation errors $\| \bbeta_2 - (\bthetastar_3 + \bthetastar_4)/2 \|$ and $\| \bbeta_3 - \bthetastar_5 \|$ are mostly determined by the configurations within group 2, and these errors remain almost constant when the inter-group distance $\deltamax \to \infty$.
    
    \paragraph{Dependence on the auxiliary parameter $\deltacell(\B)$} 
    The approximation error bounds \eqref{eqn:approx_bound.main.1} and \eqref{eqn:approx_bound.main.2} depend on the ``cellular width'' parameter $\deltacell(\B)$. This dependence stems from our proofs of Proposition~\ref{prop:small_bdr2} and Lemma \ref{lem:association_bound}. 
    Using the ``$2$-estimates-and-$3$-centers'' scenario in Figure \ref{fig:illustration_cell}, we outline below the key proof steps that yield the dependence on $\deltacell(\B)$ and argue that this dependence might be unavoidable in general. Using the first-order optimality condition \eqref{eqn:stationary.1}, we obtain $\sum_{s=0}^2 \E_{s}\left[\Asso_{i}(\bbeta_{i}-\sf x)\right]= 0$ for all $i \in \{1,2\}$. 
    Therefore, the approximation error for $\bbeta_2$ satisfies $\bbeta_2 - \bthetastar_0 \approx \E_1[\Asso_2] \cdot (\bthetastar_1 - \bthetastar_0) + \E_2[\Asso_2] \cdot (\bthetastar_2 - \bthetastar_0)$, because $\E_0[\Asso_2] \approx 1$ and $\E_1[\Asso_2], \E_1[\Asso_2] \approx 0$. 
    Suppose we increase $\deltacell(\B)$ by moving $\bthetastar_1$ and $\bthetastar_2$ away from each other along the horizontal direction. Note that the association coefficient $\E_1[\Asso_2]$, which is mostly determined by the distance difference $\|\bbeta_2-\bthetastar_1\|^2 - \|\bbeta_1-\bthetastar_1\|^2$, remain largely unchanged; similarly for $\E_2[\Asso_1]$. Consequently, the error $\bbeta_2 - \bthetastar_0$ grows with $\bthetastar_1 - \bthetastar_0$ and $\bthetastar_2 - \bthetastar_0$, which in turn grow with $\deltacell(\B)$. 
    While not a formal proof, the above observation suggests that some form of dependence on $\deltacell(\B)$ may be necessary in the presence of ``one-fit-many''.

\subsubsection{Proof Ideas for Theorem \ref{thm:main}} 

Here we outline the key ideas of the proof for Theorem \ref{thm:main}, which will be detailed in Section \ref{sec:proof_main_theorem}. 
Our analysis centers on the expectation of the product of association coefficients, $\E_s\left[ \Asso_i \Asso_j \right]$, for $\bbeta_i$ and $\bbeta_j$ with respect to the component measure $f_s^*$. 
This quantity reflects the strength of interaction between the estimates $\bbeta_i$, $\bbeta_j$ and the true cluster mean $\bthetastar_s$. 
Specifically, $\E_s\left[ \Asso_i \Asso_j \right]$ is very close to $0$ unless $\bbeta_i, \bbeta_j$ are both close to $\bthetastar_s$. Also recall that the Hessian $\nabla^2 L(\B)$, whose expression~\eqref{eq:hessian} involves $\E_s\left[ \Asso_i \Asso_j \right]$, must be positive semidefinite at a local minimizer $\B$. Combining these two facts allow us to extract structural information of $\B$.

We implement the above strategy by quantifying the connection between $\E_s\left[ \Asso_i \Asso_j \right]$ and $\| \bbeta_i - \bthetastar_s \|$, as well as $\E_s\left[ \Asso_i \right]$ and $\P_s\left( \vor_i \right)$, through a set of auxiliary propositions (see Section \ref{sec:prep_propositions}) based on probabilistic and geometric arguments. 
Collecting these propositions, we argue that every \emph{second-order stationary point} of $L$ must possess a certain set of combinatorial and geometric properties, which depend on a thresholding parameter $\delta > 0$ that is used to distinguish between `large' and `small' values of $\E_s\left[ \Asso_i \Asso_j \right]$ in our analysis. 
When the signal-to-noise ratio, $\deltamin/\sigma$, is sufficiently large, we choose a specific value of $\delta$ (see Section \ref{sec:proof_master}), and the combinatorial and geometric properties reduce to those presented in Theorem \ref{thm:main}.

\subsection{Improvability of the Approximation Error Bounds in Theorem \ref{thm:main}}\label{sec:boost}


We believe it is possible to improve upon the approximation error bounds in Claim 3 of Theorem \ref{thm:main}. To explore this possibility, we focus on the exemplary setting with $\kfit=\ktrue=3$ and $d=1$, that is, ones fits three centers to a one-dimensional three-component GMM. 
By a refined analysis, we show that the coarse approximation error bounds, expressed in \eqref{eqn:approx_bound.main.1} and \eqref{eqn:approx_bound.main.2}, can be sharpened exponentially to the form $e^{-\Omega(\Delta^2/\sigma^2)}$, where $\Delta = \deltamin = 1/2 \cdot \deltamax$. 

\begin{theorem}[Tight error bounds for 3-component GMM]\label{thm:boost}
    Let $d =1$, $\kfit=\ktrue = 3$, $\sigma >0$, and $\bThetastar = (\bthetastar_1, \bthetastar_2, \bthetastar_3) = (-\Delta, 0, \Delta)$ for some $\Delta > 0$. 
    If $\Delta/\sigma > 2^{16} \cdot 3^{10} \cdot  (\sqrt{2\pi} + 1)$, then there exists a constant $C>0$ such that every local minimizer $\B = (\bbeta_1, \bbeta_2, \bbeta_3)$ of $L$ satisfies exactly one of the following possibilities (up to permutation of the indices of $\{\bbeta_i\}$ and that of $\{\bthetastar_1, \bthetastar_3\}$):
    \begin{enumerate}
        \item\label{enu:boost_case1}
        $\left\Vert \bbeta_{1}-\frac{1}{2}\left(\bthetastar_{1}+\bthetastar_{2}\right)\right\Vert \le \sigma \cdot e^{-C\Delta^{2}/\sigma^2}$,
        $\left\Vert \bbeta_{2}-\bthetastar_{3}\right\Vert \le \sigma \cdot e^{-C\Delta^{2}/\sigma^2}$,
        and $\E_{*}\left[\Asso_{3}\right]\le e^{-C\Delta^{2}/\sigma^2}$;
        \item\label{enu:boost_case2}
        $\left\Vert \bbeta_{1}-\frac{1}{2}\left(\bthetastar_{1}+\bthetastar_{2}\right)\right\Vert \le \sigma \cdot e^{-C\Delta^{2}/\sigma^2}$
        and $\left\Vert \bbeta_{i}-\bthetastar_{3}\right\Vert \le \sigma \cdot e^{-C\Delta^{2}/\sigma^2},i\in\{2,3\}$;
        \item\label{enu:boost_case3}
        $\left\Vert \bbeta_{i}-\bthetastar_{i}\right\Vert \le \sigma \cdot e^{-C\Delta^{2}/\sigma^2}$,
        $i\in\{1,2,3\}$.
    \end{enumerate}
\end{theorem}

Theorem~\ref{thm:boost}, whose proof is presented in Section~\ref{sec:proof_main_boost}, provides tighter bounds on the approximation error and the association coefficient, $\E_{*}\left[\Asso_{3}\right]$, both of which decrease exponentially fast as a function of the SNR $\Delta/\sigma$. 
As a result, as the SNR increases, each fitted enters is either exponentially close to a true center
(or to the mean of two), or its association coefficient (and hence its Voronoi cell) becomes exponentially small. In fact, since $L$ has no other local minima near the true centers $\bthetastar$ due to existing local results on GMM \cite{balakrishnan2014EM}, the approximation errors in Possibility~\ref{enu:boost_case3} above are actually zero, in which case $\B=\bThetastar$ is the exact global minimizer.

Compared to Theorem~\ref{thm:main}, Theorem~\ref{thm:boost} provides more refined information on the possible configurations that a local minimizer can admit. 
For instance, Theorem~\ref{thm:boost} rules out the possibility of having one center $\bbeta_{1}\approx\frac{1}{2}(\bthetastar_{1,}+\bthetastar_{3})$ fitting two non-adjacent true centers, thanks to the one-dimension assumption $d=1$. 
Theorem~\ref{thm:boost} also eliminates the possibility that one estimate fits all three true centers, i.e., $\bbeta_{1}\approx\frac{1}{3}(\bthetastar_{1}+\bthetastar_{2}+\bthetastar_{3})$, while the other two estimates $\bbeta_{2},\bbeta_{3}$ are far away from the true centers and have non-association. 
Such a one-fit-all configuration fails to capture the mixture structure of the data, but we note that formally excluding this possibility in a general setting is not easy due to the issue of ``local minimum at infinity''.\footnote{When $\bbeta_{2}$ and $\bbeta_{3}$ go to infinity, and hence, move away from the data, the negative log-likelihood $L$ approaches a constant function of $(\bbeta_2, \bbeta_3)$, and the function $\bbeta_1 \mapsto L((\bbeta_1,\bbeta_2,\bbeta_3))$ approaches the negative log-likelihood of fitting a single Gaussian to a mixture of three. In this limiting case, $\bbeta_{1}=\frac{1}{3}(\bthetastar_{1}+\bthetastar_{2}+\bthetastar_{3})$ is indeed a local minimum (see Corollary~\ref{cor:1_fit_many}).}
Theorem~\ref{thm:boost} addresses this issue when $d=1$ and shows that any \emph{finite} $\bbeta$ of this form cannot be a local minimum.
It remains open whether one can exclude Scenario~\ref{enu:boost_case1} in Theorem~\ref{thm:boost} that involves a non-associated center.

The proof of Theorem~\ref{thm:boost} builds upon the coarse characterization of local minima state in Theorem \ref{thm:main} and its general form Theorem~\ref{thm:master}. The coarse characterization restricts the local minimizers of $L$ into a small neighborhood of a few ideal solutions, such as $\left(\bthetastar_{1}, \bthetastar_{2},\bthetastar_{3}\right)$ and $\left((\bthetastar_{1}+\bthetastar_{2})/2,\bthetastar_{3},\bthetastar_{3}\right)$. 
By further exploiting the first-order stationary condition~\eqref{eq:station_cond}, we show that the local minimizers must be exponentially close to these ideal solutions within this neighborhood. 
We believe that this strategy could be extended beyond the one-dimensional three-component setting, and we leave this to future work.

\paragraph{The case of a mis-specified model} 
The techniques used in the proof of Theorem \ref{thm:boost} also apply to the case where the number of mixture components is under-specified.
Recall that Corollary~\ref{cor:1_fit_many} addresses the scenario where a single Gaussian is used to fit a mixture of three Gaussians, in which case the overall mean $\bbeta_{1}=\frac{1}{3}\left(\bthetastar_{1}+\bthetastar_{2}+\bthetastar_{3}\right)$ is the only stationary point. 
The corollary below deals with fitting a two-component GMM to a three-component GMM. 

\begin{corollary}[Tight bounds for underfitting 3-component GMM]\label{cor:under}
    Let $d =1$, $\kfit=2$, $\ktrue = 3$, $\sigma >0$ and $\bThetastar = (\bthetastar_1, \bthetastar_2, \bthetastar_3) = (-\Delta, 0, \Delta)$ for some $\Delta > 0$.
    If $\Delta/\sigma > 2^{16} \cdot 3^{10} \cdot  (\sqrt{2\pi} + 1)$, then there exists a constant $C>0$ for which every local minimizer $\B = (\bbeta_1, \bbeta_2, \bbeta_3)$ of $L$ satisfies (up to permutation of the indices of $\{\bbeta_i\}$ and that of $\{\bthetastar_1, \bthetastar_3\}$)
    \begin{equation}\label{eq:under}
    \begin{aligned}
        \left\Vert \bbeta_{1}-\frac{1}{2}\left(\bthetastar_{1}+\bthetastar_{2}\right)\right\Vert 
            &\le\std \cdot e^{-C\Delta^{2}/\std^{2}}\quad\text{and}\quad\\
        \left\Vert \bbeta_{2}-\bthetastar_{3}\right\Vert 
            &\le\std \cdot e^{-C\Delta^{2}/\std^{2}}.
    \end{aligned}
    \end{equation}
\end{corollary}

We prove Corollary \ref{cor:under} in Section~\ref{sec:proof_under} using intermediate results from the proof of Theorem~\ref{thm:boost}. 

We remark that Corollary~\ref{cor:under} is related to the recent work by Dwivedi \textit{et al}.~\cite{dwivedi2018misspecified}, who study a similar under-specification setting. 
They consider fitting a \emph{symmetric} 2-mixture $\frac{1}{2}\mathcal{N}(\bbeta_{1},\std^{2})+\frac{1}{2}\mathcal{N}(-\bbeta_{1},\std^{2})$ to a 3-mixture of the form $\frac{1}{4}\mathcal{N}(\bthetastar_{1}(1+\alpha),\std^{2})+\frac{1}{4}\mathcal{N}(\bthetastar_{1}(1-\alpha),\std^{2})+\frac{1}{2}\mathcal{N}(-\bthetastar_{1},\std^{2})$. 
They provide finite-sample convergence rates for the EM algorithm, assuming that the EM starts from an initial solution sufficiently close to the global minimizer. 
Note that in the setting of Corollary~\ref{cor:under}, we effectively establish that there is no other local minimizer besides the global minimizer $\B \approx \left(\frac{1}{2}\left(\bthetastar_{1}+\bthetastar_{2}\right),\bthetastar_{3}\right)$, and thus the EM converges to this solution from any initialization.

\section{Discussion}\label{sec:Discussion}

In this paper, we study the population negative log-likelihood of GMMs with a general number of components, and show that all local minimizers have the common structure that partially reveals the locations of the true components in the global minimum. Our findings have many algorithmic implications and point to a variety of ways of improving iterative methods for learning GMMs---we mention a few here. 
\begin{enumerate}
    \item[(i)] Better initialization schemes, such as those that pick $\kfit$ initial centers that are far away from each other, are useful for finding a local minimizer free from the one-fits-many sub-configuration and therefore facilitate the retrieval of a global minimimizer. This ideas underlies the k-means++ and several other clustering algorithms~\cite{ball1967promenade, astrahan1970speech, barakbah2005optimized, barakbah2009pillar}.
    \item[(ii)] Given any local minimizer (e.g., obtained by the EM algorithm), it may be possible to refine the solution iteratively and recover the remaining components by deflating the components already recovered.  
    \item[(iii)] Another natural idea is to resolve the one-fits-many and many-fit-one sub-configurations by adding more center estimates or combining redundant centers. In fact, several improved versions of the EM are based on this ``split-and-merge'' operation~\cite{ueda1998smem,ueda2000split,zhang2003split_merge}.
\end{enumerate}

A particularly promising algorithmic idea suggested by our results is \emph{over-parameterization}, which helps avoid spurious local minima.  
In particular, when the number of fitted centers $\kfit$ is sufficiently larger than the number of true components $\ktrue$, one expects that the one-fits-many sub-configuration is unlikely to occur. Therefore, an overparmeterized iterative method is likely to converge to a local minimizers that involves only many-fit-one and near-empty subconfigurations, which can then be pruned to identify most (if not all) true components and hence the global minimizer.
A similar idea was proposed in the work by Dasgupta and Schulman~\cite{dasgupta2007probabilistic}, who propose to over-specify $\kfit$ in the EM algorithm and, after convergence, merge fitted centers that are close to each other (corresponding to many-fit-one) and remove fitted centers with low mixing weights (corresponding to non-association). Recent work on non-parametric maximum likelihood estimation~\cite{polyanskiy2020self,yan2023wasserstein}, which can be interpreted as an extreme form of overpameterization with $\kfit \to \infty$, demonstrates great potential of this approach.

There are several avenues for future research to improve upon the analysis presented in this paper. 
First, it is of immediate interest to obtain sharper results regarding the SNR requirement, approximation error bounds, and the strength of association across different sub-configurations (see Remarks on Theorem \ref{thm:main} in Section \ref{sec:main_theorem}). 
Achieving these goals may require developing analytical techniques that do not rely on Voronoi-cells, as their use in our analysis essentially impedes the propagation of influence from a component beyond its own cell, which possibly have led to suboptimal guarantees. 
Another direction for future research is to extend our population-level results to the finite-sample setting, for which the uniform concentration and localization techniques developed in \cite{mei2016landscape,dwivedi2020singularity} could be useful. 
In addition, it would be interesting to explore the low-SNR regime, where the mixture components have small or even no separation, as the structures of the local minima may become more complicated in such cases. 
Finally, it would be valuable to understand whether the phenomenon of structured local minima holds more generally in other mixture and latent variable models; for example, empirical evidences in a recent study \cite{buhai2020empirical} suggest that this phenomenon may be universal.

\section{Deferred Proofs from Section \ref{sec:prelim}}\label{sec:proof_stationary}
\subsection{Proofs of Lemma \ref{lem:gradient_NLL} and Lemma \ref{lem:hessian_NLL}}

\begin{proof}[Proof of Lemma \ref{lem:gradient_NLL}]
    Recall that $f_i(\x):=\phi(\x\mid\bbeta_{j}, \sigma^2)=\frac{1}{\sqrt{2\pi} \sigma}\exp\big(-\left\Vert \x-\bbeta_{i}\right\Vert ^{2}/2 \sigma^2 \big)$, and hence,
    \begin{equation}\label{eq:Gaussian_derivative}
        \frac{\partial f_{i}(\x)}{\partial\bbeta_{j}}
            = \begin{cases}
                    \frac{1}{\sigma^2} \cdot f_{i}(\x)\cdot(\x-\bbeta_{i}) &   \text{if }j = i,\\
                0   &   \text{if }j \neq i.
            \end{cases}
    \end{equation}
    Since $f:=\frac{1}{\kfit}\sum_{i\in[\kfit]}f_{i}$, it follows that
    \begin{align*}
        \frac{\partial}{\partial\bbeta_{j}}L(\B) 
            & = - \frac{\partial}{\partial\bbeta_{j}} \E_{*}\left[\log f(\sf x)\right]\\
            & \stackrel{(a)}{=} -\E_{*}\left[\frac{\frac{1}{\kfit}\sum_{i \in [\kfit]} \frac{\partial}{\partial\bbeta_{j}}f_{i}(\sf x)}{f(\sf x)}\right]\\
            & \stackrel{(b)}{=} -\E_{*}\left[\frac{\frac{1}{\kfit} \cdot \frac{1}{\sigma^2} \cdot f_{j}(\sf x)\cdot(\sf x-\bbeta_{j})}{f(\sf x)}\right]\\
            & \stackrel{(c)}{=} \frac{1}{\sigma^2} \cdot  \E_{*}\big[\Asso_{j}\cdot(\bbeta_{j}-\sf x)\big],
    \end{align*}
where (a) follows from exchanging the expectation and differentiation using the dominated convergence theorem\footnote{We omit explicitly mentioning this argument in the rest of this paper.}, (b) follows from \eqref{eq:Gaussian_derivative}, and (c) follows from the definition of $\Asso_j$ in \eqref{eq:asso_rv} (see Definition \ref{defn:coeff_asso} as well).
\end{proof}

\begin{proof}[Proof of Lemma \ref{lem:hessian_NLL}]
    First of all, we compute the partial derivative of $\asso_{i}(\x)$. For each $j\in[\kfit]$, we observe that
    \begin{align}
        \frac{\partial}{\partial\bbeta_{j}}\asso_{i}(\x) 
            &= \frac{\partial}{\partial\bbeta_{j}} \frac{\frac{1}{k} f_i(\x)}{ f(\x) }      \nonumber\\
            &= \frac{1}{k} \frac{ \frac{\partial}{\partial\bbeta_{j}} f_i(\x) \cdot f(\x) - f_i(\x) \cdot \frac{\partial}{\partial\bbeta_{j}} f(\x) }{ f(\x)^2 }      \nonumber\\
            &\stackrel{(a)}{=} \frac{1}{k} \bigg[ \frac{\frac{1}{\sigma^2} \cdot f_{i}(\x)\cdot(\x-\bbeta_{i}) \cdot \delta_{ij}}{f(\x)}    \nonumber\\
                &\qquad\quad
                - \frac{f_{i}(\x)\cdot \frac{1}{k} \cdot \frac{1}{\sigma^2} \cdot f_{j}(\x) \cdot (\x-\bbeta_{j}) }{f^2(\x)} \bigg]     \nonumber\\
            &= \frac{1}{\sigma^2} \big[ \asso_i(\x) \cdot \delta_{ij} - \asso_i(\x) \asso_j(\x) \big] \cdot (\x - \bbeta_j ).
                \label{eq:asso_derivative}
    \end{align}
    where (a) follows from \eqref{eq:Gaussian_derivative}. 
    Therefore, we obtain that 
    \begin{align*}
        \std^2 \cdot \frac{\partial^{2}}{\partial\bbeta_{i}\partial\bbeta_{j}}L(\B)
            &\stackrel{(a)}{=} \std^2 \cdot \frac{\partial}{\partial\bbeta_{j}} \left( \frac{1}{\sigma^2} \cdot \E_{*}\big[\Asso_{i} \cdot (\bbeta_{i}-\sf x)\big] \right) \\
            &\stackrel{(b)}{=} \E_{*}\left[\left(\frac{\partial}{\partial\bbeta_{j}}\Asso_{i}\right)\cdot(\bbeta_{i}-\sf x)^{\top}+\Asso_{i}\cdot\left(\frac{\partial}{\partial\bbeta_{j}}\bbeta_{i}^{\top}\right)\right]\\
            &\stackrel{(b)}{=}\E_{*}\left[\left(\frac{1}{\sigma^2} \big[ \Asso_i \cdot \delta_{ij} - \Asso_i \Asso_j \big] \cdot (\sf x - \bbeta_j )\right)\cdot(\bbeta_{i}-\sf x)^{\top} \right]
                 + \E_{*} \left[\Asso_{i}\cdot\left(\frac{\partial}{\partial\bbeta_{j}}\bbeta_{i}^{\top}\right)\right]    \\
            &= \frac{1}{\sigma^2} \cdot \E_{*}\Big[ \Asso_i \Asso_j \cdot ( \bbeta_j - {\sf x} ) ( \bbeta_i - {\sf x} )^{\top} \Big]
                 + \delta_{ij} \cdot \E_{*} \left[ \Asso_i \cdot \Big\{ \Id_{d} - \frac{1}{\std^2} \cdot ( \bbeta_i - {\sf x} ) ( \bbeta_i - {\sf x} )^{\top}   \Big\} \right],
    \end{align*}
    where (a) follows from Lemma \ref{lem:gradient_NLL}, (b) is by the chain rule, and (c) is due to \eqref{eq:asso_derivative}. 
    The proof is complete after rearranging terms.
\end{proof}

\subsection{Proof of Theorem~\ref{thm:equiv}}\label{sec:proof_equiv}

We apply the Stein's identity \cite{stein1981estimation} to prove Theorem~\ref{thm:equiv}. 
Specifically, we make use of the following multivariate version of the Stein's identity specialized to the identity covariance setting. 
\begin{lemma}[Stein's identity]\label{lem:stein}
    Suppose ${\sf x} \sim\mathcal{N}(\bmu,\sigma^{2}\Id_{d})$ and $g:\real^{d}\to\real$ is a differentiable function. Then 
    \[
        \E\left[g({\sf x}) \cdot({\sf x}-\bmu)\right]=\sigma^{2} \cdot \E\left[\nabla g({\sf x})\right].
    \]
\end{lemma}

Next, we present a proof of Theorem~\ref{thm:equiv}. 
\begin{proof}[Proof of Theorem~\ref{thm:equiv}]    
    Fix an arbitrary $i\in[\kfit]$ and compute the derivative $\nabla\asso_{i}(\x)$ with respect to $\x$. 
    Recall from \eqref{eq:asso} that 
    \begin{align*}
        \psi_{i}(\x)
            &:=\frac{e^{-\left\Vert \x-\bbeta_{i}\right\Vert ^{2}/(2\std^{2})}}{\sum_{j\in[\kfit]}e^{-\left\Vert \x-\bbeta_{j}\right\Vert ^{2}/(2\std)^{2}}}\\
            &=\frac{1}{\sum_{j\in[\kfit]}e^{\left(\left\Vert \x-\bbeta_{i}\right\Vert ^{2}-\left\Vert \x-\bbeta_{j}\right\Vert ^{2}\right)/(2\std^{2})}}.
    \end{align*}
    Taking the derivative, we have 
    \begin{align*}
        \nabla\asso_{i}(\x)
            & =- \sum_{j\in[\kfit]}\frac{e^{\left(\left\Vert \x-\bbeta_{i}\right\Vert ^{2}-\left\Vert \x-\bbeta_{j}\right\Vert ^{2}\right)/(2\std)^{2}} \cdot(\bbeta_{j}-\bbeta_{i}) \cdot \std^{-2}}{\left(\sum_{\ell\in[\kfit]}e^{\left(\left\Vert \x-\bbeta_{i}\right\Vert ^{2}-\left\Vert \x-\bbeta_{\ell}\right\Vert ^{2}\right)/(2\std)^{2}}\right)^{2}}\\
            & =\sum_{j\in[\kfit]}\frac{e^{-\left\Vert \x-\bbeta_{i}\right\Vert ^{2}/(2\std)^{2}}e^{-\left\Vert \x-\bbeta_{j}\right\Vert ^{2}/(2\std)^{2}}}{\left(\sum_{\ell\in[\kfit]}e^{-\left\Vert \x-\bbeta_{\ell}\right\Vert ^{2}/(2\std)^{2}}\right)^{2}}\cdot(\bbeta_{i}-\bbeta_{j})\cdot\std^{-2}\\
            & =\std^{-2}\sum_{j\in[\kfit]} \asso_{i}(\x)\asso_{j}(\x) \cdot (\bbeta_{i}-\bbeta_{j})\\
            & =\std^{-2} \cdot \bigg(\asso_{i}(\x) \cdot \bbeta_{i} -\sum_{j\in[\kfit]}\asso_{i}(\x)\asso_{j}(\x) \cdot \bbeta_{j}\bigg),
    \end{align*}
    where the last step follows from the fact that $\sum_{j\in[\kfit]}\asso_{j}(\x)=1$.
    Noting that ${\sf x} \sim\mathcal{N}(\bthetastar_{s},\std^{2}\Id_{d})$ under the density $f^*_s$, we apply the Stein's identity (Lemma \ref{lem:stein}) to obtain that for all $(s,i)\in[\ktrue] \times [\kfit]$,
    \begin{align}
        \E_{s}\left[\Asso_{i} \cdot {\sf x} \right] 
            & =\bthetastar_{s} \cdot \E_{s}\left[\Asso_{i}\right] + \std^{2}\cdot\E_{s}\left[\nabla\asso_{i}({\sf x})\right]\nonumber \\
            & =\bthetastar_{s} \cdot \E_{s}\left[\Asso_{i}\right] + \bbeta_{i} \cdot \E_{s}\left[\Asso_{i}\right]-\sum_{j\in[\kfit]}\bbeta_{j}\cdot \E_{s}\left[\Asso_{i}\Asso_{j}\right].\label{eq:PsiX_after_stein}
    \end{align}
    
    Recall from the stationarity condition \eqref{eq:station_cond} that $\bbeta$ is a stationary point of $L$ if and only if 
    \[
        \bbeta_{i}=\frac{\sum_{s\in[\ktrue]}\E_{s}\left[\Asso_{i}{\sf x}\right]}{\sum_{s\in[\ktrue]}\E_{s}\left[\Asso_{i}\right]},
        \qquad\forall i\in[\kfit].
    \]
    Combining this condition with the above expression for $\E_{s}\left[\Asso_{i} \cdot \sf x\right]$ in \eqref{eq:PsiX_after_stein}, we obtain that for all $i\in[\kfit]$,
    \begin{align*}
        \sum_{s\in[\ktrue]}\E_{s}\left[\Asso_{i}\right] \cdot \bbeta_{i} 
            & = \sum_{s\in[\ktrue]}\bigg(\bthetastar_{s} \cdot \E_{s}\left[\Asso_{i}\right] + \bbeta_{i} \cdot \E_{s}\left[\Asso_{i}\right] - \sum_{j\in[\kfit]}\bbeta_{j} \cdot \E_{s}\left[\Asso_{i}\Asso_{j}\right]\bigg).
    \end{align*}
    Rearranging terms yields 
    \begin{align*}
        \sum_{s\in[\ktrue]}\sum_{j\in[\kfit]}\bbeta_{j} \cdot \E_{s}\left[\Asso_{i}\Asso_{j}\right] 
            & =\sum_{s\in[\ktrue]}\bthetastar_{s} \cdot \E_{s}\left[\Asso_{i}\right],\quad\forall i\in[\kfit],
    \end{align*}
    which is equivalent to the condition in \eqref{eq:equiv}. Thus, the proof is complete. 
\end{proof}

\subsection{Proof of Corollary \ref{cor:span}} \label{sec:proof_span}

\begin{proof}[Proof of Corollary \ref{cor:span}]
    Given $\B = (\bbeta_1, \dots, \bbeta_k) \in \real^{d \times k}$, let $k' \in \NN_+$ denote the number of distinct elements (=columns) in $\B$. 
    Note that $k' \leq k$, and that we may assume $\bbeta_1, \dots, \bbeta_{k'}$ are all distinct without loss of generality (by permuting the order). 
    For each $i \in [k']$, let $\mathcal{I}_{i} := \left\{ i'\in[\kfit]:\bbeta_{i'}=\bbeta_{i}\right\}$, $m_i := \big| \mathcal{I}_i \big|$, and $\Asso'_i := \sum_{i' \in \mathcal{I}_i} \Asso_{i'} = m_i \cdot \Asso_i$. 

    Then we observe that for all $i \in [k']$,
    \begin{align}
        \sum_{s \in [\ktrue]} \bthetastar_s \cdot \E_s \left[ \Asso'_i \right]
            &= \sum_{i' \in \mathcal{I}_i} \sum_{s \in [\ktrue]} \bthetastar_s \cdot \E_s \left[ \Asso_{i'} \right]     \nonumber\\
            &= \sum_{i' \in \mathcal{I}_i} \sum_{s \in [\ktrue]} \sum_{j\in[\kfit]}\bbeta_{j} \cdot \E_{s}\left[\Asso_{i'}\Asso_{j}\right]   &&\because \text{\eqref{eq:equiv}}\nonumber\\
            &= \sum_{i' \in \mathcal{I}_i} \sum_{s \in [\ktrue]} \sum_{j\in[k']} \sum_{j' \in \mathcal{I}_j } \bbeta_{j} \cdot \E_{s}\left[\Asso_{i'}\Asso_{j'}\right]      \nonumber\\
            &= \sum_{s \in [\ktrue]} \sum_{j\in[k']}  \bbeta_{j} \cdot \E_{s}\left[\Asso'_{i}\Asso'_{j}\right].     \label{eqn:condition_balance}
    \end{align}
    Letting $\vec{\Asso'} := ( \Asso'_1, \dots, \Asso'_{k'})^{\top} \in \real^{k' \times 1}$, we can rewrite the condition \eqref{eqn:condition_balance} as the following matrix equation:
    \begin{equation}\label{eqn:condition_balance_matrix}
        \bThetastar \cdot \mathbf{\Phi}' = \mathbf{B}' \cdot \E_*\left[ \vec{\Asso}' \cdot {\vec{\Asso'}}^{\top} \right]  
    \end{equation}
    where
    \begin{align*}
            \mathbf{\Phi}' &= \begin{bmatrix} \E_1[ \vec{\Asso'}] & \dots & \E_{\ktrue}[\vec{\Asso'} ] \end{bmatrix}^{\top} \in \real^{\ktrue \times k'},\quad\text{and}\\
            \B' &= \B([k']) \in \real^{d \times k'}.
    \end{align*}

    Next, we claim that the $k'$-by-$k'$ matrix $\E_*\big[ \vec{\Asso}' \cdot {\vec{\Asso'}}^{\top} \big]$ is invertible. 
    If we assume otherwise, there exists a nonzero vector $\u \in \real^{k'}$ such that
    \[
        \E_* \left[ (\u^T \vec{\Asso'})^2 \right]
            = \u^T \E_*\left[ \vec{\Asso}' \cdot {\vec{\Asso'}}^{\top} \right] \u
            = 0,
    \]
    which implies that 
    \[
        \u^T \vec{\Asso'}
            = \sum_{i \in [k']} u_i \cdot m_i \cdot \Asso_i
            = 0
    \]
    almost surely with respect to $f^*$. 
    This is equivalent to the condition that
    \[
        \sum_{i \in [k']} u_i \cdot m_i \cdot \asso_i(\x) = 0
    \]
    almost everywhere (with respect to the Lebesgue measure) because $f^*(\x) > 0$ for all $\x \in \real^d$. 
    Because $m_i \geq 1 > 0$ for all $i \in [k']$ and $\{ \bbeta_i: ~i \in [k']\}$ are distinct, the $k'$ functions $m_i \cdot \asso_i: \x \mapsto m_i \cdot e^{-\|\x - \bbeta_i\|^2 / (2\sigma^2)}$, $i \in [k']$ are linearly independent. 
    This renders a contradiction.

    Finally, we obtain from \eqref{eqn:condition_balance_matrix} using the invertibility of $\E_*\big[ \vec{\Asso}' \cdot {\vec{\Asso'}}^{\top} \big]$ that
    \[
        \mathbf{B}'
            = \bThetastar \cdot \mathbf{\Phi}' \cdot \E_*\left[ \vec{\Asso}' \cdot {\vec{\Asso'}}^{\top} \right]^{-1}.  
    \]
    Therefore, $\bbeta_i$ belongs to the column space of $\bThetastar$ for all $i \in [k']$, and the proof of Corollary \ref{cor:span} is complete.
\end{proof}
\section{Proof of Theorem~\ref{thm:main} (the Main Theorem)}\label{sec:proof_main_theorem}
This section is dedicated to proving the main theorem of this paper, namely, Theorem \ref{thm:main}. 
To this end, we start in Section \ref{sec:master_statement} by presenting Theorem \ref{thm:master}, which serves as the master theorem of this work and from which Theorem \ref{thm:main} is derived as a special instance. 
In Section \ref{sec:master_proof}, we prove Theorem \ref{thm:master} by relying on five propositions that we introduce in Section \ref{sec:prep_propositions}. 
We then proceed in Section \ref{sec:proof_main} to derive Theorem \ref{thm:main} from Theorem \ref{thm:master} by choosing specific parameters (Section \ref{sec:remarks_theorem}) and making slight adjustments (Section \ref{sec:distillation}) when necessary. 

Throughout this section, we assume that $\B = (\bbeta_{1},\ldots,\bbeta_{\kfit})\in\real^{d\times\kfit}$ is an arbitrary, yet fixed, local minimizer of $L$, unless stated otherwise. 
To avoid clutter, we may omit the explicit reference to $\B$ in our notation, such as when referring to the coefficient of association $\asso_i(\x)$ or the Voronoi cell $\cV_i$, when it is clear from the context.


\subsection{A Complete Version of Theorem \ref{thm:main}}\label{sec:master_statement}

To state the master theorem, we define a relaxed notion of a partition of a finite set. To better understand of this relaxation, it is helpful to first recall the formal definition of a partition.
\begin{definition}\label{defn:partition}
    A family of sets $\bbS$ is a \emph{partition} of $\cX$ if and only if all of the following conditions hold:
    \begin{itemize}
        \item
        $\bbS$ does not contain the empty set, i.e., $\emptyset \notin \bbS$.
        \item
        The sets in $\bbS$ cover $\cX$, i.e., $\bigcup_{A \in \bbS} A = \cX$.
        \item
        The elements of $\bbS$ are pairwise disjoint, i.e., $\forall A, B \in \bbS$, if $A \neq B$ then $A \cap B = \emptyset$.
    \end{itemize}
\end{definition}

We now introduce the notion of a quasi-partition, which relaxes the third requirement above.
\begin{definition}[Quasi-partition of a set]\label{defn:quasi_partition}
    Let $\bbS$ be a family of sets and $\cI, \cJ \subset \bbS$ such that $\cI \cap \cJ = \emptyset$. 
    $\bbS$ is an \emph{$(\cI, \cJ)$-quasi-partition} of $\cX$ if and only if all of the following conditions hold:
    \begin{itemize}
        \item
        $\bbS$ does not contain the empty set, i.e., $\emptyset \notin \bbS$.
        \item
        The sets in $\bbS$ cover $\cX$, i.e., $\bigcup_{A \in \bbS} A = \cX$.
        \item
        The elements of $\bbS$ are pairwise disjoint except for pairs across $\cI$ and $\cJ$, i.e., $\forall A, B \in \bbS$ such that $A \neq B$, 
        if $A \cap B \neq \emptyset$, then either (1) $A \in \cI$ and $B \in \cJ$ or (2) $A \in \cJ$ and $B \in \cI$.
    \end{itemize}
\end{definition}

\begin{remark}
    Suppose that $\bbS$ is an $(\cI, \cJ)$-quasi-partition of $\cX$ for some $\cI, \cJ \subset \bbS$. 
    Then for each $x \in \cX$, there exist either one or two sets in $\bbS$ that contain $x$. 
    Moreover, if there are two sets that contain $x$, then one should belong to $\cI$ with the other being a member of $\cJ$.  
\end{remark}

\subsubsection{Statement of the Master Theorem}

\begin{theorem}[Master theorem]\label{thm:master}
    Let $\kfit, \ktrue \in \NN$, $\bThetastar \in \real^{d \times \ktrue}$, and $\B \in \real^{d \times \kfit}$ be a local minimum of $L(\,\cdot \,|\, \bTheta^*)$.
    If $\delta \in \real_+$ satisfies $\frac{4 \ktrue \cdot \sigma}{\deltamin} < \delta \leq \frac{1}{ 18 (\sqrt{2\pi} + 1 ) \cdot  \kfit^4 }$, 
    then there exist $q \in \NN$, and two collections of sets $\bbS^{\delta} := \left\{ \cS^{\delta}_a \in [\kfit]: a \in [q]_0 \right\}$ and $\bbT^{\delta} := \left\{ \cT^{\delta}_a \in [\ktrue]: a \in [q] \right\}$, for which the following properties hold.
    \begin{enumerate}
        \item (Simple quasi-partitions) 
        There exist $q_0$ with $0 \leq q_0 \leq q$ such that the following properties hold:
        \begin{enumerate}
            \item
            $\bbS^{\delta} \setminus \{ \cS^{\delta}_0 \}$ is a $\big( \{ \cS^{\delta}_a \}_{a=1}^{q_0}, \{ \cS^{\delta}_a \}_{a=q_0+1}^{q} \big)$-quasi-partition of $[\kfit] \setminus \cS^{\delta}_0$.
            \item
            $\bbT^{\delta}$ is a partition of $[\ktrue]$.
            \item
            $|\cS^{\delta}_a| = 1$ for all $a \in [q_0]$.
            \item
            $|\cS_a| \geq 2$ and $|\cT_a|=1$ for all $a \in [q] \setminus [q_0]$.
        \end{enumerate}

        \item (Mutual exclusiveness)
        Suppose that $i \in \cS^{\delta}_a$ and $j \in \cS^{\delta}_b$ for $a, b \in [q]_0$ with $a \neq b$. If $i \neq j$, then $\bbeta_i \neq \bbeta_j$.
        
        \item (Approximation error) 
        \begin{enumerate}
            \item (one-fits-many)
            Let 
            \[
                \deltacell(\B) \coloneqq \max_{j \in [\kfit]} \max_{s \in [\ktrue]: \bthetastar_s \in \vor_j} \| \bthetastar_s - \bbeta_j \|.
            \]
            For $a \in [q_0] $, 
            \begin{equation}\label{eqn:approx_bound.1}
                \begin{aligned}
                    \frac{1}{\sigma} \left\|\, \bbeta_{i_a} - \frac{1}{\left| \cT^{\delta}_a \right| } \sum_{s \in \cT^{\delta}_a} \bthetastar_s \,\right\|
                        &\leq
                        18 \left(\sqrt{2\pi} + 1 \right) \cdot \kfit^3 \cdot \left( 5\ktrue + 2 \kfit \right) \cdot \frac{\deltacell(\B)}{\std} \cdot \delta\\
                        &\quad + \frac{1}{|\cT^{\delta}_a|} \cdot \left\{ \frac{8\ktrue}{3 }\cdot \frac{1}{\delta} + 2\ktrue \cdot (\sqrt{d}+4) + \frac{4}{3\sqrt{2\pi}} \right\}\\
                        &\quad  + 3.
                \end{aligned}
            \end{equation}
            where $i_a$ denotes the unique element in $\cS^{\delta}_a$.
            
            \item (many-fit-one)
            For $a \in [q] \setminus [q_0]$, 
            \begin{equation}\label{eqn:approx_bound.2}
                \frac{1}{\sigma} \left\| \bbeta_i - \bthetastar_{s_a} \right\|
                    \leq \frac{2 \ktrue }{ \delta  },   \qquad \forall i \in \cS^{\delta}_a,
            \end{equation}
            where $s_a$ denotes the unique element in $\cT^{\delta}_a$.
        \end{enumerate}

        \item (Near-empty cross-association)
        Let $(a,b) \in [q] \times [q]_0$ such that $a \neq b$. 
        If $s \in \cT^{\delta}_a$ and $i \in \cS^{\delta}_b$, then
        \begin{equation}\label{eqn:weak_association}
            \begin{aligned}
                \E_s \left[ \Asso_i \right] &\leq 9 \big(\sqrt{2\pi} + 1 \big) \cdot \tilde{\kappa}(a) \cdot \delta, \quad\text{and}\\
                \P_s \left( \vor_i \right)  &\leq 3 \big(\sqrt{2\pi} + 1 \big) \cdot \tilde{\kappa}(a) \cdot \delta
            \end{aligned}
        \end{equation}
        where
        \[
            \tilde{\kappa}(a) =
            \begin{cases}
                \kfit^4    &\text{if }a \in [q_0],\\
                \kfit^3    &\text{if }a \in [q] \setminus [q_0].
            \end{cases}
        \]
    \end{enumerate}
\end{theorem}

\subsubsection{Construction of the Collections of Sets \texorpdfstring{$\bbS^{\delta}$}{Sd} and \texorpdfstring{$\bbT^{\delta}$}{Td} in Theorem \ref{thm:master}}\label{sec:partition_construction}
Here, we describe an algorithm that produces the collections of sets, $\bbS^{\delta}$ and $\bbT^{\delta}$, in the statement of Theorem \ref{thm:master}. 
For the convenience of presentation, we define two types of sets. 
For any $i \in [\kfit]$ and for any $\delta \geq 0$, we let 
\begin{equation}\label{eq:setA}
\begin{aligned}
    \setA_i^{\delta} \coloneqq \bigg\{ s \in [\ktrue] ~\bigg|~ \iota_{\B}(\bthetastar_s) = i ~~~\text{and}~~~
        \max_{j, j' \in [\kfit]}  \frac{ \| \bbeta_j - \bbeta_{j'} \| }{\sigma} \cdot \E_s \left[ \Asso_j \Asso_{j'} \right] < \delta \bigg\}.
\end{aligned}
\end{equation}
For any $s \in [\ktrue]$ and for any $\delta > 0$, let
\begin{equation}\label{eqn:entangled}
    \cE_s^{\delta} \coloneqq \left\{ i \in [\kfit] ~\bigg|~  \max_{j \in [\kfit] \setminus \{i\}} \frac{ \| \bbeta_i - \bbeta_{j} \| }{\sigma} \cdot \E_s \left[ \Asso_i \Asso_j \right] \geq \delta \right\}.
\end{equation}

\begin{remark}
    Intuitively, $\setA_i^{\delta}$ is the set of true cluster indices that are exclusively associated to $\bbeta_i$ at level $\delta$. 
    Similarly, we can interpret $\cE_s^{\delta}$ as the set of indices of estimated centers that contend for the possesion of $\bthetastar_s$. 
\end{remark}

Then we describe a procedure to construct covers of $[\ktrue]$ and $[\kfit]$, whose pseudocode can be found in Algorithm \ref{alg:partition}. 
For any inputs $\delta > 0$ and $\B \in \real^{d \times k}$, Algorithm \ref{alg:partition} outputs $q \in \NN$ and $q \in \NN$ along with two collections of sets, namely, 
$\bbS^{\delta} = \left\{ \cS^{\delta}_a \right\}_{a=0}^q$ and $\bbT^{\delta} = \left\{ \cT^{\delta}_a \right\}_{a=1}^q$. 
    
\begin{algorithm}[h!]
\caption{Construction of partitions of $[\ktrue]$ and $[\kfit]$} 
\label{alg:partition}
\textbf{Input:} $\delta \in \real_+$, $\B \in \real^{d \times k}$\\
\textbf{Output:} $q_0$, $q$, $\bbS^{\delta} = \{ \cS^{\delta}_a \}_{a=0}^{q}$, $\bbT^{\delta} = \{ \cT^{\delta}_a \}_{a=1}^q$
\begin{algorithmic}[1]
    \STATE Initialize $a \gets 0$
    \FOR {$i = 1,2,\ldots, \kfit$}
        \IF { $\setA_i^{\delta} \neq \emptyset$ }
            \STATE $a \gets a + 1$
            \STATE $\cS^{\delta}_a \gets \{ i \}$
            \STATE $\cT^{\delta}_a \gets \setA_i^{\delta}$
                \hfill\COMMENT{See \eqref{eq:setA} for definition of $\setA_i^{\delta}$}
        \ENDIF
    \ENDFOR
    \STATE $q_0 \gets a$
    \STATE $\cR^{\delta} \gets [\ktrue] \setminus \big(\bigcup_{a'=1}^{q_0} \cT^{\delta}_{a'} \big)$
        \hfill\COMMENT{`Remaining'}
    \FOR {$s = 1, 2, \dots, \ktrue$}
        \IF{ $s \in \cR^{\delta}$ }
            \STATE $a \gets a + 1$
            \STATE $\cS^{\delta}_a \gets \cE_s^{\delta}$ 
                \hfill\COMMENT{See \eqref{eqn:entangled} for definition of $\cE_s^{\delta}$}
            \STATE $\cT^{\delta}_a \gets \{ s \}$
        \ENDIF
    \ENDFOR
    \STATE $q \gets a$
    \STATE $\cS^{\delta}_0 \gets [\kfit] \setminus \bigcup_{a=1}^q \cS^{\delta}_a$
\end{algorithmic} 
\end{algorithm}


\subsection{Proof of Theorem \ref{thm:master}}\label{sec:master_proof}
\subsubsection{Five Preparatory Propositions}\label{sec:prep_propositions}
Here we present five key propositions that establish useful properties of association coefficients (Propositions \ref{prop:small_bdr1} and \ref{prop:exclusive}), geometry of local minima (Propositions \ref{prop:proposition1} and \ref{prop:small_bdr2}), and combinatorial properties of the covers $\bbS^{\delta}, \bbT^{\delta}$ constructed by Algorithm \ref{alg:partition} (Proposition \ref{prop:cover_properties}). 
We will use these propositions to prove Theorem \ref{thm:master} in Section \ref{sec:proof_master}. 
Proofs of Propositions \ref{prop:small_bdr1} through \ref{prop:cover_properties} are deferred to the Appendix (Appendices \ref{sec:proof_proposition.2} through \ref{sec:proof_covering}).


\paragraph{Strength of associations}
First of all, we present a proposition that controls the strength of association between the estimate $\bbeta_i$ and the true mean $\bthetastar_s$, measured in two quantities $\P_s\left( \vor_i \right)$ and $\E_s[\Asso_i]$.

\begin{proposition}[Association bounds]\label{prop:small_bdr1}
    Let $\B \in \real^{d \times k}$ be an arbitrary ordered set of vectors. 
    For any $(s, i) \in [\ktrue] \times [\kfit]$ and any sequence $\big( \alpha_j \in \real_+: j \in [\kfit]\setminus\{i\} \big)$, 
    if $\bthetastar_{s}\notin\intr\vor_{i}$, then 
    \begin{enumerate}
        \item
        $\P_s\left( \vor_i \right) \leq M_{si}(\boldsymbol{\alpha, \B})$, and
        \item
        $\E_{s}\left[\Asso_{i}\right] \leq 3 M_{si}(\boldsymbol{\alpha, \B})$.
    \end{enumerate}
    where
    \begin{align*}
        M_{si}(\boldsymbol{\alpha, \B})
        &\coloneqq \left(\sqrt{2\pi} + 1 \right) \cdot \kfit^2 \sum_{j\in [\kfit]\setminus\{i\}} \max\left\{ 1, ~\frac{1}{\alpha_j} \right\} \cdot \E_{s}\left[ \Asso_i \Asso_j  \right] \cdot \exp \left( 3\alpha_j \frac{ \| \bbeta_i - \bbeta_j \| }{\sigma} \right).
    \end{align*}
\end{proposition}


Proposition \ref{prop:small_bdr1} states that if $\bthetastar_{s}\notin\intr\vor_{i}$ and $\E_{s}\left[\Asso_{i}\Asso_{j}\right]$ is small for \emph{all} $j \in [\kfit]\setminus\{i\}$, then the Voronoi cell associated with $\bbeta_{i}$ must have a small size (under $\nu^*_s$) as well as a small coefficient of association to $\nu_{s}^{*}$. We present a proof of Proposition \ref{prop:small_bdr1} in Appendix \ref{sec:proof_proposition.2}.

\begin{corollary}\label{cor:small_bdr}
    Let $\B \in \real^{d \times k}$ be an arbitrary ordered set of vectors and let $(s, i) \in [\ktrue] \times [\kfit]$.
    If $\bthetastar_{s}\notin\intr\vor_{i}$, then 
    \begin{enumerate}
        \item
        $\P_s\left( \vor_i \right) \leq
            3 \left(\sqrt{2\pi} + 1 \right) \cdot \kfit^2 \sum_{j\in [\kfit]\setminus\{i\}} \frac{ \| \bbeta_i - \bbeta_j \| }{\sigma} \cdot \E_{s}\left[ \Asso_i \Asso_j  \right]$; and
        \item
        $\E_{s}\left[\Asso_{i}\right] \leq
            9 \left(\sqrt{2\pi} + 1 \right) \cdot \kfit^2 \sum_{j\in [\kfit]\setminus\{i\}} \frac{ \| \bbeta_i - \bbeta_j \| }{\sigma} \cdot \E_{s}\left[ \Asso_i \Asso_j  \right]$.
    \end{enumerate}
\end{corollary}
\begin{proof}[Proof of Corollary \ref{cor:small_bdr}]
    Let $z \in \real_+$ and Let $f_z: \real_+ \to \real_+$ be a function such that $f_z(\alpha) = \max\big\{1, \frac{1}{\alpha}\big\} \cdot \exp( \alpha \cdot z )$.
    Observe that 
    \begin{align*} 
        \inf_{\alpha \in \real_+} f_z(\alpha)
            &= \inf \left\{ \inf_{\alpha \in (0,1)} \Big\{\frac{1}{\alpha} \cdot e^{\alpha \cdot z }\Big\},~ \inf_{\alpha \in [1, \infty)]} \big\{ e^{ \alpha \cdot z } \big\} \right\}\\
            &= \begin{cases}
                e \cdot z   & \text{if }z > 1 \text{ (minimum attained at }\alpha= \frac{1}{z}),\\
                e^z         & \text{if }z \leq 1,
            \end{cases}\\
            &\leq e \cdot z.
    \end{align*}
    Then the conclusion of this corollary immediately follows from Proposition \ref{prop:small_bdr1}.
\end{proof}


\paragraph{Exclusion principle for associations}
Recall from \eqref{eq:setA} that for any $i \in [\kfit]$ and any $\delta \geq 0$,
\begin{align*}
    \setA_i^{\delta} \coloneqq \bigg\{ s \in [\ktrue] ~\bigg|~ \iota_{\B}(\bthetastar_s) = i ~~~\text{and}~~~
        \max_{j, j' \in [\kfit]}  \frac{ \| \bbeta_j - \bbeta_{j'} \| }{\sigma} \cdot \E_s \left[ \Asso_j \Asso_{j'} \right] < \delta \bigg\}.
\end{align*}

\begin{proposition}\label{prop:exclusive}
    Let $\B \in \real^{d \times k}$ be an arbitrary ordered set of vectors, let $i\in[\kfit]$, and let $\delta \in \real_+$. 
    If $\setA_i^{\delta} \neq \emptyset$ and $\delta \leq \frac{1}{18 \left(\sqrt{2\pi} + 1 \right) \cdot \kfit^4}$, the following three statements hold:
    \begin{enumerate}
        \item
        $\bbeta_i \neq \bbeta_j$ for all $j \in [\kfit] \setminus \{i\}$;
        \item
        $ 1 - \E_s \left[ \Asso_i \right] 
            \leq 9 \left(\sqrt{2\pi} + 1 \right) \cdot \kfit^4 \cdot \delta$ for all $s \in \setA_i^{\delta}$;
        \item
        $\P_s \left( \vor_i^c \right) 
            \leq 3 \left(\sqrt{2\pi} + 1 \right) \cdot \kfit^4 \cdot \delta$ for all $s \in \setA_i^{\delta}$.
    \end{enumerate}
\end{proposition}
Proposition \ref{prop:exclusive} states that when the parameter $\delta$ is sufficiently small, the mixture components with $s \in \setA^{\delta}_i$ are almost exclusively associated to $\beta_i$. 
This claim is quantified in two different measures, namely, $\E_s \left[ \Asso_i \right]$ and $\P_s \left[ \vor_i \right]$. 
We present a proof of Proposition \ref{prop:exclusive} in Appendix \ref{sec:proof_proposition}.


\paragraph{Proximity of $\B$ controlled by product of association coefficients} 
Next, we discuss several useful properties of $\B$, which is a local minimum of $L$. 
First, we present a proposition that establishes several upper bounds on the proximity of the estimates, $\bbeta_i$, to the true centers, $\bthetastar_s$.

\begin{proposition}[Proximity bound, I]\label{prop:proposition1}
    Let $\B$ be a local minimum of $L$. For all $i, j \in [\kfit]$ such that $i \neq j$, the following three statements hold:
    \begin{enumerate}
        \item
        $ \frac{\| \bbeta_i - \bbeta_j \|^2}{ \sigma^2 } \leq \frac{\E_{*}[\Asso_i + \Asso_j]}{ \E_{*}[ \Asso_i \cdot \Asso_j ]}$, 
        \item
        $\min_{s \in [\ktrue]} \frac{\| \bbeta_i - \bthetastar_s \|^2}{ \sigma^2 } 
            \leq \frac{2\left(  \E_{*}\left[\Asso_{i}\right] + 1 \right) }{ \E_{*}\left[ \Asso_i \cdot \Asso_j \right]  }$, and
        \item
        $ \frac{\| \bbeta_i - \bthetastar_s\|^2 }{ \sigma^2 }
        \leq 2 \ktrue \cdot \frac{ \E_{*}\left[\Asso_{i}\right] + 1 }{\E_{s}\left[\Asso_{i} \cdot \Asso_{j} \right]}$ for all $s \in [\ktrue]$.
    \end{enumerate}
\end{proposition}

Proposition \ref{prop:proposition1} states that if $\E_{*}\left[\Asso_{i}\Asso_{j}\right]$ is large for some $i, j \in [\kfit]$ with $i \neq j$, then $\bbeta_{i}$ and $\bbeta_{j}$ must be both close to each other (Claim 1). 
Moreover, there exists a true center $\bthetastar_{s}$ that is close to both $\bbeta_i$ and $\bbeta_j$ (Claim 2). 
Claim 3 of Proposition \ref{prop:proposition1} provides a weaker, yet still useful, upper bound for our analysis. 
We postpone our proof of Proposition \ref{prop:proposition1} to Appendix \ref{sec:proof_proposition.1}.

\begin{corollary}\label{cor:proximity}
    Let $\B$ be a local minimum of $L$, let $\delta \in \real_+$, and $s \in [\ktrue]$. 
    If $\cE^{\delta}_s \neq \emptyset$, then for all $i \in \cE^{\delta}_s$,
    \[
        \frac{\| \bbeta_i - \bthetastar_s\| }{ \sigma }
            \leq \frac{2\ktrue}{\delta}.
    \]
\end{corollary}
\begin{proof}[Proof of Corollary \ref{cor:proximity}]
    Observe that if $\cE^{\delta}_s \neq\emptyset$, then $|\cE^{\delta}_s| \geq 2$ by definition. 
    Choose any $i \in \cE^{\delta}_s$ and any $j \in \cE^{\delta}_s \setminus \{i\}$. 
    Then
    \begin{align*}
         \frac{\| \bbeta_i - \bbeta_j \|}{ \sigma } 
            &\overset{(a)}{\leq} \frac{\E_{*}[\Asso_i + \Asso_j]}{ \frac{\| \bbeta_i - \bbeta_j \|}{ \sigma }  \cdot \E_{*}[ \Asso_i \cdot \Asso_j ]}\\
            &\overset{(b)}{\leq} \frac{\ktrue \cdot \E_{*}[\Asso_i + \Asso_j]}{ \frac{\| \bbeta_i - \bbeta_j \|}{ \sigma }  \cdot \E_{s}[ \Asso_i \cdot \Asso_j ]}\\
            &\overset{(c)}{\leq} \frac{\ktrue}{\delta}
    \end{align*}
    where (a) follows from Proposition \ref{prop:proposition1}, Claim 1; (b) follows from $\E_{*}[ \Asso_i \cdot \Asso_j ] \geq \frac{1}{\ktrue} \E_{s}[ \Asso_i \cdot \Asso_j ]$; and (c) is from $i, j \in \cE^{\delta}_s$. Next, we obtain by Proposition \ref{prop:proposition1} that
    \begin{align*}
        \frac{\| \bbeta_i - \bthetastar_s\|^2 }{ \sigma^2 }
            &\leq 2 \ktrue \cdot \frac{ \E_{*}\left[\Asso_{i}\right] + 1 }{\E_{s}\left[\Asso_{i} \cdot \Asso_{j} \right]} \\
            &= 2 \ktrue \cdot \frac{ \E_{*}\left[\Asso_{i}\right] + 1 }{ \frac{\| \bbeta_i - \bbeta_j \|}{ \sigma }  \cdot \E_{s}\left[\Asso_{i} \cdot \Asso_{j} \right]} \cdot \frac{\| \bbeta_i - \bbeta_j \|}{ \sigma } \\
            &\leq 2 \ktrue \cdot \frac{2}{\delta} \cdot \frac{\ktrue}{\delta} \\
            &= \left(\frac{2\ktrue}{\delta}\right)^2.
    \end{align*}
\end{proof}

\paragraph{Approximation error bound for local minima of $L$}
Here, we argue that the mean estimates $\bbeta_i$ (for $i \in [\kfit]$) in the so-called `one-fits-many' configuration are well approximated by the barycenter of several true component means, $\bthetastar_s$. 
We recall the definition of $\setA_i^{\delta}$ from \eqref{eq:setA}. 
With aid of the set $\setA_i^{\delta}$, we state an approximation error bound as follows.

\begin{proposition}[Proximity bound, II]\label{prop:small_bdr2}
    Let $\B \in \real^{d \times k}$ be a local minimum of $L$. 
    If $\delta \in \real_+$ satisfies that $\frac{4 \ktrue \cdot \sigma}{\deltamin} < \delta \leq \frac{1}{ 18 (\sqrt{2\pi} + 1 ) \cdot  \kfit^4 }$, 
    then for all $i \in [\kfit]$ with $\setA_i^{\delta} \neq \emptyset$, 
    \begin{equation}\label{eqn:prop_approx}
    \begin{aligned}
            \frac{1}{\sigma} \left\| \bbeta_i - \frac{1}{\left| \setA_i^{\delta} \right| } \sum_{s \in \setA_i^{\delta}} \bthetastar_s \right\|
            &\leq
                 18 \left(\sqrt{2\pi} + 1 \right) \cdot \left( \frac{ 5 \ktrue }{|\setA_i^{\delta}|} \cdot \frac{\deltacell(\B)}{\std} + \kfit \cdot \frac{\tdeltacell^{i,\delta}(\B)}{\std} \right) \cdot \delta  + \frac{ 4\ktrue }{ |\setA_i^{\delta} | }  \cdot \frac{1}{\delta}   \\
            &\quad
                + \left\{ \frac{1}{|\setA_i^{\delta}|} \cdot \left( 2\ktrue \cdot (\sqrt{d}+4) + \sqrt{\frac{2}{\pi}} \right) + 3 \right\}.
    \end{aligned}
    \end{equation}
    where $\deltacell(\B) \coloneqq \max_{j \in [\kfit]} \max_{s \in [\ktrue]: \bthetastar_s \in \vor_j} \| \bthetastar_s - \bbeta_j \|$ and $\tdeltacell^{i, \delta}(\B) \coloneqq \min_{s \in \setA_i^{\delta}} \max_{s' \in \setA_i^{\delta}} \| \bthetastar_s - \bthetastar_{s'} \|$. 
\end{proposition}


We remark that $\tdeltacell^{i, \delta}(\B) \leq 2 \deltacell(\B)$. 
Also, we conjecture that the $\sqrt{d}$ term in \eqref{eqn:prop_approx} is an artifact of our analysis, and thus, is removable. 
Our proof of Proposition \ref{prop:small_bdr2} is deferred to Appendix \ref{sec:proof_proposition.3}.

\paragraph{Properties of the covers constructed by Algorithm \ref{alg:partition}}
Lastly, we discuss some properties of the collections $\left\{ \cS^{\delta}_a \right\}_{a=0}^q$ and $\left\{ \cT^{\delta}_a \right\}_{a=1}^q$ produced by Algorithm \ref{alg:partition}. 
Specifically, we argue that these are collections of non-empty sets that cover $[\kfit]$ and $[\ktrue]$, respectively, which possess useful combinatorial properties.

\begin{proposition}\label{prop:cover_properties}
    Let $\delta \in \real_+$ and $\B \in \real^{d \times k}$ be a local minimum of $L$. 
    Let $\bbS^{\delta} = \left\{ \cS^{\delta}_a \right\}_{a=0}^q$ and $\bbT^{\delta} = \left\{ \cT^{\delta}_a \right\}_{a=1}^q$ be the collections of sets produced by Algorithm \ref{alg:partition}. 
    If $\frac{4 \ktrue \cdot \sigma}{\deltamin} < \delta \leq \frac{1}{ 18 (\sqrt{2\pi} + 1 ) \cdot  \kfit^4 }$, then $\bbS^{\delta}$ and $\bbT^{\delta}$ possess the following properties.
    \begin{enumerate}
        \item
        For all $a \in [q]$, both $\cS^{\delta}_a$ and $\cT^{\delta}_a$ are not empty. 
        Moreover, 
        \begin{enumerate}
            \item
            $|\cS^{\delta}_a|=1$ for all $a \in [q_0]$, and
            \item
            $|\cS^{\delta}_a|\geq2$ and $|\cT^{\delta}_a|=1$ for all $a \in [q] \setminus [q_0]$.
        \end{enumerate} 
        \item
        $\left\{ \cS^{\delta}_a \right\}_{a=0}^q$ covers $[\kfit]$ and $\left\{ \cT^{\delta}_a \right\}_{a=1}^q$ covers $[\ktrue]$, i.e.,
        \[
            \bigcup_{a=0}^q \cS^{\delta}_a = [\kfit]
            \qquad\text{and}\qquad
            \bigcup_{a=1}^q \cT^{\delta}_a = [\ktrue].
        \]
        \item
        Let $a \in [q]$ and $s \in [\ktrue]$. 
        If $s \in \cT^{\delta}_a$, then $\iota_{\B}(\bthetastar_s) \in \cS^{\delta}_a$. 
        \item
        $\bbT^{\delta}$ is a collection of disjoint sets, i.e., $\cT^{\delta}_a \cap \cT^{\delta}_b = \emptyset$ for all $a, b \in [q]$ such that $a \neq b$.
        \item
        $\bbS^{\delta}$ is a collection of partially disjoint sets. That is, 
        \begin{enumerate}
            \item
            $\cS^{\delta}_a \cap \cS^{\delta}_0 = \emptyset$ for all $a \in [q]$;
            \item
            $\cS^{\delta}_a \cap \cS^{\delta}_b = \emptyset$ for all $a, b \in [q_0]$ such that $a \neq b$, and
            \item
            $\cS^{\delta}_a \cap \cS^{\delta}_b = \emptyset$ for all $a, b \in [q]\setminus[q_0]$ such that $a \neq b$.
        \end{enumerate}
    \end{enumerate}
\end{proposition}

We defer the proof of Proposition \ref{prop:cover_properties} to Appendix \ref{sec:proof_covering}.

\begin{remark}
    Proposition \ref{prop:cover_properties} states that $\bbT^{\delta}$ is a partition of $[\ktrue]$. 
    Nevertheless, $\bbS^{\delta}$ falls short of being a partition of $[\kfit]$ for two reasons: (i) $\cS^{\delta}_0$ may or may not be empty; and (ii) we do not have disjointness for index pairs $(a,b) \in [q_0] \times ([q]\setminus[q_0])$. 
    Indeed, this motivated us to define the notion of quasi-partitions as in Definition \ref{defn:quasi_partition}. 
    While our current analysis cannot resolve these issues, we conjecture that $\cS^{\delta}_0 = \emptyset$ and $\cS^{\delta}_a \cap \cS^{\delta}_b = \emptyset$ for all $(a,b) \in [q_0] \times ([q]\setminus[q_0])$, i.e., $\bbS^{\delta} \setminus \{ \cS^{\delta}_{0} \}$ forms a partition of $[\kfit]$. 
    
\end{remark}

\subsubsection{Completing the Proof of Theorem \ref{thm:master}}\label{sec:proof_master}
\begin{proof}[Proof of Theorem \ref{thm:master}]
    We prove the four claims one by one, using the propositions stated in Section \ref{sec:prep_propositions}.

    \bigskip
    \textit{\underline{Proof of Claim 1 (Simple quasi-partition).}}
    This is straightforward from Proposition \ref{prop:cover_properties}.

    \bigskip
    \textit{\underline{Proof of Claim 2 (Mutual exclusiveness).}}
    First, suppose that $i \in \cS^{\delta}_a$ for some $a \in [q_0]$. 
    Then $\setA^{\delta}_i \neq \emptyset$, and thus, $\bbeta_j \neq \bbeta_i$ for all $j \in [\kfit] \setminus \{i\}$ by Proposition \ref{prop:exclusive}, Claim 1. 

    Second, suppose that $i \in \cS^{\delta}_a$ for some $a \in [q] \setminus [q_0]$. 
    Then $\cT^{\delta}_a = \{ s_a \}$ for some $s_a \in [\ktrue]$. 
    Let $b \in [q]_0 \setminus \{a\}$ and assume that there exists $j \in \cS^{\delta}_b \setminus \{i\}$ such that $\bbeta_i = \bbeta_j$. 
    Then $\asso_i (\x) = \asso_j (\x)$ for all $\x \in \real^d$, and thus, $j \in \cS^{\delta}_{a}$ because $i \in \cE^{\delta}_{s_a}$ implies $j \in \cE^{\delta}_{s_a}$. 
    It follows that $j \in \cS^{\delta}_a \cap \cS^{\delta}_b$, however, this contradicts the disjointness property of $\bbS^{\delta}$ proved above (Claim 1 of this theorem).

    \bigskip
    \textit{\underline{Proof of Claim 3 (Approximation error bounds).}}
    \begin{enumerate}[label=(\alph*)]
        \item
        For $a \in [q_0]$, notice that $\cS^{\delta}_a = \{i_a\}$ and $\cT^{\delta}_a = \setA^{\delta}_{i_a}$. 
        Then the inequality \eqref{eqn:approx_bound.1} immediately follows from Proposition \ref{prop:small_bdr2}, cf. \eqref{eqn:prop_approx}.
        \item
        For $a \in [q] \setminus [q_0]$, note that $\cT^{\delta}_a = \{ s_a \}$ and $\cS^{\delta}_a = \cE^{\delta}_{s_a} = \left\{ i \in [\kfit]: \max_{j \in [\kfit] \setminus \{i\}} \frac{\|\bbeta_i - \bbeta_j \|}{\sigma} \cdot \E_{s_a} \left[ \Asso_i \Asso_j \right] \geq \delta \right\}$. 
        Thus, we obtain the inequality \eqref{eqn:approx_bound.2} by applying Corollary \ref{cor:proximity}.
    \end{enumerate}

    \bigskip
    \textit{\underline{Proof of Claim 4 (Weak cross-associations).}}
    First, suppose that $s \in \cT^{\delta}_a$ for some $a \in [q_0]$, and let $i_a$ denote the unique element in $\cS^{\delta}_a$. 
    We observe that $s \in \setA^{\delta}_{i_a}$ by construction of the sets $\cS^{\delta}_a$ and $\cT^{\delta}_a$ in Algorithm \ref{alg:partition}. 
    Then it follows from Proposition \ref{prop:exclusive} that for all $i \in [\kfit] \setminus \{i_a\}$,
    \begin{align*}
        \E_s \left[ \Asso_i \right]
            &\leq 1 - \E_s \left[ \Asso_{i_a} \right]
            \leq 9 \left(\sqrt{2\pi} + 1 \right) \cdot \kfit^4 \cdot \delta,
        \quad\text{and}\\
        \P_s \left( \vor_i \right)
            &\leq \P_s \left( \vor_{i_a}^c \right)
            \leq 3 \left(\sqrt{2\pi} + 1 \right) \cdot \kfit^4 \cdot \delta.
    \end{align*}

    Second, suppose that $s \in \cT^{\delta}_a$ for some $a \in [q] \setminus [q_0]$. 
    Let $s_a$ denote the unique element in $\cT^{\delta}_a$. 
    By Claim 3 of Proposition \ref{prop:cover_properties}, if $\iota_{\B}(\bthetastar_s) = i$, then $i \in \cS^{\delta}_a$. 
    Then we observe that if $i \in [\kfit] \setminus \cS^{\delta}_{a}$, then (1) $\iota_{\B}(\bthetastar_s) \neq i$, and thus, $\bthetastar_{s_a} \not\in \intr\vor_i$; and (2) $\max_{j \in [\kfit] \setminus \{i\}} \frac{\| \bbeta_{i} - \bbeta_{j}\|}{\sigma}\cdot\E_{s_a}\left[ \Asso_i \cdot \Asso_j \right] < \delta$ by definition of $\cE^{\delta}_{s_a}$. 
    Therefore, it follows from Proposition \ref{prop:small_bdr1} (Corollary \ref{cor:small_bdr}) that for all $i \in [\kfit] \setminus \cS^{\delta}_{a}$,
    \begin{align*}
        \E_{s_a} \left[ \Asso_i \right]
            &\leq 9 \left(\sqrt{2\pi} + 1 \right) \cdot \kfit^3 \cdot \delta,
            \quad\text{and}\\
        \P_{s_a} \left( \vor_i \right)
            &\leq 3 \left(\sqrt{2\pi} + 1 \right) \cdot \kfit^3 \cdot \delta.
    \end{align*}

\end{proof}


\subsection{Obtaining Theorem \ref{thm:main} from Theorem \ref{thm:master}}\label{sec:proof_main}

\subsubsection{Remarks that Simplify Theorem \ref{thm:master}}\label{sec:remarks_theorem}
In this section, we make three remarks on Theorem \ref{thm:master}, which will be used to simplify the conditions and conclusions of the theorem. 
First of all, we remark on the minimum separation between the components in the true Gaussian mixture model required to apply Theorem \ref{thm:master}. 
Second, we discuss the reduction of dimensionality from the dimension of ambient space, $d$, to the effective dimension (Definition \ref{defn:dim_eff}) of the given problem instance, $\deff$, which leads to a significant improvement of approximation error bound \eqref{eqn:approx_bound.1} when $\ktrue \ll d$. 
Lastly, we determine the optimal value for the auxiliary parameter $\delta \in \real_+$ that approximately minimizes the approximation error bound \eqref{eqn:approx_bound.1}.

\paragraph{Minimum requirement on the signal-to-noise ratio}
Note that Theorem \ref{thm:master} is vacuous unless there exists $\delta \in \real_+$ satisfies $\frac{4 \ktrue \cdot \sigma}{\deltamin} < \delta \leq \frac{1}{ 18 (\sqrt{2\pi} + 1 ) \cdot  \kfit^4 }$. 
This requires the minimum signal-to-noise ratio, cf. \eqref{eq:separation}, to exceed a certain threshold. 
Specifically, in order to apply Theorem \ref{thm:master}, we need the minimum SNR requirement \eqref{eqn:SNR_requirement} to be satisfied:
\begin{equation*}
    \frac{\deltamin}{\sigma} > 72 (\sqrt{2\pi} + 1 ) \cdot \ktrue \cdot \kfit^4.
\end{equation*}

\paragraph{Effective dimensionality}
For any given instance $\GMM(\kfit, \bThetastar)$ of Gaussian mixture learning problem, we define its effective dimension as follows. 
\begin{definition}\label{defn:dim_eff}
    Given a Gaussian mixture learning problem instance $\GMM(\kfit, \bThetastar)$, its effective dimension, denoted by $\deff$, is defined as
    \begin{equation}\label{eqn:eff_dim}
        \deff:= \dim \bTheta^* = \dim \spn \left\{ \bthetastar_s: ~ s \in [\ktrue] \right\}.
    \end{equation}
\end{definition}

Note that $\deff \leq d$ by definition. 
It follows from Corollary \ref{cor:span} that if $\B \in \real^{d \times \kfit}$ is a stationary point of $L(\cdot \mid \bTheta^*)$, then $\bbeta_i \in V_{\bThetastar}$ for all $i \in [\kfit]$, where $V_{\bThetastar} := \spn \left\{ \bthetastar_s: ~s \in [\ktrue] \right\}$. 
If $\deff < d$, then we may choose an (ordered) orthonormal basis of $\real^d$ whose first $\deff$ basis vectors span the subspace $V_{\bThetastar}$. 
Under the coordinate system specified by this basis, we can represent $\bthetastar_s = ({\bthetastar}'_s, 0)$ with ${\bthetastar}'_s \in \real^{\deff}$ for all $s \in [\ktrue]$. 
Likewise, for a stationary point $\B \in \real^{d \times \kfit}$ of $L$, we can represent its elements as $\bbeta_i = (\bbeta'_i, 0)$ with $\bbeta'_i \in \real^{\deff}$, for all $i \in [\kfit]$. 

Then we observe that the probability density function of an isotropic Gaussian distribution is coordinate-wisely decomposable.  
That is, for any $\u, \x \in \real^d$ and any $\sigma \in \real_+$, 
\[
    \phi( \x \mid \u, \sigma^2 \Id_d ) = \prod_{i=1}^d \phi( x_1 \mid u_1, \sigma^2 ).
\]
Under the coordinate system specified above,
\begin{align*}
    f^*_s( \x ) 
        &= \phi( \x \mid \bthetastar_s, \sigma^2 \Id_d )\\
        &= \phi( \x_{1:\deff} \mid {\bthetastar}'_s, \sigma^2 \Id_{\deff} ) \cdot \phi( \x_{\deff+1: d} \mid 0, \sigma^2 \Id_{d - \deff} ),\\
    f_i( \x ) 
        &= \phi( \x \mid \bbeta_i, \sigma^2 \Id_d )\\
        &= \phi( \x_{1:\deff} \mid {\bbeta}'_i, \sigma^2 \Id_{\deff} ) \cdot \phi( \x_{\deff+1: d} \mid 0, \sigma^2 \Id_{d - \deff} ),
\end{align*}
for all $s \in [\ktrue]$ and for all $i \in [\kfit]$, respectively. 
Therefore, for all $\X = (\x_1, \dots, \x_{\kfit}) \in V_{\bThetastar}^{\kfit} \cong \real^{\deff \times \kfit}$,
\[
    L(\X \mid \bThetastar ) = L( \X' \mid {\bThetastar}')
\]
where ${\bThetastar}' = ( {\bthetastar}'_1 \dots, {\bthetastar}'_{\ktrue} ) \in \real^{\deff \times \ktrue}$. 
Thus, a local minimum of $L( \cdot \mid \bThetastar )$ is a local minimum of $L( \cdot \mid {\bThetastar}')$ (with a change of coordinates), and vice versa. 
Consequently, we may replace the dimension $d$ in \eqref{eqn:approx_bound.1} of Theorem \ref{thm:master} by the effective dimension $\deff$, which satisfies $\deff \leq \min \{ d, \ktrue\}$. 

\begin{remark}
    If $\deff < d$ and $\B \in \real^{d \times k}$ is a stationary point of $L(\cdot \mid \bThetastar)$, then for every $i \in [\kfit]$, the $i$-th Voronoi cell of $B$ takes the form $\vor_i \cong \vor'_i \times \real^{d - \deff}$ for some $\vor'_i \subseteq V_{\bThetastar}$.
\end{remark}

\paragraph{Approximately optimal choice of parameter $\delta$}
Lastly, we optimize the value of $\delta$ to minimize the approximation error bounds \eqref{eqn:approx_bound.1} and \eqref{eqn:approx_bound.2}. 
If $q_0 = 0$, there does not exists ``one-fits-many.'' Thus, we may simply choose the largest value $\delta = \frac{1}{ 18 (\sqrt{2\pi} + 1 ) \cdot  \kfit^4 }$ to optimize the approximation error bound \eqref{eqn:approx_bound.2}.

If $q_0 \neq 0$, we choose a particular value of $\delta$ that minimizes the upper bound \eqref{eqn:approx_bound.1}. 
Observe, e.g., by the AM-GM inequality that for any $A, B > 0$, $\min_{\delta > 0} \big\{ A \delta + \frac{B}{\delta} \big\} = 2\sqrt{AB}$, which is achieved by $\delta^{\star} \coloneqq \arg\min_{\delta > 0} \big\{ A \delta + \frac{B}{\delta} \big\} = \sqrt{B/A}$. 
With the choice of $A = 18 \left(\sqrt{2\pi} + 1 \right) \cdot \kfit^3 \cdot \left( 5\ktrue + 2 \kfit \right) \cdot \frac{\deltacell(\B)}{\std}$ and 
$B = \frac{4\ktrue}{ | \cT^{\delta}_a | }$, we obtain
\begin{align}
    \delta^{\star} 
        &= \sqrt{\frac{B}{A}}   \nonumber\\
        &= \min_{a \in [q_0]} \Bigg\{ \bigg( \frac{ 2 \ktrue }{ 9 \left(\sqrt{2\pi} + 1 \right) \cdot \kfit^3 \cdot \left( 5\ktrue + 2 \kfit \right) } \bigg)^{\frac{1}{2}}   \cdot \bigg( \frac{1}{|\cT^{\delta}_a|} \cdot \frac{\std}{\deltacell(\B)} \bigg)^{\frac{1}{2}} \Bigg\}.
                \label{eqn:delta_star.0}
\end{align}

Next, we ensure $\delta = \delta^{\star}$ satisfies the requirement $\frac{4 \ktrue \cdot \sigma}{\deltamin} < \delta \leq \frac{1}{ 18 (\sqrt{2\pi} + 1 ) \cdot  \kfit^4 }$ in Theorem \ref{thm:master}. 
Firstly, since $|\cT^{\delta}_a| \leq \ktrue$ for all $a \in [q_0]$, the lower bound $\frac{4 \ktrue \cdot \sigma}{\deltamin} < \delta^{\star}$ is satisfied if 
\begin{align*}
    \frac{4 \ktrue \cdot \sigma}{\deltamin} <  \sqrt{ \frac{ 2 }{ 9 \left(\sqrt{2\pi} + 1 \right) \cdot \kfit^3 \cdot \left( 5\ktrue + 2 \kfit \right) } \cdot \frac{\std}{\deltacell(\B)} },
\end{align*}
or equivalently, if the inequality in \eqref{eqn:SNR_requirement.2} holds:
\begin{align*}
    \frac{\deltacell(\B)}{\std}
    &<   \frac{ 1 }{ 72 \left(\sqrt{2\pi} + 1 \right) \cdot \ktrue^2 \cdot \kfit^3 \cdot \left( 5\ktrue + 2 \kfit \right) } \cdot \left(\frac{\deltamin}{\std} \right)^2.
\end{align*}

Secondly, the upper bound $\delta^{\star} \leq \frac{1}{ 18 (\sqrt{2\pi} + 1 ) \cdot  \kfit^4 }$ is automatically satisfied when $\frac{\deltamin}{\sigma} > 72 (\sqrt{2\pi} + 1 ) \cdot \ktrue \cdot \kfit^4$ as in \eqref{eqn:SNR_requirement}. 
To see this, it suffices to notice that $\deltacell(\B) \geq \frac{1}{2} \deltamin$ when $q_0 \geq 1$. 
Then we observe that 
\begin{align*}
    \delta^{\star}
        &= \left( \frac{ 2 \ktrue }{ 9 \left(\sqrt{2\pi} + 1 \right) \cdot \kfit^3 \cdot \left( 5\ktrue + 2 \kfit \right) } \cdot \frac{1}{|\cT^{\delta}_a|} \cdot \frac{\std}{\deltacell(\B)} \right)^{\frac{1}{2}}
            \\
        &\stackrel{(a)}{\leq} \left( \frac{ 4 \ktrue }{ 9 \left(\sqrt{2\pi} + 1 \right) \cdot \kfit^3 \cdot \left( 5\ktrue + 2 \kfit \right) }\cdot \frac{\std}{\deltamin} \right)^{\frac{1}{2}}
            \\
        &\stackrel{(b)}{\leq} \frac{1}{ 18 \left(\sqrt{2\pi} + 1 \right) \cdot \kfit^4} \cdot \left( \frac{ 2 \kfit }{ 5\ktrue + 2 \kfit } \right)^{\frac{1}{2}}
            \\
        &\leq \frac{1}{ 18 \left(\sqrt{2\pi} + 1 \right) \cdot \kfit^4},
\end{align*}
where (a) follows from $|\cT^{\delta}_a| \geq 1$ and $\deltacell(\B) \geq \frac{1}{2} \deltamin$, and (b) is due to \eqref{eqn:SNR_requirement}.

All in all, if the conditions \eqref{eqn:SNR_requirement} and \eqref{eqn:SNR_requirement.2} are met, then the approximation error bounds \eqref{eqn:approx_bound.1} and \eqref{eqn:approx_bound.2} in Theorem \ref{thm:master} reduce to the following: $\forall a \in [q_0]$ and $\forall a \in [q] \setminus [q_0]$ respectively,
\begin{align*}
    \frac{1}{\sigma} \left\|\, \bbeta_{i_a} - \frac{1}{\left| \cT^{\delta}_a \right| } \sum_{s \in \cT^{\delta}_a} \bthetastar_s \,\right\|
            &\lesssim \left( \frac{\deltacell(\B)}{\std} \cdot \ktrue^2 \cdot \kfit^3 \cdot (\ktrue + \kfit) \right)^{1/2} + \ktrue \cdot d^{1/2}\\
            &\lesssim \frac{\deltamin}{\std} + \ktrue \cdot d^{1/2},
                \\
    \frac{1}{\sigma} \left\| \bbeta_i - \bthetastar_{s_a} \right\|
            &\lesssim \left( \frac{\deltacell(\B)}{\std} \cdot \ktrue^2 \cdot \kfit^3 \cdot (\ktrue + \kfit) \right)^{1/2}\\
            &\lesssim  \frac{\deltamin}{\std}.
\end{align*}
Likewise, the cross-association bounds in \eqref{eqn:weak_association} are bounded from above by
\[
    k^4 \cdot \delta^{\star} \lesssim \left( \frac{\kfit^5 \cdot \ktrue}{\ktrue + \kfit} \cdot \frac{\std}{\deltacell(\B)} \right)^{1/2}
        \lesssim \left( \frac{\kfit^5 \cdot \ktrue}{\ktrue + \kfit} \cdot \frac{\std}{\deltamin} \right)^{1/2}.
\]
Here, the last inequality follows from the observation that if $q_0 \neq 0$, then $\deltacell(\B) \geq \frac{1}{2} \deltamin$.

\subsubsection{Distillation of Partitions from Quasi-partitions}\label{sec:distillation}
We extract partitions of $[\kfit]$ and $[\ktrue]$ by slightly altering the quasi-partitions $\bbS^{\delta}$ and $\bbT^{\delta}$ in Theorem \ref{thm:master}. 

\paragraph{Alteration process to distill decomposable partitions} 
Let $\B$ be a local minimum of $L(\,\cdot \,|\, \bTheta^*)$, and suppose that we are given $q, q_0, \bbS, \bbT$ as stated in Theorem \ref{thm:master}. 
We define the set of `problematic indices' by letting
\begin{equation}\label{set:problematic}
    \setprob \coloneqq \left\{ a\in[q]\setminus[q_{0}]: \setconf_a\neq\emptyset \right\} 
\end{equation}
where
\[
    \setconf_{a} \coloneqq \left\{ b\in[q_{0}]:\cS_{a}\cap\cS_{b}\neq\emptyset\right\}.
\]
We resolve the conflict in $\cS_{b}$ for each of $a \in \setprob$ sequentially with a simple `surgery' (Algorithm \ref{alg:distillation}).

\begin{algorithm}[ht!]
    \caption{Alteration process to distill fully decomposable partitions} 
    \label{alg:distillation}
    \textbf{Input:} $q_{0},q,\bbS,\bbT$ that satisfy the conclusions of Theorem \ref{thm:master}\\
    \textbf{Output:} $q_{0},q,\bbS,\bbT$; these may be different from the inputs
    \begin{algorithmic}[1]
        \FOR {$a \in \setprob$}
            \IF { $\cS_a \setminus \bigcup_{b \in \setconf_a} \cS_b \neq \emptyset$ }
                \STATE $\cS_a \gets \cS_a \setminus \bigcup_{b \in \setconf_a} \cS_b$
                    \hfill\COMMENT{$\bbS$ is also updated accordingly}
            \ELSE
                \STATE $i \gets \arg\min \cS_a$
                \STATE $b_0 \gets$ the unique element $b \in \setconf_a$ such that $\cS_b = \{i\}$
                \STATE $\cT_{b_0} \gets \cT_{b_0} \cup \cT_a$       
                    \hfill\COMMENT{Merge $\cT_a$ to $\cT_{b_0}$}
                \STATE $\bbS \gets \bbS \setminus \{ \cS_a \}$, $\bbT \gets \bbT \setminus \{ \cT_a \}$
                \FOR {$a' \in [q] \setminus [a]$}
                    \STATE $\cS_{a' - 1} \gets \cS_{a'}$, $\cT_{a' - 1} \gets \cT_{a'}$
                        \hfill\COMMENT{Update the group index for $a' > a$}
                \ENDFOR
                \STATE $q \gets q - 1$
            \ENDIF
        \ENDFOR
    \end{algorithmic} 
\end{algorithm}

\paragraph{Properties of the output of Algorithm \ref{alg:distillation}} 
Let $(q, q_0, \bbS, \bbT)$ be any tuple that satisfies the conclusions of Theorem \ref{thm:master}, and let $(q', q_0', \bbS', \bbT') = \Alg(q, q_0, \bbS, \bbT)$ where $\Alg$ is Algorithm \ref{alg:distillation}. 
We observe that $(q', q_0', \bbS', \bbT')$ satisfy the conclusions in Claims 1 through 3 of Theorem \ref{thm:master}, and moreover, $\bbS'$ is a partition of $[\kfit]$. 
This observation is formally stated in the following proposition, whose proof is deferred to Appendix \ref{sec:proof_distillation}.

\begin{proposition}\label{prop:distillation}
    Let $\kfit, \ktrue \in \NN$, $\bThetastar \in \real^{d \times \ktrue}$, and $\B \in \real^{d \times \kfit}$ be a local minimum of $L(\,\cdot \,|\, \bTheta^*)$. 
    Let $q, q_0 \in \NN$, and $\bbS, \bbT$ be collections of subsets of $[\kfit]$, $[\ktrue]$, respectively. 
    If $q, q_0, \bbS, \bbT$ satisfy the conclusions of Theorem \ref{thm:master}, then $(q', q_0', \bbS', \bbT') = \Alg(q, q_0, \bbS, \bbT)$ where $\Alg$ is Algorithm \ref{alg:distillation} has the following properties:
    \begin{enumerate}
        \item 
        Claim 1 in Theorem \ref{thm:master} holds for $(q', q_0', \bbS', \bbT')$. 
        Moreover, $\bbS'$ is a partition of $[\kfit]$.
        \item
        Claim 2 in Theorem \ref{thm:master} holds for $(q', q_0', \bbS', \bbT')$. 
        \item
        Claim 3 in Theorem \ref{thm:master} holds for $(q', q_0', \bbS', \bbT')$. 
    \end{enumerate}
    
\end{proposition}

\subsubsection{Derivation of Theorem \ref{thm:main} from Theorem \ref{thm:master}}\label{sec:reduction_to_main}

\begin{proof}[Proof of Theorem \ref{thm:main}]
    Given $\B$, we choose $\delta = \delta^{\star}$ per \eqref{eqn:delta_star.0}. 
    Then, we let $\tilde{q}, \tilde{q}_0, \tilde{\bbS}, \tilde{\bbT}$ be the outputs of the construction algorithm (Algorithm \ref{alg:partition}), and let $q, q_0, \bbS, \bbT = \Alg\left( \tilde{q}, \tilde{q}_0, \tilde{\bbS}, \tilde{\bbT} \right)$ where $\Alg$ is the distillation algorithm (Algorithm \ref{alg:distillation}). 
    With these, Theorem \ref{thm:main} immediately follows from Theorem \ref{thm:master} and Proposition \ref{prop:distillation}.
\end{proof}

\section{A Fine-grained Analysis for One-dimensional GMM} \label{sec:proof_main_boost}

In this section, we prove Theorem~\ref{thm:boost} (and Corollary~\ref{cor:under}). 
Our proof consists of four steps outlined here:
\begin{enumerate}
    \item 
    \textbf{(Step 1)}
    Enumerate all possible configurations of tuples $(q, q_0, |\cS_0|)$ that satisfy the simple quasi-partition property for local minima; see Claim 1 of Theorem \ref{thm:main}.
    \item
    \textbf{(Step 2)}
    For each $(q, q_0, |\cS_0|)$, identify all possible compositions of $\bbS$ and $\bbT$ that satisfy the near-empty cross-association property established in Claim 4 of Theorem \ref{thm:main}.
    \item
    \textbf{(Step 3)}
    For each composition, establish refined approximation error bounds individually. 
    \item
    \textbf{(Step 4)}
    Regroup the compositions for the convenience of presentation.
\end{enumerate}

In Section \ref{sec:1dim_lemmas}, we state and prove useful technical lemmas. 
Thereafter, in Section \ref{sec:1dim_proof}, we complete the proof of Theorem~\ref{thm:boost} by detailing Steps 1 through 3.
Lastly, in Section \ref{sec:proof_under}, we prove Corollary~\ref{cor:under} following essentially the same, yet simpler, argument in the proof of Theorem~\ref{thm:boost}.

\subsection{Useful Technical Lemmas}\label{sec:1dim_lemmas}
\subsubsection{Additional Notation and Basic Facts about Gaussians}
Let $\phi: \real \to \real$ denote the probability density of the standard Gaussian, i.e., 
\begin{equation}
    \phi(x) \coloneqq \frac{1}{\sqrt{2\pi}} e^{-\frac{x^2}{2}}.
\end{equation}
Let $\Phi$ and $Q$ denote the Gaussian cumulative distribution function and the Gaussian $Q$-function, respectively:
\begin{equation}\label{eqn:Gaussian_Q}
    \Phi(t) \coloneqq \int_{-\infty}^{t} \phi(z) ~\ddup z
    \quad
    \text{and}
    \quad
    Q(t) \coloneqq \int_{t}^{\infty} \phi(z) ~\ddup z.
\end{equation}
It is well known \cite{borjesson1979simple} that 
\begin{equation}\label{eqn:Q_bounds}
    \frac{t}{t^2+1} \cdot \phi(t) 
        \leq Q(t) 
        \leq \frac{1}{t} \cdot \phi(t), \qquad \forall t > 0.
\end{equation}
\begin{lemma}[Gaussian tail bounds]\label{lem:gaussian_tail}
    For any $t\ge0$,
    \begin{align*}
        \frac{1}{t+1}\phi(t) 
            \leq \frac{1}{t+\sqrt{t^{2}+4}} \phi(t)
            \leq
            Q(t)
            \leq \sqrt{2\pi}\cdot\phi(t).
    \end{align*}
\end{lemma}
\begin{proof}[Proof of Lemma \ref{lem:gaussian_tail}]
    The upper bound is standard. The lower bounds can be found in \cite[Formula 7.1.13]{abramowitz1948handbook}.
\end{proof}

Also, it can be verified that for all $t \in \real$,
\begin{equation}\label{eqn:phi_integral}
    \int_{-\infty}^t z \cdot \phi(z) ~\ddup z = -\phi(t)
    \quad\text{and}\quad
    \int_t^{\infty} z \cdot \phi(z) ~\ddup z = \phi(t).
\end{equation}
A more extensive list of Gaussian integrals can be found in \cite{owen1980table} for example.

Moreover, we observe a simple equation, namely, for any $x, y, \delta \in \real$,
\begin{equation}\label{eqn:pdf_ratio}
    \frac{\phi(x + \delta)}{\phi(y+\delta)} \cdot \frac{\phi(y)}{\phi(x)} = e^{\delta (y-x)}.
\end{equation}

Lastly, we make a simple observation that for any $\alpha, x_0 \in \real$,
\begin{equation}\label{eqn:gaussian_shift}
    e^{-\frac{x^2}{2}} \cdot e^{\alpha(x-x_0)} = e^{- \frac{(x-\alpha)^2}{2}} \cdot e^{\frac{\alpha}{2} (\alpha - 2 x_0)}.
\end{equation}
This readily implies that $\phi(x)\cdot e^{\alpha(x-x_0)} = \phi(x-\alpha) \cdot e^{\frac{\alpha}{2} (\alpha - 2 x_0)}$.

\subsubsection{Useful Technical Lemmas}\label{sec:useful_lemmas}

For $\alpha, \beta \in \real$, we define $\psiab: \real \to \real$ be a function such that
\begin{equation}\label{eqn:psiab}
    \psiab(x) = \frac{e^{-(x-\beta)^2/2}}{e^{-(x-\alpha)^2/2} + e^{-(x-\beta)^2/2}}.
\end{equation}

\begin{lemma}\label{lem:separation}
    Let $\alpha, \beta \in \real$ such that $\alpha < 0 < \beta$ and $\beta \geq |\alpha|$. 
    Then
    \begin{align*}
        \E_{\sfx \sim \cN(0,1)} \big[ \psiab(\sfx) \cdot \left( \beta - \sfx \right) \big]
            &\geq \beta \cdot \E_0 \left[ \psiab(\sfx) \cdot \big( 2 \psiab(\sfx) - 1 \big) \right]\\
            &\geq \frac{\beta}{8} \cdot Q \left( \frac{\alpha + \beta}{2} - \frac{1}{\alpha} \right).
    \end{align*}
\end{lemma}

\begin{remark}
    We note that it is possible to obtain a tighter lower bound by refining the proof of Lemma \ref{lem:separation}. 
    Specifically, we can derive a tighter lower bound for $\E_0 \left[ \psiab(\sfx) \cdot \big( 2 \psiab(\sfx) - 1 \big) \right]$ 
    by continuing from the expression in \eqref{eqn:lem_separation_int} as follows\footnote{See \eqref{eqn:g_function} and \eqref{eqn:lem_separation_int} in Appendix \ref{sec:proof_lemma.tech1} (the proof of Lemma \ref{lem:separation}) for the definition of the functions $g, h_1, h_2$ and more details.}:
    \begin{align}
        \E_0 \left[ \psiab(\sfx) \cdot \big( 2\psiab(\sfx) - 1 \big) \right] 
            &\geq \frac{1}{\sqrt{2\pi}} \int_{1/\delta}^{\infty} g(z + c) \cdot \big[ h_1(z) - h_2(z) \big] ~\ddup z   \nonumber\\
            &\geq \frac{1}{\sqrt{2\pi}} \int_{1/\delta}^{\infty} \frac{1}{4} e^{(z-\delta)^2/2} \cdot \big[ 1 - e^{2\alpha z} \big] \cdot h_1(z) ~\ddup z   \nonumber\\
            &\geq \frac{1}{4 \sqrt{2\pi}} \int_{1/\delta}^{\infty} e^{-(z-c)^2/2} ~\ddup z  
             - \frac{1}{4 \sqrt{2\pi}} \cdot \frac{\phi(c)}{\phi(c+2\alpha)} \int_{1/\delta}^{\infty} e^{-(z-c-2\alpha)^2/2} ~\ddup z   \nonumber\\
            &= \frac{1}{4} \left[ Q\left( \frac{1}{\delta} - c \right) -  \frac{\phi(c)}{\phi(c+2\alpha)} \cdot Q\left( \frac{1}{\delta} - c - 2\alpha\right) \right].  \label{eqn:lem_separation_alt}
    \end{align}
\end{remark}

\begin{lemma}[Variance lower bound]\label{lem:var_bound}
    Let $\alpha, \beta \in \real$ such that $\alpha \neq \beta$, and let $\psiab$ be the function as defined in \eqref{eqn:psiab}. 
    Then
    \begin{align*}
        \var_{\sfx \sim \cN(0,1)} \big( \psiab(\sfx) \big)
            &\geq \frac{e^{-4 \left( \frac{|\alpha+\beta|}{2} + |\beta-\alpha| \right)^2}}{32 \sqrt{2\pi}} (\alpha - \beta)^2   
                \cdot \left( \frac{|\alpha+\beta|}{2} + | \beta - \alpha | \right)^3.
    \end{align*}
\end{lemma}

\begin{lemma}[Exponential association]\label{lem:Psi_domination}
    Let $\sigma = 1$, $d=1$, $s \in [\ktrue]$ and $i_s \coloneqq \arg\min_{i \in [\kfit]} |\bbeta_i - \bthetastar_s|$. 
    For $j \in [\kfit]\setminus \{i_s\}$, 
    \begin{equation}
        \E_{s}\left[\Asso_{j}\right] 
            \leq \left( 1 + \frac{4}{\sqrt{2\pi} \cdot \delta_j } \right) e^{- \frac{\delta_j^2}{32}}.
    \end{equation}
    where $\delta_j \coloneqq |\bbeta_j - \bthetastar_s| - |\bbeta_{i_s} - \bthetastar_s|$.
\end{lemma}

\begin{lemma}[Exponential accuracy]\label{lem:accu_domination}
    Let $\sigma = 1$ and $d=1$. Suppose that $\B$ be a stationary point of $L$. 
    Let $i \in [\kfit]$, $\cT  \subseteq [\ktrue]$, and $\delta \in \real_+$ such that $\delta \geq \max \left\{ \frac{4}{\sqrt{2\pi}}, \, 8 \sqrt{\log ( 2 \sqrt{2} \cdot \kfit \ktrue)} \right\}$. 
    If 
    \begin{equation}\label{eqn:accu_premise}
    \begin{aligned}
        \min_{j \in [\kfit]\setminus\{i\}} \| \bbeta_j - \bthetastar_s \| - \| \bbeta_i - \bthetastar_s \| &\geq \delta, \qquad \forall s \in \cT,\\
        \| \bbeta_i - \bthetastar_s \| - \min_{j \in [\kfit]\setminus\{i\}} \| \bbeta_j - \bthetastar_s \| &\geq \delta, \qquad \forall s \in [\ktrue]\setminus \cT,
    \end{aligned}
    \end{equation}
    then
    \begin{align*}
        \left\| \bbeta_i - \frac{1}{|\cT|} \sum_{s \in \cT} \bthetastar_s \right\| 
            \leq 4 \kfit \ktrue \deltamax \cdot e^{- \frac{\delta^2}{64}}.
    \end{align*}
\end{lemma}

\subsection{Completing the Proof of Theorem~\ref{thm:boost}}\label{sec:1dim_proof}
This entire subsection is dedicated to the proof of Theorem~\ref{thm:boost}, which follows the four-step strategy outlined in the preamble of Section \ref{sec:proof_main_boost}. 
Here we begin by making several preliminary observations. 
Thereafter, we prove Theorem~\ref{thm:boost} by detailing Step 1 (Section \ref{sec:proof_step1}), Step 2 (Section \ref{sec:proof_step2}), and Step 3 (Section \ref{sec:proof_step3}) of the outlined argument. 

Note that we may assume $\sigma = 1$ by treating $\Delta/\sigma$ as new $\Delta$; thus, in the rest of the proof, we assume $\sigma = 1$ and write $\Delta$ in place of $\Delta/\sigma$. 
If $\Delta > 2^{16} \cdot 3^{10} \cdot  (\sqrt{2\pi} + 1)$, then the approximation error bounds in Claim 3 of Theorem \ref{thm:main}, cf. \eqref{eqn:approx_bound.main.2}, reduce to the following.
\begin{itemize}
    \item
    For $a \in [q_0]$,
    \begin{equation}\label{eqn:approx_bound.1dim.a}
    \begin{aligned}
        \left\|\, \bbeta_{i_a} - \frac{1}{\left| \cT_a \right| } \sum_{s \in \cT_a} \bthetastar_s \,\right\|
            &\leq 
                \left(2^8 \cdot 3^{8} \cdot (\sqrt{2\pi} + 1) \cdot \Delta \right)^{1/2} + 2 \\
            &< \frac{\Delta}{32}
    \end{aligned}
    \end{equation}
    where $i_a$ denotes the unique element in $\cS_a$.
                
    \item
    For $a \in [q] \setminus [q_0]$, and for all $i \in \cS_a$,
    \begin{equation}\label{eqn:approx_bound.1dim.b}
    \begin{aligned}
        \left\| \bbeta_i - \bthetastar_{s_a} \right\|
            &\leq 
                \left( 2^4 \cdot 3^8 \cdot  (\sqrt{2\pi} + 1) \cdot \Delta \right)^{1/2}\\
            &< \frac{\Delta}{192},
    \end{aligned}
    \end{equation}
    where $s_a$ denotes the unique element in $\cT_a$.
\end{itemize}
Moreover, the cross-association bound in Claim 4 of Theorem \ref{thm:main}, cf. \eqref{eqn:weak_association.main}, reduces to the following. 
\begin{quote}
    Let $(a,b) \in [q] \times [q]_0$ such that $a \neq b$. 
        If $s \in \cT_a$ and $i \in \cS_b$, then
        \begin{equation}\label{eqn:weak_association.1dim}
            \E_s \left[ \Asso_i \right] 
                < \frac{1}{2^9 \cdot 3^2}
            \quad\text{and}\quad
            \P_s \left( \vor_i \right)  
                < \frac{1}{2^9 \cdot 3^3}.
        \end{equation}
\end{quote}

\subsubsection{Step 1. Preliminary Screening}\label{sec:proof_step1}
First of all, we enumerate all possible configurations of parameter tuples $(q, q_0, |\cS_0|)$ that are allowed by the quasi-partition property stated in Theorem \ref{thm:master}. 
To this end, we first observe that (i) $1 \leq q \leq \min\{\kfit, \ktrue \} = 3$, (ii) $0 \leq q_0 \leq q$, and (3) $0 \leq |\cS_0| \leq \kfit = 3$. 
Next, we exclude some of these parameter combinations based on the simple quasi-partition property of the collections $\bbS$ and $\bbT$, cf. claim 1 of Theorem \ref{thm:master}.

\begin{itemize}
    \item 
    Suppose that $(q,q_0) = (1,0)$. 
    This is not allowed as $|\cT_1| = 1$ contradicts $\cT_1 = \bigcup_{a=1}^q \cT_a = [\ktrue] = \{1, 2, 3\}$.
    
    \item
    Suppose that $(q,q_0) = (1,1)$. 
    Then $|\cS_1 | = 1$. 
    Thus, $|\cS_0| = 2$ because $\cS_0 \cap \cS_1 = \emptyset$ and $\cS_0 \cup \cS_1 = [\kfit]$.

    \item
    Suppose that $(q,q_0) = (2,0)$. 
    This is not allowed as $|\cT_1| = |\cT_2| = 1$ contradicts $\cT_1 \cup \cT_2 = \{1, 2, 3\}$.
    
    \item
    Suppose that $(q,q_0) = (2,1)$. 
    Then $|\cS_1| = 1$ and $|\cS_2| \geq 2$. 
    Observe that $\cS_0 \cap \cS_2 = \emptyset$ and $\bigcup_{a=0}^2 \cS_a = [\kfit]$. 
    Thus, $|\cS_0| \leq 1$.

    \item
    Suppose that $(q,q_0) = (2,2)$. 
    Then $|\cS_1 | = |\cS_2| = 1$. 
    Observe that $\cS_a \cap \cS_b = \emptyset$ for all $a,b \in \{0, 1, 2\}$ such that $a \neq b$, and that $\bigcup_{a=0}^2 \cS_a = [\kfit] = \{1, 2, 3\}$. 
    Therefore, $|\cS_0| = 1$.

    \item
    Suppose that $q = 3$ and $q_0 \in \{0, 1\}$. 
    Then $|\cS_2 | \geq 2$, $\cS_3| \geq 2$, and $\cS_2 \cap \cS_3 = \emptyset$. 
    This contradicts $\bigcup_{a=0}^3 \cS_a = [\kfit]$ because $|\cS_2 \cup \cS_3| \geq 4$. 
    Thus, these configurations are forbidden.

    \item
    Suppose that $(q,q_0) = (3,2)$. 
    Then $|\cS_1|=|\cS_2| = 1$ and $\cS_1 \cap \cS_2 = \emptyset$. 
    Since $\cS_0 \cap \cS_a = \emptyset$ for $a \in \{1, 2\}$ and $\bigcup_{a=0}^3 \cS_a = [\kfit] = \{1, 2, 3\}$, it must hold that $|\cS_0| \leq 1$. 

    \item
    Suppose that $(q,q_0) = (3,3)$. 
    Then $|\cS_1|=|\cS_2|=|\cS_3| = 1$ and $\cS_1, \cS_2, \cS_3$ are mutually disjoint. 
    Since $\cS_0 \cap \cS_a = \emptyset$ for $a \in \{1, 2, 3\}$ and $\bigcup_{a=0}^3 \cS_a = [\kfit] = \{1, 2, 3\}$, it must hold that $|\cS_0| = 0$. 
\end{itemize}

\begin{table}[h!]
    \centering
    \caption{All possible configurations of parameter tuples $(q, q_0, |\cS_0|)$ for the collections $\bbS, \bbT$ that satisfy the simple quasi-partition property stated in Theorem \ref{thm:master}.}
    \label{tab:possible_configs}
    \begin{tabular}{c c c c}
        \toprule
        ~\quad$q$\quad~ &   ~\quad$q_0$\quad~   &   ~\quad$|\cS_0|$\quad~   &     \qquad\qquad\qquad~  \\
        \midrule
        1   &   0       &   0, 1, 2 or 3    &   impossible\\
            \cline{2-4}
            &   1       &   0 or 1          &   impossible\\
            &           &   2               &   -   \\
            &           &   3               &   impossible\\
        \midrule
        2   &   0       &   0, 1, 2 or 3    &   impossible\\
            \cline{2-4}
            &   1       &   0 or 1          &   -\\
            &           &   2 or 3          &   impossible\\
            \cline{2-4}
            &   2       &   0               &   impossible\\
            &           &   1               &   -   \\
            &           &   2 or 3          &   impossible\\
        \midrule
         3   &   0 or 1 &   0, 1, 2 or 3    &   impossible\\
            \cline{2-4}
            &   2       &   0 or 1          &   -   \\
            &           &   2 or 3          &   impossible\\
            \cline{2-4}
            &   3       &   0               &   -   \\
            &           &   1, 2 or 3       &   impossible\\
        \bottomrule
    \end{tabular}
\end{table}

These are summarized in Table \ref{tab:possible_configs}. 
In the next step, we consider each of these configurations $(q, q_0, |\cS_0|)$ and identify all possible compositions of $\bbS$ and $\bbT$.

\subsubsection{Step 2. Identification of All Possible Compositions of \texorpdfstring{$\bbS$}{S} and \texorpdfstring{$\bbT$}{T} for Local Minima}\label{sec:proof_step2}
Next, we list all possible compositions of $\bbS$ and $\bbT$ for the remaining configurations of $(q, q_0, |\cS_0|)$. 
Note that we may freely relabel the indices of estimates $\{ \bbeta_i: i \in [k] \}$ by a permutation of $[\kfit]$. 
Also, we can relabel the indices of the true component means $\{ \bthetastar_s: s \in [\ktrue] \}$ by flipping the order: $(1,2,3) \mapsto (3,2,1)$. 
Thus, the possible compositions of $\bbS$ and $\bbT$ are presented up to permutation of $[k]$ and the flipping of the order in $[\ktrue]$.
\begin{enumerate}
    \item 
    $(q, q_0, |\cS_0|) = (1,1,2)$. 
    The only possibility is $\cS_0 = \{2, 3\}$; $\cS_1 = \{ 1 \}$; $\cT_1 = \{1, 2, 3 \}$.
    \item 
    $(q, q_0, |\cS_0|) = (2,1,0)$. 
    There are two possibilities: (a) $|\cS_1| = 1$ and $|\cS_2| = 2$; or (b) $|\cS_1| = 1$ and $|\cS_2| = 3$. In both cases, $\cS_0 = \emptyset$.
    \begin{enumerate}[label=(\alph*)]
        \item 
        $\cS_1 = \{1\}$, $\cS_2 = \{2, 3\}$.
        \begin{enumerate}
            \item 
            $\cT_1 = \{1, 2\}$, $\cT_2 = \{3\}$.
            \item
            $\cT_1 = \{1, 3\}$, $\cT_2 = \{2\}$.
        \end{enumerate}
        \item
        $\cS_1 = \{1\}$, $\cS_2 = \{1, 2, 3\}$.
        \begin{enumerate}
            \item 
            $\cT_1 = \{1, 2\}$, $\cT_2 = \{3\}$.
            \item
            $\cT_1 = \{1, 3\}$, $\cT_2 = \{2\}$.
        \end{enumerate}
    \end{enumerate}
    \item 
    $(q, q_0, |\cS_0|) = (2,1,1)$. 
    There are two possibilities: (a) $|\cS_1| = 1$ and $|\cS_2| = 1$; or (b) $|\cS_1| = 1$ and $|\cS_2| = 2$. However, (a) is forbidden by the requirement $|\cS_a| \geq 2$ for $a \in [q]\setminus[q_0]$. Thus, we may write $\cS_0 = \{ 3 \}$, $\cS_1 = \{1\}$, $\cS_2 = \{1, 2\}$.
    \begin{enumerate}[label=(\alph*)]
        \item 
        $\cT_1 = \{1, 2\}$, $\cT_2 = \{3\}$.
        \item
        $\cT_1 = \{1, 3\}$, $\cT_2 = \{2\}$.
    \end{enumerate}
    \item 
    $(q, q_0, |\cS_0|) = (2,2,1)$. 
    We may write $\cS_0 = \{3\}$; $\cS_1 = \{1\}$, $\cS_2 = \{2\}$.
    \begin{enumerate}[label=(\alph*)]
        \item 
        $\cT_1 = \{1, 2\}$, $\cT_2 = \{3\}$.
        \item
        $\cT_1 = \{1, 3\}$, $\cT_2 = \{2\}$.
    \end{enumerate}
    \item 
    $(q, q_0, |\cS_0|) = (3,2,0)$. 
    $\cS_0 = \emptyset$. There are two possibilities: (a) $|\cS_3|=2$; or (b) $|\cS_3| = 3$.
    \begin{enumerate}[label=(\alph*)]
        \item
        $\cS_1 = \{1\}$, $\cS_2 = \{2\}$, $\cS_3 = \{1, 3\}$.
        \begin{enumerate}
            \item 
            $\cT_1 = \{1\}$, $\cT_2 = \{2\}$, $\cT_3 = \{3\}$.
            \item
            $\cT_1 = \{1\}$, $\cT_2 = \{3\}$, $\cT_3 = \{2\}$.
            \item
            $\cT_1 = \{2\}$, $\cT_2 = \{1\}$, $\cT_3 = \{3\}$.
        \end{enumerate}
        \item
        $\cS_1 = \{1\}$, $\cS_2 = \{2\}$, $\cS_3 = \{1, 2, 3\}$; 
        \begin{enumerate}
            \item 
            $\cT_1 = \{1\}$, $\cT_2 = \{2\}$, $\cT_3 = \{3\}$.
            \item
            $\cT_1 = \{1\}$, $\cT_2 = \{3\}$, $\cT_3 = \{2\}$.
        \end{enumerate}
    \end{enumerate}
    \item 
    $(q, q_0, |\cS_0|) = (3,2,1)$.
    $\cS_0 = \{3\}$; $\cS_1 = \{1\}$, $\cS_2 = \{2\}$, $\cS_3 = \{1, 2\}$.
    \begin{enumerate}[label=(\alph*)]
        \item 
        $\cT_1 = \{1\}$, $\cT_2 = \{2\}$, $\cT_3 = \{3\}$.
        \item
        $\cT_1 = \{1\}$, $\cT_2 = \{3\}$, $\cT_3 = \{2\}$.
    \end{enumerate}
    \item 
    $(q, q_0, |\cS_0|) = (3,3,0)$. 
    The only possibility is $\cS_0 = \emptyset$; $\cS_1 = \{ 1 \}$, $\cS_2 = \{2\}$, $\cS_3 = \{3\}$; $\cT_1 = \{1\}$, $\cT_2 = \{2\}$, $\cT_3 = \{3\}$.
\end{enumerate}

Now we rule out some of these compositions via geometric considerations.
\begin{itemize}
    \item 
    \textbf{The compositions with $\cS_1 = \{1\}$ and $\cT_1 = \{1, 3\}$.} 
    We claim that it is impossible for a single estimate $\bbeta_1$ to fit two non-adjacent centers $\{\bthetastar_1, \bthetastar_3\}$. 
    These include the compositions 2-(a)-ii, 2-(b)-ii, 3-(b) and 4-(b). 
    Without loss of generality, we may assume $\bbeta_1 \leq \bbeta_2$; also, we may assume $\bbeta_2 \leq \bbeta_3$ in 2-(a)-ii, 2-(b)-ii.
    Moreover, $\bbeta_1 \neq \bbeta_2$ by Theorem \ref{thm:main}, Claim 2, and thus, $\bbeta_1 < \bbeta_2$. 
    Let $b = \frac{\bbeta_1 + \bbeta_2}{2}$. Then $b > \bthetastar_3$ because $\vor_1$ is a convex set and $\iota_{\B}(\bthetastar_3) = 1$ (which implies $\bthetastar_3 \in \vor_1$; see Definition \ref{defn:voronoi} for the definition of index function $\iota_{\B}$) by construction; recall Algorithm \ref{alg:partition} and the definition of $\setA_i^{\delta}$ in \eqref{eq:setA}. 
    Thus, $\vor_2 \subseteq [b, \infty) \subseteq (\bthetastar_3, \infty)$. 
    Therefore, $\P_2(\vor_2) \leq \P_2 \left((\bthetastar_3, \infty) \right) = Q(\Delta)$ where $Q$ is the Gaussian $Q$-function, cf. \eqref{eqn:Gaussian_Q}. 
    Using the well-known upper bound for the $Q$-function, cf. \eqref{eqn:Q_bounds}, it follows that $\P_2(\vor_2) \leq Q(\Delta) \leq \frac{1}{\Delta} \frac{1}{\sqrt{2\pi}} e^{-\frac{\Delta^2}{2}} < \frac{1}{4}$. 
    Observe that $\P_2(\vor_1) < \frac{1}{2^9 \cdot 3^3 } \leq \frac{1}{4}$ by \eqref{eqn:weak_association.1dim}, which follows from Theorem 2, Claim 4. 
    Also, we can see that $\P_2(\vor_3) \leq \frac{1}{4}$, either by using $\bbeta_2 \leq \bbeta_3$ (when $3 \in \cS_2$), or by using \eqref{eqn:weak_association.1dim} (when $3 \in \cS_0$). 
    Then we have 
    \[
        1 = \P_2( \real ) \leq \sum_{i=1}^3 \P_2(\vor_i) < \frac{3}{4},
    \]
    which is a contradiction. Therefore, these compositions are forbidden.

    \item 
    \textbf{The compositions with overlapping $\cS_a$'s, i.e., $\cS_1 \cap \cS_3 = \{1\}$.} 
    We claim that when $\Delta$ is sufficiently large, it is impossible for two groups of estimates, namely, $\cS_1$ and $\cS_a$ ($a = 2$ or $3$), to share a common estimate $\bbeta_1$. 
    These include the conpositions 2-(b) and 3 ($a = 2$ and $q_0 = 1$) as well as cases 5 and 6 ($a = 3$ and $q_0 = 2$) with all of their subcases.  
    Note that $1 \in \cS_1 \cap \cS_a$.  
    Let $\bthetastar_s$ be the unique element of $\cT_1$ and choose an arbitrary $\bthetastar_t \in \cT_a $. 
    Then we observe that $| \bbeta_1 - \bthetastar_s | < \Delta / 32$ by \eqref{eqn:approx_bound.1dim.a} and $| \bbeta_1 - \bthetastar_t | < \Delta / 192$ by \eqref{eqn:approx_bound.1dim.b}. 
    This yields $\Delta \leq | \bthetastar_s - \bthetastar_t | \leq | \bbeta_1 - \bthetastar_s | + | \bbeta_1 - \bthetastar_t | < 7\Delta / 192$, which is a contradiction.
    Consequently, these compositions are excluded from our consideration.
\end{itemize}
\begin{table*}[t]
    \centering
    \caption{All possible compositions (up to permutation symmetry) of the collections $\bbS, \bbT$ that are remaining after Step 2 of the proof of Theorem~\ref{thm:boost}.}
    \label{tab:possible_compositions}
    \begin{tabular}{l c | c c c c | c c c}
        \toprule
        Case number     & $(q, q_0, |\cS_0|)$ &   \,~~$\cS_0$~~\,   &   \,~~$\cS_1$~~\,   &     \,~~$\cS_2$~~\,  &     \,~~$\cS_3$~~\,
            &   \,~~~$\cT_1$~~~\,   &     \,~~~$\cT_2$~~~\,  &     \,~~~$\cT_3$~~~\\
        \midrule
        Case A / comp 1         &   $(1,1,2)$       &   $\{2,3\}$   &   $\{1\}$     & -   & -   
            &   $\{1,2,3\}$ & - & -   \\
        Case B / comp 2-(a)-i   &   $(2,1,0)$       &   $\emptyset$ &   $\{1\}$     & $\{2,3\}$     & -   
            &   $\{1,2\}$   & $\{3\}$   & -   \\
        Case C / comp 4-(a)      &   $(2,2,1)$       &   $\{3\}$     &   $\{1\}$     & $\{2\}$       & -  
            &   $\{1,2\}$   & $\{3\}$   & -   \\
        Case D / comp 7          &   $(3,3,0)$       &   $\emptyset$ &   $\{1\}$     & $\{2\}$       & $\{3\}$
            &   $\{1\}$ &   $\{2\}$     &   $\{3\}$\\
        \bottomrule
    \end{tabular}
\end{table*}

In the end, we are left with only four cases as summarized in Table \ref{tab:possible_compositions}. 
In Step 3, we investigate each of these four cases and establish more refined approximation error bounds.

\subsubsection{Step 3. Refined Approximation Error Analysis for the Four Remaining Cases}\label{sec:proof_step3}
In this step, we analyze the four remaining cases (Case A through Case D listed in Table \ref{tab:possible_compositions}) one by one.

\paragraph{Step 3 - Case A} 
We draw a contradiction to rule argue that Case A is impossible to happen. 

\begin{figure}[h]
    \begin{centering}
    \includegraphics[width=0.6\linewidth]{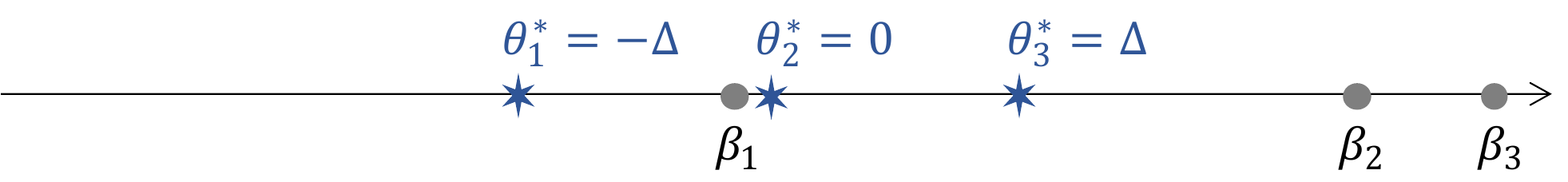}
    \par\end{centering}
    \caption{Illustration of Case A in Step 3.}
    \label{figure:caseA}
\end{figure}

To this end, we begin by observing that $| \bbeta_1 - \bthetastar_s | < | \bbeta_i - \bthetastar_s |$ for all $i \in \{2,3\}$ and all $s \in [\ktrue]$. 
If we assume otherwise, there exists $(i,s) \in \{2,3\} \times [\ktrue]$ such that $| \bbeta_i - \bthetastar_s | \leq | \bbeta_1 - \bthetastar_s |$, which implies $\P_s[\vor_1] \leq 1/2$. Then it follows that $\max_{i \in \{2,3\}}\{ \P_s[\vor_i] \} \geq 1/4$, which contradicts \eqref{eqn:weak_association.1dim}.
Note that $\bthetastar_2 = \frac{1}{3} \sum_{s \in [3]} \bthetastar_s = \bzero$, and thus, $|\bbeta_1| < \Delta / 32$, cf. \eqref{eqn:approx_bound.1dim.a}. 
Thus, $\bbeta_2, \bbeta_3 \notin [ \bthetastar_1 - 31\Delta/32, \bthetastar_3 + 31\Delta/32 ]$. 
In what follows, we assume that $\bbeta_2$ and $\bbeta_3$ are on the same side of this interval, namely, $\bthetastar_3 < \bbeta_2 \leq \bbeta_3$. 
This is illustrated in Figure \ref{figure:caseA}.
However, the opposite-side case, i.e., the case where $\bbeta_2 < \bthetastar_1 < \bthetastar_3 < \bbeta_3$, can be analyzed in a similar manner with a minor modification to the argument that follows.

From the first-order stationarity condition, cf. \eqref{eq:station_cond}, it follows that $\bbeta_{i}=\frac{\E_{*}\left[\Asso_{i}\sf x\right]}{\E_{*}\left[\Asso_{i}\right]}, ~\forall i\in[\kfit]$, and therefore,
\begin{equation}\label{eqn:caseA_stationary}
    0 = \sum_{s \in [\ktrue]} \E_s\left[\Asso_{i} \cdot ( \bbeta_i - \sf x ) \right], \qquad\forall i \in [\kfit].
\end{equation}
Summing up the equations \eqref{eqn:caseA_stationary} for $i \in \{2, 3\}$ yields
\begin{align}
    0 
        &= \sum_{i \in \{2,3\}} \E_1\left[\Asso_{i} \cdot ( \bbeta_i - \sf x ) \right]  
            + \sum_{s \in \{2,3\}} \left( \E_s\left[\Asso_{2} \cdot ( \bbeta_2 - \sf x ) \right] + \E_s\left[\Asso_{3} \cdot ( \bbeta_3 - \sf x ) \right] \right)   \nonumber\\
        &\geq \sum_{i \in \{2,3\}} \E_1\left[\Asso_{i} \cdot ( \bbeta_i - \sf x ) \right] 
            + \sum_{s \in \{2,3\}} \E_s\left[ \big(\Asso_{2} + \Asso_{3}\big) \cdot ( \bbeta_2 - \sf x ) \right],   \label{eqn:caseA.eq1}
\end{align}
where the last inequality follows from $\bbeta_2 \leq \bbeta_3$. 
Rewriting the first term in \eqref{eqn:caseA.eq1} using the Stein's identity (Lemma \ref{lem:stein}), cf. \eqref{eq:PsiX_after_stein}, we obtain that
\begin{align*}
    \sum_{i \in \{2,3\}} \E_1\left[\Asso_{i} \cdot ( \bbeta_i - \sf x ) \right]
        &= \sum_{i \in \{2,3\}}  \left( -\bthetastar_1 \cdot \E_1\left[\Asso_{i} \right] + \sum_{j \in [\kfit]} \bbeta_j \cdot \E_1\left[ \Asso_{i} \Asso_{j} \right] \right)   \\
        &\stackrel{(a)}{=} \sum_{i \in \{2,3\}}  \sum_{j \in [\kfit]} \big( \bbeta_j - \bthetastar_1 \big) \cdot \E_1\left[ \Asso_{i} \Asso_{j} \right] \\
        &\stackrel{(b)}{\geq} 0,
\end{align*}
because (a) $\sum_{j \in [k]}\Asso_j = 1$ with probability $1$ and (b) $\bbeta_j > \bthetastar_1, ~\forall j \in [\kfit]$. 
Therefore, it follows from \eqref{eqn:caseA.eq1} that
\begin{align}
    0
        &\geq \sum_{s \in \{2,3\}} \E_s\left[ \big(\Asso_{2} + \Asso_{3}\big) \cdot ( \bbeta_2 - \sf x ) \right]    \nonumber\\
        &= \sum_{s \in \{2,3\}} \E_s\left[ \big( 1 - \Asso_{1} \big) \cdot ( \bbeta_2 - \sf x ) \right].    \label{eqn:caseA_master_ineq}
\end{align}
In the rest, we will argue that $\E_s\left[ \big( 1 - \Asso_{1} \big) \cdot ( \bbeta_2 - \sf x ) \right] > 0$ for $s \in \{2, 3\}$ to draw a contradiction.

We define a function $\assop: \real \to \real$ so that 
\[
    \assop(x) = \frac{e^{-\|x - \bbeta_2\|^2/2}}{e^{-\|x - \bbeta_1\|^2/2} + e^{-\|x - \bbeta_2\|^2/2}},
\]
and define a random variable $\Assop = \assop(\sfx)$. Note that $\assop(x)$ is equal to the association coefficient of $\bbeta_2$ at $x$ if there were only $\bbeta_1$ and $\bbeta_2$ (with $\bbeta_3$ removed). 
We use $\Assop$ as a proxy of $1 - \Asso_1$ to facilitate our subsequent analysis of Case A. 
Specifically, we observe that for each $s \in \{2, 3\}$,
\begin{equation}\label{eqn:exp_discr}
\begin{aligned}
    \E_s\left[ \big( 1 - \Asso_{1} \big) \cdot ( \bbeta_2 - \sf x ) \right]
        &= \E_s\left[ \Assop \cdot ( \bbeta_2 - \sf x ) \right]
        + \E_s\left[ \big( 1 - \Asso_1 - \Assop \big) \cdot ( \bbeta_2 - \sf x ) \right].
\end{aligned}
\end{equation}
Then we can easily verify that
\begin{equation}\label{eqn:term.prox}
\begin{aligned}
    \E_s\left[ \Assop \cdot ( \bbeta_2 - \sf x ) \right]
        &\geq \frac{ \bbeta_2 - \bthetastar_s }{ 8 } \cdot Q \left( \frac{\bbeta_1 + \bbeta_2}{2} - \bthetastar_s - \frac{1}{\bbeta_1 - \bthetastar_s} \right)
\end{aligned}
\end{equation}
by Lemma \ref{lem:separation}. 
Then we also observe that
\begin{align}
    \E_s\left[ \big( 1 - \Asso_1 - \Assop \big) \cdot ( \bbeta_2 - \sf x ) \right] \nonumber
        &\stackrel{(a)}{\geq} \E_s\left[ \big( 1 - \Asso_1 - \Assop \big) \cdot ( \bbeta_2 - \sf x ) \cdot \indic\{ \sfx \geq \bbeta_2 \} \right]   
            \nonumber\\
        &\stackrel{(b)}{\geq} - \E_s\left[ ( \sf x - \bbeta_2 ) \cdot \indic\{ \sfx \geq \bbeta_2 \} \right]   
            \nonumber\\
        &\stackrel{(c)}{\geq} - \E_s\left[ ( \sf x - \bthetastar_s ) \cdot \indic\{ \sfx \geq \bbeta_2 \} \right]   
            \nonumber\\
        &= - \int_{\bbeta_2 - \bthetastar_s}^{\infty} z \cdot \phi(z) ~\ddup z
            \nonumber\\
        &\stackrel{(d)}{=} - \phi( \bbeta_2 - \bthetastar_s )    \label{eqn:term.prox_rest}
\end{align}
because (a) $1 - \Asso_1 - \Assop \geq 0$; (b) $\big| 1 - \Asso_1 - \Assop \big| \leq 1$; (c) $\bbeta_2 > \bthetastar_s$; and (d) follows from \eqref{eqn:phi_integral}.

Letting $\tau_s :=  \frac{\bbeta_1 + \bbeta_2}{2} - \bthetastar_s - \frac{1}{\bbeta_1 - \bthetastar_s}$ for $s \in \{2,3\}$, we observe that $\tau_s \geq 0$, and thus,
\begin{equation}\label{eqn:Q_lower_bound}
\begin{aligned}
    Q(\tau_s) 
        &\stackrel{(a)}{\geq} \frac{1}{\tau_s + 1} \phi(\tau_s) \stackrel{(b)}{\geq} e^{- \tau_s} \cdot \phi(\tau_s)\\
        &\geq \phi(\tau_s + 1),
\end{aligned}
\end{equation}
where (a) follows from Lemma \ref{lem:gaussian_tail}, (b) is due to $e^t \geq t+1$, $\forall t \geq 0$, and (c) is from the observation $t^2/2 + t \leq (t+1)^2$, $\forall t \in \real$. 
Collecting \eqref{eqn:exp_discr}, \eqref{eqn:term.prox}, \eqref{eqn:term.prox_rest}, and \eqref{eqn:Q_lower_bound} together with \eqref{eqn:caseA_master_ineq}, we have
\begin{align}
    0 
        &\geq \sum_{s \in \{2,3\}} \E_s\left[ \big( 1 - \Asso_{1} \big) \cdot ( \bbeta_2 - \sf x ) \right]  
            \nonumber\\
        &\geq \sum_{s \in \{2,3\}} \left( \frac{ \bbeta_2 - \bthetastar_s }{ 8 } \cdot \phi(\tau_s + 1) - \phi (\bbeta_2 - \bthetastar_s) \right)
            \nonumber\\
        &\stackrel{(a)}{\geq} \frac{ \bbeta_2 - \bthetastar_3 }{ 8 } \cdot \phi(\tau_3 + 1) - \sum_{s \in \{2,3\}} \phi (\bbeta_2 - \bthetastar_s)
            \nonumber\\
        &\stackrel{(b)}{>} 0    \label{eqn:caseA_contradiction}
\end{align}
where (a) is due to $(\bbeta_2 - \bthetastar_s ) \cdot \phi(\tau_s + 1) > 0$. 
The last inequality (b) follows from the observations that $\phi(t)$ is strictly monotone decreasing for $t \geq 0$ and that
\begin{align*}
    \tau_3 + 1 &= \frac{\bbeta_1 + \bbeta_2}{2} - \bthetastar_3 - \frac{1}{\bbeta_1 - \bthetastar_3} + 1\\ 
        &\leq (\bbeta_2 - \bthetastar_3 ) - \frac{31\Delta}{32} + \frac{32}{31\Delta} + 1\\
        &< \bbeta_2 - \bthetastar_3,\\
    \frac{\bbeta_2 - \bthetastar_3}{8} &> \frac{31\Delta}{256} > 2,
\end{align*}
because $\bbeta_2 - \bthetastar_3 > \bthetastar_3 - \bbeta_1 > \frac{31\Delta}{32}$ and $\Delta$ is assumed to be sufficiently large (to be precise, $\Delta > 2^{16} \cdot 3^{10} \cdot  (\sqrt{2\pi} + 1)$).

The inequality $0 > 0$ in \eqref{eqn:caseA_contradiction} is a contradiction, and we conclude that case A cannot happen.

\paragraph{Step 3 - Case B} 
Recall from \eqref{eqn:approx_bound.1dim.a} and \eqref{eqn:approx_bound.1dim.b} in Step 0 of this proof that 
\begin{align*}
    \left\| \bbeta_1 - \frac{\bthetastar_1 + \bthetastar_2}{2} \right\| &< \frac{\Delta}{32},\\
    \left\| \bbeta_i - \bthetastar_3 \right\| &< \frac{\Delta}{192}, \qquad \forall i \in \{2,3\}.
\end{align*}
This implies that (1) $\| \bbeta_1 - \bthetastar_s \| < \frac{17\Delta}{32}$, $\forall s \in \{1,2\}$; (2) $\left\| \bbeta_1 - \bthetastar_3 \right\| \geq  \left\| \frac{\bthetastar_1 + \bthetastar_2}{2} - \bthetastar_3 \right\| - \left\| \bbeta_1 -\frac{\bthetastar_1 + \bthetastar_2}{2} \right\| > \frac{3\Delta}{2} - \frac{\Delta}{32} = \frac{47\Delta}{32}$; and (3) $\| \bbeta_i - \bthetastar_s \| > \frac{191 \Delta}{192}$ for all $(i,s) \in \{2,3\} \times \{1,2\}$. 

Recall $\cT_1 = \{ 1, 2 \}$ in this case and let $\delta_1 = 89\Delta/192$. 
Observe that $\delta_1 \geq \max \left\{ \frac{4}{\sqrt{2\pi}}, \, 8 \sqrt{\log ( 2 \sqrt{2} \cdot \kfit \ktrue)} \right\}$ and that the conditions in \eqref{eqn:accu_premise} are satisfied for $i=1$ with $\delta = \delta_1$ and $\cT = \cT_1$ because
\begin{equation}\label{eqn:caseB.separation}
\begin{aligned}
    \min_{j \in \{2,3\}} \| \bbeta_j - \bthetastar_s \| - \| \bbeta_1 - \bthetastar_s \| 
        &\geq \frac{191}{192} \Delta - \frac{17}{32} \Delta\\
        &= \frac{89}{192}\Delta, \quad \forall s \in \{1,2\},\\
    \| \bbeta_1 - \bthetastar_3 \| - \min_{j \in \{2,3\}} \| \bbeta_j - \bthetastar_3 \| 
        &\geq \frac{47}{32}\Delta - \frac{1}{192}\Delta \\
        &= \frac{281}{192}\Delta.
\end{aligned}
\end{equation}
Applying Lemma \ref{lem:accu_domination}, we obtain
\begin{equation}\label{eqn:caseB_beta1}
\begin{aligned}
    \left\| \bbeta_1 - \frac{\bthetastar_1 + \bthetastar_2}{2} \right\|
        &\leq 4 \kfit \ktrue \deltamax \cdot e^{- \frac{\delta_1^2}{64}}\\
        &= 8 \kfit \ktrue \Delta \cdot e^{- \frac{7921}{2359296} \Delta^2}\\
        &\leq 8 \kfit \ktrue \Delta \cdot e^{- \frac{1}{298} \Delta^2}.
\end{aligned}
\end{equation}

It remains to establish an upper bound for $\max_{i \in \{2,3\}} \| \bbeta_i - \bthetastar_3 \|$. 
To accomplish this, for each $i \in \{2,3\}$, we define a function $\assop_i: \real \to \real$ so that 
\[
    \assop_i(x) = \frac{e^{-\|x - \bbeta_i\|^2/2}}{e^{-\|x - \bbeta_2\|^2/2} + e^{-\|x - \bbeta_3\|^2/2}},
\]
and define a random variable $\Assop_i = \assop_i(\sfx)$. 
Note that $\assop_i(x)$ is equal to the association coefficient of $\bbeta_i$ at $x$ if there were only $\bbeta_2$ and $\bbeta_3$ (with $\bbeta_1$ removed). 
We use $\Assop_i$ as a proxy of $\Asso_i$ to facilitate our subsequent analysis of Case B.

Next, we make three preparatory observations as follows.
\begin{enumerate}[label = (\alph*)]
    \item 
    First, it follows from the first-order stationarity condition, cf. \eqref{eq:station_cond}, that
    \[
        \sum_{s \in [\ktrue]} \E_s \left[ \Asso_i \cdot (\sfx - \bbeta_i ) \right] = 0, \qquad \forall i\in[\kfit].
    \]
    This leads to the following: for all $i \in [3]$,
    \begin{equation}\label{caseB:eqA}
        \E_3 \left[ \Asso_i \cdot (\sfx - \bbeta_i ) \right] 
            = - \sum_{s \in \{1,2\}}  \E_s \left[ \Asso_i \cdot (\sfx - \bbeta_i ) \right].
    \end{equation}
    \item
    Second, by the Stein's identity (Lemma \ref{lem:stein}), we have the following equation, cf. \eqref{eq:PsiX_after_stein}: for all $i\in \{2,3\}$,
    \begin{align*}
        \E_3 \left[ \Assop_i \cdot (\sfx - \bbeta_i ) \right] 
            &= \bthetastar_s \cdot \E_3 \left[ \Assop_i \right]
            - \sum_{j \in \{2,3\}} \bbeta_j \cdot \E_3 \left[ \Assop_i \Assop_j \right].
    \end{align*}
    Then it follows that
    \begin{align}\label{caseB:eqB}
        &\E_3 \left[ \Assop_3 \right] \cdot \E_3 \left[ \Assop_2 \cdot (\sfx - \bbeta_2 ) \right]
             - \E_3 \left[ \Assop_2 \right] \cdot \E_3 \left[ \Assop_3 \cdot (\sfx - \bbeta_3 ) \right] \nonumber\\
        &= - \sum_{j \in \{2,3\}} \Big\{ \bbeta_j \cdot \E_3 \left[ \Assop_3 \right] \cdot \E_3 \left[ \Assop_2 \Assop_j \right] - \bbeta_j \cdot  \E_3 \left[ \Assop_2 \right] \cdot \E_3 \left[ \Assop_3 \Assop_j \right] \Big\}  \nonumber\\
        &= - \left( \bbeta_2 -\bbeta_3 \right) \cdot \var_3\left[ \Assop_2 \right]
    \end{align}
    where $\var_3 \left[ \sfz \right] := \E_3 [ \sfz^2 ] - \E_3[ \sfz ]^2$ for a random variable $\sfz$.
    \item
    Third, for each $i \in \{2,3\}$, the proxy $\assop_i(x) \approx \asso_i(x)$ in the sense that 
    \begin{align*}
        \E_3 \left[ \left( \Assop_i - \Asso_i \right)^2 \right]
            &= \E_3 \left[ (\Assop_i \cdot \Asso_1 )^2 \right]\\
            &\leq \E_3 \left[ \Asso_1 \right].     
    \end{align*}
    As we observed $\| \bbeta_1 - \bthetastar_3 \| - \min_{i \in [\kfit]} \| \bbeta_i - \bthetastar_3 \| > \frac{281}{192}\Delta$ in \eqref{eqn:caseB.separation}, we apply Lemma \ref{lem:Psi_domination} to obtain 
    \begin{equation}\label{eqn:caseB.prox_approx}
    \begin{aligned}
        \E_3 \left[ \left( \Assop_i - \Asso_i \right)^2 \right]
            &\leq \E_3 \left[ \Asso_1 \right]\\
            &\leq \left( 1 + \frac{768}{281 \sqrt{2\pi} \cdot \Delta } \right) e^{- \frac{78961}{1179648}\Delta^2}\\
            &\leq 2 e^{-\frac{1}{15}\Delta^2}.
    \end{aligned}
    \end{equation}
\end{enumerate}

Combining these observations together, we get
\begin{align}
    \left\| \bbeta_2 -\bbeta_3 \right\| \cdot \var_3\left[ \Assop_2 \right]
        &=  \Big\| \E_3 \left[ \Assop_2 \right] \cdot \E_3 \left[ \Assop_3 \cdot (\sfx - \bbeta_3 ) \right] 
            - \E_3 \left[ \Assop_3 \right] \cdot \E_3 \left[ \Assop_2 \cdot (\sfx - \bbeta_2 ) \Big] \right\|
            &&\because \eqref{caseB:eqB}    \nonumber\\
        &\leq \sum_{i \in \{2,3\}} \Big\| \E_3 \left[ \Assop_i \cdot (\sfx - \bbeta_i ) \right] \Big\|
            \nonumber\\
        &\leq \sum_{i \in \{2,3\}} \big\| \E_3 \left[ \Asso_i \cdot (\sfx - \bbeta_i ) \right] \big\| 
            + \sum_{i \in \{2,3\}} \big\| \E_3 \left[ \big(\Assop_i - \Asso_i \big) \cdot (\sfx - \bbeta_i ) \right] \big\|.
            \label{eqn:caseB_intermediate}
\end{align}
Then we obtain separate upper bounds for the two terms in \eqref{eqn:caseB_intermediate} as the following.

\begin{itemize}
    \item 
    \emph{First term in \eqref{eqn:caseB_intermediate}.} For each $i \in \{2,3\}$,
    \begin{align}
        \big\| \E_3 \left[ \Asso_i \cdot (\sfx - \bbeta_i ) \right] \big\|
            &\stackrel{(a)}{\leq} \sum_{s \in \{1,2\}} \big\| \E_s \left[ \Asso_i \cdot (\sfx - \bbeta_i ) \right] \big\|   
                \nonumber\\
            &\stackrel{(b)}{\leq} \sum_{s \in \{1,2\}} \left( \E_s \left[ \Asso^2_i \right] \cdot \E_s \left[ \big\| \sfx - \bbeta_i \big\|^2 \right] \right)^{1/2}
                \nonumber\\
            &\stackrel{(c)}{\leq} 12 \Delta \cdot e^{- \frac{1}{298}\Delta^2}, 
                \label{eqn:caseB_intermediate.1}
    \end{align}
    because (a) is due to \eqref{caseB:eqA}; (b) follows from Cauchy-Schwarz inequality; and (c) for any $(i,s) \in \{2,3\} \times \{1,2\}$, it follows from \eqref{eqn:caseB.separation} and Lemma \ref{lem:Psi_domination} that
    \begin{align*}
        \E_s \left[ \Asso^2_i \right]
            &\leq \E_s \left[ \Asso_i \right]\\
            &\leq \left( 1 + \frac{768}{89\sqrt{2\pi} \cdot \Delta } \right) e^{- \frac{7921}{1179648}\Delta^2} \\
            &\leq 2e^{- \frac{1}{149}\Delta^2},\\
        \E_s \big\| \sfx - \bbeta_i \big\|^2
            &\leq 2 \left( \E_s \| \sfx - \bthetastar_s \|^2 + \E_s \|\bbeta_i - \bthetastar_s\|^2 \right)\\
            &\leq 2 \left( 1 + \max_{i \in \{2,3\}} \| \bbeta_i - \bthetastar_1 \|^2 \right)\\
            &< 2 \left( 1 + \left( \frac{385}{192} \Delta \right)^2 \right)        \\
            &\leq 2 \cdot (3 \Delta)^2.
    \end{align*}

    \item 
    \emph{Second term in \eqref{eqn:caseB_intermediate}.} For each $i \in \{2,3\}$,
    \begin{align}\label{eqn:small_MSE}
        \E_3 \big\| \sfx - \bbeta_i \big\|^2
            &\leq 2 \left( \E_3 \| \sfx - \bthetastar_3 \|^2 + \E_3 \|\bbeta_i - \bthetastar_3\|^2 \right)  \nonumber\\
            &\leq 2 \left( 1 + \max_{i \in \{2,3\}} \| \bbeta_i - \bthetastar_3 \|^2 \right)    \nonumber\\
            &< 2 \left( 1 + \left( \frac{1}{192} \Delta \right)^2 \right)        \nonumber\\
            &\leq 2 \cdot \left(\frac{1}{96} \Delta \right)^2.
    \end{align}
    and therefore,
    \begin{align}
        \big\| \E_3 \left[ \big(\Assop_i - \Asso_i \big) \cdot (\sfx - \bbeta_i ) \right] \big\| 
            &\leq \left( \E_3 \left[ \big(\Assop_i - \Asso_i \big)^2 \right] \cdot \E_3 \left[ \| \sfx - \bbeta_i ) \|^2 \right] \right)^{1/2}  \nonumber\\
            &\leq \frac{1}{48} \Delta \cdot e^{-\frac{1}{30}\Delta^2}.
                \qquad\because \eqref{eqn:caseB.prox_approx} \And \eqref{eqn:small_MSE}
            \label{eqn:caseB_intermediate.2}
    \end{align}
\end{itemize}
Combining \eqref{eqn:caseB_intermediate.1} and \eqref{eqn:caseB_intermediate.2} with \eqref{eqn:caseB_intermediate}, we have
\begin{equation}\label{eqn:caseB_intermediate.3}
\begin{aligned}
    \left\| \bbeta_2 -\bbeta_3 \right\| \cdot \var_3\left[ \Assop_2 \right]
        &\leq 24 \Delta \cdot e^{- \frac{1}{298}\Delta^2}  
            + \frac{1}{24} \Delta \cdot e^{-\frac{1}{30}\Delta^2} \\
        &\leq 25 \Delta \cdot e^{- \frac{1}{298}\Delta^2}.
\end{aligned}
\end{equation}

Lastly, we apply Lemma \ref{lem:var_bound} to obtain a variance lower bound
\begin{equation}\label{eqn:caseB_intermediate.4}
    \var_3\left[ \Assop_2 \right]
        \geq \frac{1}{32 \sqrt{2\pi}} \| \bbeta_2 - \bbeta_3 \|^5 \cdot e^{-4 \left( \frac{\Delta}{96} \right)^2}
\end{equation}
because $\| \frac{\bbeta_2 + \bbeta_3}{2} - \bthetastar_3 \| + \|\bbeta_2 - \bbeta_3\| \leq 2 \max \{ \| \bbeta_2 - \bthetastar_3\|, \|\bbeta_3 - \bthetastar_3\| \} < \frac{1}{96}\Delta$. 
Combining \eqref{eqn:caseB_intermediate.3} and \eqref{eqn:caseB_intermediate.4} yields
\begin{align*}
    \left\| \bbeta_2 -\bbeta_3 \right\|^6
        &\leq 800\sqrt{2\pi} \Delta \cdot e^{(\frac{1}{2304} - \frac{1}{298} ) \Delta^2}\\
        &\leq 800\sqrt{2\pi} \Delta \cdot e^{- \frac{1}{343} \Delta^2}.
\end{align*}
Consequently,
\begin{equation}\label{eqn:caseB_beta23}
\begin{aligned}
    \max_{i \in \{2,3\}} \left\| \bbeta_i \bthetastar_3 \right\|
        &\leq \left\| \bbeta_2 -\bbeta_3 \right\|\\
        &\leq \left( 800\sqrt{2\pi} \Delta \cdot e^{- \frac{1}{343} \Delta^2} \right)^{1/6}.
\end{aligned}
\end{equation}
We conclude this case with the two upper bounds \eqref{eqn:caseB_beta1} and \eqref{eqn:caseB_beta23}.

\paragraph{Step 3 - Case C} 
In this case, $q=q_0=2$, i.e., there are two `one-fits-many' clusters. 
We observe that 
\begin{equation}\label{obs:c1}
    \bthetastar_1, \bthetastar_2 \in \vor_1
    \quad\text{and}\quad
    \bthetastar_3 \in \vor_2,
\end{equation}
which readily follows from the construction of these clusters; see Section \ref{sec:partition_construction}, cf. Algorithm \ref{alg:partition} and the definition of the set $\setA_i^{\delta}$ in \eqref{eq:setA}.
Moreover, it follows from \eqref{eqn:approx_bound.1dim.a} that 
\begin{equation}\label{eqn:caseC_approx}
    \left\| \bbeta_1 - \frac{\bthetastar_1 + \bthetastar_2}{2} \right\| < \frac{\Delta}{32}
    \quad\text{and}\quad
    \left\| \bbeta_2 - \bthetastar_3 \right\| < \frac{\Delta}{32}.
\end{equation}

We consider three possible subcases based on the location of $\bbeta_3$, namely,
\begin{enumerate}[label=(\roman*)]
    \item\label{case:C1}
    $\bbeta_3 < \bbeta_1$;
    \item\label{case:C2}
    $\bbeta_3 > \bbeta_2$;
    \item\label{case:C3}
    $\bbeta_3 \in (\bbeta_1, \bbeta_2)$.
\end{enumerate}
In the following, we discuss each of these three subcases individually.

\vspace{6pt}
{\emph{Subcase \ref{case:C1}: $\bbeta_3 < \bbeta_1$.}} 
Letting $\c = (\bbeta_3 + \bbeta_1)/2$, we observe that $\vor_3 = (-\infty, \c]$ and that $\c \leq \bthetastar_1$ by observation \ref{obs:c1}. 
Thus, $\P_1(\vor_3) = Q(\bthetastar_1 - \c)$. 
Because 
\begin{align*}
    \P_1(\vor_3) &= Q(\bthetastar_1 - \c)
        \geq \phi( \bthetastar_1 - \c + 1 ) \quad\text{by }\eqref{eqn:Q_lower_bound} 
    \quad\text{and}\\
    \P_1(\vor_3) &< \frac{1}{2^9 \cdot 3^3}  \quad\text{by }\eqref{eqn:weak_association.1dim},
\end{align*}
we have $\bthetastar_1 - \c > \tau_0$ where
\[
    \tau_0 \coloneqq \left( 2 \log (2^9 \cdot 3^3 ) \right)^{1/2} - 1 \approx 3.367.
\]

Next, we recall from the first-order stationary condition, cf. \eqref{eq:station_cond}, that 
\begin{equation}\label{eqn:caseC.sub1.beta3}
    \bbeta_3 = \frac{\sum_{s=1}^3 \E_s[\Asso_3 \cdot \sfx]}{\sum_{s=1}^3 \E_s[\Asso_3]}.
\end{equation}
We derive a lower bound for $\bbeta_3$ to argue that $\bbeta_3 \approx \c$. To this end, we begin by observing that for all $x \leq \frac{\bbeta_1 + \bbeta_2}{2}$,
\begin{align*}
    \asso_3(x) 
        &\geq \frac{\phi(x - \bbeta_3)}{2 \phi(x - \bbeta_1) + \phi(x - \bbeta_3) }\\
        &= \frac{1}{1 + 2 e^{(\bbeta_1 - \bbeta_3) \cdot ( x - \c )}}.
\end{align*}
In particular, this implies that
\begin{equation}\label{eqn:asso3_bound.1}
    \asso_3(x) \geq 1 - 2 e^{(\bbeta_1 - \bbeta_3) \cdot ( x - \c )}, \qquad \forall x \leq \c.
\end{equation}

Then we consider the numerator of \eqref{eqn:caseC.sub1.beta3}. Due to the translation invariance, we may assume $\c = 0$ without loss of generality, and therefore, it follows that
\begin{align}
    \sum_{s=1}^3 \E_s \left[ \Asso_3 \cdot \sfx \right] 
        &= \sum_{s=1}^3 \E_s \left[ \Asso_3 \cdot \sfx \cdot \indic\{ \sfx \leq \c \} \right]
            +\sum_{s=1}^3 \E_s \left[ \Asso_3 \cdot \sfx \cdot \indic\{ \sfx > \c \} \right]   \nonumber\\
        &\stackrel{(a)}{\geq} \sum_{s=1}^3 \E_s \left[ \Asso_3 \cdot \sfx \cdot \indic\{ \sfx \leq \c \} \right]   \nonumber\\
        &\stackrel{(b)}{\geq} \sum_{s=1}^3 \E_s \left[ \sfx \cdot \indic\{ \sfx \leq \c \} \right]  \nonumber\\
        &\stackrel{(c)}{=} - \sum_{s=1}^3 \phi\left( \bthetastar_s - \c \right)
            + \sum_{s=1}^3 \left( \bthetastar_s - \c \right) \cdot Q \left( \bthetastar_s - \c \right),  \label{caseC.sub1.num}
\end{align}
because (a) $\asso_3(x) \cdot x \cdot \indic\{ x > \c \} \geq 0$; (b) $\asso_3(x) \in [0,1] \And x \cdot \indic\{ x \leq \c \} \leq 0$; and (c) follows from the Gaussian integral formula \eqref{eqn:phi_integral}.

For each $s \in [3]$,
\begin{align*}
    -\phi(\bthetastar_s - \c) + \left( \bthetastar_s - \c \right) \cdot Q \left( \bthetastar_s - \c \right)
        &\geq \left( -1 + \frac{\bthetastar_s - \c }{\bthetastar_s - \c  + 1} \right) \phi \left( \bthetastar_s - \c \right)\\
        &= -\frac{1}{\bthetastar_s - \c  + 1} \phi \left( \bthetastar_s - \c \right)
\end{align*}
due to the upper and lower bounds for the Gaussian Q-function, cf. \eqref{eqn:Q_bounds} and Lemma \ref{lem:gaussian_tail}. 
Plugging this into \eqref{caseC.sub1.num}, we get 
\begin{equation}\label{caseC.sub1.num2}
\begin{aligned}
    \sum_{s=1}^3 \E_s \left[ \Asso_3 \cdot \sfx \right] 
        &\geq - \sum_{s=1}^3 \frac{1}{\bthetastar_s - \c  + 1} \phi \left( \bthetastar_s - \c \right)\\
        &\geq - \frac{3}{\bthetastar_1 - \c + 1}\phi \left( \bthetastar_1 - \c \right).
\end{aligned}
\end{equation}

Thereafter, we consider the denominator of \eqref{eqn:caseC.sub1.beta3} and observe that
\begin{align*}
    \sum_{s=1}^3 \E_s \left[ \Asso_3 \right] 
        &\stackrel{(a)}{\geq} \E_1 \left[ \Asso_3  \cdot \indic\{\sfx \leq \c\} \right] \\
        &\stackrel{(b)}{\geq} \E_1 \left[ \indic\{\sfx \leq \c\} \right] 
            - \E_1 \left[ 2 e^{(\bbeta_1 - \bbeta_3) \cdot ( \sfx - \c )} \cdot \indic\{\sfx \leq \c\} \right]
\end{align*}
because (a) $\asso_3(x) \geq 0$ and (b) $\asso_3(x) \geq 1 - 2 e^{(\bbeta_1 - \bbeta_3) \cdot ( x - \c )},~\forall x \leq \c$ by \eqref{eqn:asso3_bound.1}. 
Observe that 
\[
    \E_1 \left[ \indic\{\sfx \leq \c\} \right]
        = Q(\bthetastar_1 - \c)
\]
and that
\begin{align*}
    \E_1 \left[ e^{(\bbeta_1 - \bbeta_3) \cdot ( \sfx - \c )} \cdot \indic\{\sfx \leq \c\} \right]
        &= \E_1 \left[ e^{(\bbeta_1 - \bbeta_3) \cdot ( \sfz - (\c - \bthetastar_1)  )} \cdot \indic\{\sfz \leq \c - \bthetastar_1\} \right]\\
        &= \frac{1}{\sqrt{2\pi}} \cdot \int_{-\infty}^{\c - \bthetastar_1} e^{(\bbeta_1 - \bbeta_3) \cdot ( \sfz - (\c - \bthetastar_1)  )} \cdot e^{-\frac{z^2}{2}} ~\ddup z\\
        &= \frac{1}{\sqrt{2\pi}} \cdot e^{\frac{\bbeta_1 - \bbeta_3}{2} \cdot \left( \bbeta_1 - \bbeta_3 - 2(\c-\bthetastar_1)\right)} 
            \int_{-\infty}^{\c - \bthetastar_1} e^{-\frac{1}{2}\left( z - (\bbeta_1 -\bbeta_3) \right)^2} ~\ddup z  
            \qquad\because \eqref{eqn:gaussian_shift}\\
        &= e^{-\frac{1}{2} (\bthetastar_1 - \c)^2 } \cdot e^{\frac{1}{2} \left( (\bbeta_1 - \bbeta_3) + (\bthetastar_1 - \c)\right)^2} 
            \cdot Q \big( (\bbeta_1 - \bbeta_3) + (\bthetastar_1 - \c) \big).
\end{align*}
Thus, we obtain
\begin{equation}\label{caseC.sub1.denom}
\begin{aligned}
    \sum_{s=1}^3 \E_s \left[ \Asso_3 \right] 
        &= Q(\bthetastar_1 - \c) 
         - 2 e^{-\frac{1}{2} (\bthetastar_1 - \c)^2 } \cdot e^{\frac{1}{2} \left( (\bbeta_1 - \bbeta_3) + (\bthetastar_1 - \c)\right)^2}
            \cdot Q \left( (\bbeta_1 - \bbeta_3) + (\bthetastar_1 - \c) \right)
\end{aligned}
\end{equation}

Further proceeding with the lower bound in \eqref{caseC.sub1.denom}, we use the upper and lower bounds for the Gaussian Q-function, cf. \eqref{eqn:Q_bounds} and Lemma \ref{lem:gaussian_tail}, to obtain
\begin{align}
    \sum_{s=1}^3 \E_s \left[ \Asso_3 \right] 
        &\geq \frac{1}{\bthetastar_1 - \c + 1} \phi(\bthetastar_1 - \c) - 2 \cdot \phi \big( (\bbeta_1 - \bbeta_3) + (\bthetastar_1 - \c) \big)  
            \cdot \frac{e^{-\frac{1}{2} (\bthetastar_1 - \c)^2 } \cdot e^{\frac{1}{2} \left( (\bbeta_1 - \bbeta_3) + (\bthetastar_1 - \c)\right)^2} }{(\bbeta_1 - \bbeta_3) + (\bthetastar_1 - \c)}  \nonumber\\
        &\geq \frac{1}{3} \cdot \frac{1}{\bthetastar_1 - \c + 1} \cdot \phi(\bthetastar_1 - \c)     \label{caseC.sub1.denom2}
\end{align}
with the last inequality following from $\bbeta_1 - \bbeta_3 = 2 (\bbeta_1 - \c) > 2(\bthetastar_1 + \frac{15}{32}\Delta - \c ) \geq 2 (\bthetastar_1 - \c + 1)$.

In the end, we combine the lower bounds \eqref{caseC.sub1.num2} and \eqref{caseC.sub1.denom2} with the stationary condition for $\bbeta_3$ in \eqref{eqn:caseC.sub1.beta3} to obtain 
\[
    \bbeta_3 - \c \geq \frac{- \frac{3}{\bthetastar_1 - \c + 1}\phi \left( \bthetastar_1 - \c \right)}{\frac{1}{3} \cdot \frac{1}{\bthetastar_1 - \c + 1} \cdot \phi(\bthetastar_1 - \c)} = -9.
\]
This implies that
\[
    9 \geq \| \bbeta_3 - \c \| = \| \bbeta_1 - \c \| \geq \| \bbeta_1 - \bthetastar_1\| > \frac{15}{32} \Delta,
\]
which is a contradiction. Consequently, we conclude that Subcase \ref{case:C1} of Case C cannot happen.

\vspace{6pt}
{\emph{Subcase \ref{case:C2}: $\bbeta_3 > \bbeta_2$.}} 
We recall the equivalent stationary condition (Theorem \ref{thm:equiv}), cf. \eqref{eq:equiv}:
\begin{align*}
    \sum_{j\in[\kfit]}\bbeta_{j}\sum_{s\in[\ktrue]}\E_{s}\left[\Asso_{i}\Asso_{j}\right] =\sum_{s\in[\ktrue]}\bthetastar_{s}\E_{s}\left[\Asso_{i}\right], \qquad\forall i \in [\kfit].
\end{align*}
Letting $i = 3$ and rearranging the terms, we obtain
\begin{align*}
    \sum_{j\in[\kfit]}\bbeta_{j}\sum_{s \in \{1,2\} }\E_{s}\left[\Asso_{3}\Asso_{j}\right]  -  \sum_{s\in \{1,2\} }\bthetastar_{s}\E_{s}\left[\Asso_{3}\right]
    + \sum_{j \in [\kfit]} \left( \bbeta_j - \bthetastar_3 \right) \cdot \E_3 \left[\Asso_{3}\Asso_{j}\right]  = 0,
\end{align*}
whence it follows that 
\begin{equation}\label{eqn:caseC.sub2.1}
\begin{aligned}
    &\left( \bbeta_3 - \bthetastar_3 \right) \cdot \E_3 \left[\Asso_{3}\Asso_{3}\right] - \left( \bbeta_2 - \bthetastar_3 \right) \cdot \E_3 \left[\Asso_{3}\Asso_{2}\right]
        = - \left( \bbeta_1 - \bthetastar_3 \right) \cdot \E_3 \left[\Asso_{3}\Asso_{1}\right]
            + \sum_{s \in \{1,2\} } \sum_{j\in[\kfit]} \left(\bthetastar_{s} - \bbeta_j \right) \cdot \E_{s}\left[\Asso_{3}\Asso_{j}\right] 
\end{aligned}
\end{equation}
Then we derive an upper bound for the absolute value of the expression on the right-hand side of \eqref{eqn:caseC.sub2.1}.
\begin{itemize}
    \item 
    Note that $\| \bbeta_1 - \bthetastar_3 \| \leq \frac{49}{32} \Delta \leq 2 \Delta$, cf. \eqref{eqn:approx_bound.1dim.a}. 
    Moreover, $\| \bbeta_1 - \bthetastar_3 \| - \| \bbeta_1 - \bthetastar_3 \| \geq \frac{23}{16}\Delta \geq \sqrt{2} \Delta$. 
    Applying Lemma \ref{lem:Psi_domination}, we have  
    \begin{equation}\label{eqn:caseC.sub2.upper.a}
    \begin{aligned}
        \big| - \left( \bbeta_1 - \bthetastar_3 \right) \cdot \E_3 \left[\Asso_{3}\Asso_{1}\right] \big|
            &\leq \left| \bbeta_1 - \bthetastar_3 \right| \cdot \E_3 \left[\Asso_{1}\right]\\
            &\leq 2 \Delta \cdot \left( 1 + \frac{2}{\sqrt{\pi} \Delta} \right) \cdot e^{-\frac{\Delta^2}{16}}\\
            &\leq 3 \Delta \cdot e^{-\frac{\Delta^2}{16}}.
    \end{aligned}
    \end{equation}

    \item
    Similarly, for each $(s,j) \in [2] \times [3]$, 
    \begin{equation}\label{eqn:caseC.sub2.upper.b}
    \begin{aligned}
        \left| \left(\bthetastar_{s} - \bbeta_j \right) \cdot \E_{s}\left[\Asso_{3}\Asso_{j}\right]  \right|
            & \leq \left| \bthetastar_{s} - \bbeta_j \right| \cdot \E_{s}\left[\Asso_{3} \right]     \\
            & \leq \left| (\bbeta_3 - \bthetastar_3) + 2\Delta \right| \cdot \left( 1 + \frac{128}{ 15 \sqrt{2\pi} \cdot\Delta  }\right)  
                \cdot e^{-\frac{\left( (\bbeta_3 - \bthetastar_3) + \frac{15}{32}\Delta \right)^2}{32}}\\
            & \leq 3 \Delta \cdot e^{- \frac{225}{32768} \Delta^2}.      
    \end{aligned}
    \end{equation}
\end{itemize}
Noticing that $\bbeta_3 - \bthetastar_3 \geq \bthetastar_3 - \bbeta_2$ (because $\bthetastar_2 \in \vor_2$), and gathering the upper bounds \eqref{eqn:caseC.sub2.upper.a} and \eqref{eqn:caseC.sub2.upper.b}, we obtain from \eqref{eqn:caseC.sub2.1} that
\begin{equation}\label{eqn:caseC.sub2.2}
\begin{aligned}
    \left( \bbeta_3 - \bthetastar_3 \right) \cdot \E_3 \left[\Asso_{3} \cdot (\Asso_{3} - \Asso_2) \right]
        &\leq 3 \Delta \cdot e^{-\frac{\Delta^2}{16}} + 3 \Delta \cdot e^{- \frac{225}{32768} \Delta^2}\\
        &\leq 6 \Delta \cdot e^{- \frac{1}{146} \Delta^2}.
\end{aligned}
\end{equation}

Next, we proceed to prove a lower bound for $\E_3 \left[\Asso_{3} \cdot (\Asso_{3} - \Asso_2) \right]$. 
Defining a function $\assop: \real \to \real$ so that\footnote{Note that here we define $\assop$ differently from that used in Step 3 - Case A, although the underlying intuition is the same.} 
\[
    \assop(x) = \frac{e^{-\|x - \bbeta_2\|^2/2}}{e^{-\|x - \bbeta_2\|^2/2} + e^{-\|x - \bbeta_3\|^2/2}},
\]
Note that $\assop(x) \geq \asso_2(x)$ for all $x \in \real$. Letting $\Assop = \assop(\sfx)$ and $\bAssop = 1 - \Assop$, we thus have
\begin{equation}\label{eqn:caseC.sub2.3}
\begin{aligned}
    \E_3 \left[ \Asso_3 \cdot (\Asso_3 - \Asso_2 ) \right]
        &\geq \E_3 \left[ \Asso_3 \cdot (\Asso_3 - \Assop ) \right]   \\
        &= \E_3 \left[ \bAssop \cdot (\bAssop - \Assop) \right] 
            + \E_3 \left[ \Asso_3 \cdot (\Asso_3 - \Assop ) \right]
            - \E_3 \left[ \bAssop \cdot (\bAssop - \Assop) \right]  \\
        &= \E_3 \left[ \bAssop \cdot (\bAssop - \Assop) \right] 
             + \E_3 \left[ ( \Asso_3 + \bAssop - \Assop ) \cdot (\Asso_3 - \bAssop) \right]. 
\end{aligned}
\end{equation}
Now we analyze the two terms in the right-hand side of \eqref{eqn:caseC.sub2.3}. 
\begin{itemize}
    \item 
    First, we observe the following inequality by modifying the proof of Lemma \ref{lem:separation}, cf. \eqref{eqn:lem_separation_alt}:
    \begin{equation}\label{eqn:caseC.sub2.4}
    \begin{aligned}
        \E_3 \left[ \bAssop \cdot (\bAssop - \Assop) \right]
            &= \E_3 \left[ \bAssop \cdot (2\bAssop - 1) \right]\\
            &\geq \frac{1}{4} \left[ Q\left( \frac{1}{\delta} - c \right) -  \frac{\phi(c)}{\phi(c+2\alpha)} \cdot Q\left( \frac{1}{\delta} - c - 2\alpha\right) \right]
    \end{aligned}
    \end{equation}
    where $\alpha = \bbeta_2 - \bthetastar_3$,  $c = \frac{\bbeta_2 + \bbeta_3}{2} - \bthetastar_3$, and $\delta = \frac{\bbeta_3 - \bbeta_2}{2}$. 
    Note that 
    \begin{equation}\label{eqn:caseC.sub2.5}
    \begin{aligned}
        \frac{\phi(c)}{\phi(c+2\alpha)} \cdot Q\left( \frac{1}{\delta} - c - 2\alpha\right)
            &= e^{\frac{2\alpha}{\delta}} \cdot \frac{\phi(\frac{1}{\delta} - c)}{\phi(\frac{1}{\delta} - c - 2\alpha)} \cdot Q\left( \frac{1}{\delta} - c - 2\alpha\right)
                &&\because \eqref{eqn:pdf_ratio} \\
            &\leq e^{\frac{2\alpha}{\delta}} \cdot Q\left( \frac{1}{\delta} - c \right)
    \end{aligned}
    \end{equation}
    because $Q(x)/\phi(x)$ is monotone increasing over $x \in \real$.

    \item
    Second, since $|\Asso_3 + \bAssop - \Assop | \leq 2$, we have
    \begin{align}
        &\left| \E_3 \left[ ( \Asso_3 + \bAssop - \Assop ) \cdot (\Asso_3 - \bAssop) \right] \right| \nonumber\\
            &\qquad\leq 2 \cdot \E_3 \left[ \bAssop - \Asso_3 \right] \nonumber\\
            &\qquad= 2 \cdot \E_3 \Bigg[ \frac{\phi( \sfx - \bbeta_1 ) \cdot \phi( \sfx - \bbeta_3 )}{ \big[ \phi( \sfx - \bbeta_1 )  + \phi( \sfx - \bbeta_2 )  + \phi( \sfx - \bbeta_3 ) \big] } \cdot \frac{1}{ \big[ \phi( \sfx - \bbeta_2 ) + \phi( \sfx - \bbeta_3 ) \big]  } \Bigg]      \nonumber\\
            &\qquad\leq 2 \cdot \E_3 \left[ \Asso_1 \right]   \nonumber\\
            &\qquad\stackrel{(a)}{\leq} 2 \cdot \left( 1 + \frac{2}{\sqrt{\pi} \Delta} \right) \cdot e^{-\frac{\Delta^2}{16}}
                \nonumber\\
            &\qquad\leq 3 \cdot e^{-\frac{\Delta^2}{16}},       \label{eqn:caseC.sub2.6}
    \end{align}
    where (a) follows from Lemma \ref{lem:Psi_domination}; also see \eqref{eqn:caseC.sub2.upper.a}.
\end{itemize}

Combining \eqref{eqn:caseC.sub2.4}, \eqref{eqn:caseC.sub2.5} and \eqref{eqn:caseC.sub2.6} with \eqref{eqn:caseC.sub2.3}, we obtain
\begin{equation}\label{eqn:caseC.sub2.7}
\begin{aligned}
    \E_3 \left[ \Asso_3 \cdot (\Asso_3 - \Asso_2 ) \right]
        &\geq \frac{1}{4} \cdot \left( 1 - e^{4\frac{\bbeta_2 - \bthetastar_3}{\bbeta_3 - \bbeta_2}} \right) \cdot Q \left( \frac{2}{\bbeta_3 - \bbeta_2} - \frac{\bbeta_2 + \bbeta_3}{2} + \bthetastar_3 \right)
         - 3 \cdot e^{-\frac{\Delta^2}{16}}.
\end{aligned}
\end{equation}
Inserting \eqref{eqn:caseC.sub2.7} to \eqref{eqn:caseC.sub2.2} and observing $\bbeta_3 - \bthetastar_3 \geq \frac{1}{2} (\bbeta_3 - \bbeta_2)$ yields
\begin{equation}\label{eqn:caseC.sub2.8}
\begin{aligned}
    \frac{1}{8} \cdot \left( \bbeta_3 - \bbeta_2 \right) \cdot \left( 1 - e^{4\frac{\bbeta_2 - \bthetastar_3}{\bbeta_3 - \bbeta_2}} \right) 
        \cdot Q \left( \frac{2}{\bbeta_3 - \bbeta_2} - \frac{\bbeta_2 + \bbeta_3}{2} + \bthetastar_3 \right)
        \leq 9 \Delta \cdot e^{-\frac{1}{146}\Delta^2}.
\end{aligned}
\end{equation}
If the inequality \eqref{eqn:caseC.sub2.8} holds, then either of the following must be true: (1) $\bbeta_3 - \bbeta_2$ is very small, namely, $\bbeta_3 - \bbeta_2 \leq \frac{20}{\Delta}$, or (2) $\bbeta_2 - \bthetastar_3$ is positive, or very close to $0$ so that $\bthetastar_3 - \bbeta_2 \leq e^{-C \cdot \Delta^2}$ for some sufficiently small constant (e.g., $C = 1/200$). 
Note that (1) cannot be the case, as it will violate the near-empty association condition $\P_3[\vor_3] \approx 0$.  Also, $\bbeta_2 - \bthetastar_3$ cannot be positive. 
Thus, the only possibility is to have $\bthetastar_3 - \bbeta_2 \leq e^{-C \cdot \Delta^2}$, which implies that $\bbeta_3 - \bthetastar_3 \geq C' \Delta$ (e.g., $C' = 1/20$). 

All in all, we conclude that this subcase may not be possible to happen, but if it occurs, then the following must be true: there exist some absolute constant $C > 0$ such that
\begin{align*}
    \| \bbeta_1 - \frac{1}{2}(\bthetastar_1 + \bthetastar_2) \| &\leq e^{- C \Delta^2},\\
    \| \bbeta_2 - \bthetastar_3 \| &\leq e^{- C \Delta^2},\quad\text{and}\\
    \E_*[\Asso_3] &\leq e^{-C \Delta^2}.
\end{align*}
The approximation error bounds follow form Lemma \ref{lem:accu_domination}.

\vspace{6pt}
{\emph{Subcase \ref{case:C3}: $\bbeta_3 \in (\bbeta_1, \bbeta_2)$.}} 
We reach at a similar conclusion to Subcase \ref{case:C2} using a similar argument. We omit the details.

\paragraph{Step 3 - Case D. } 
Recall from \eqref{eqn:approx_bound.1dim.a} in Step 0 of this proof that 
\[
    \| \bbeta_i - \bthetastar_i \| < \frac{\Delta}{32}, \qquad \forall i \in [3].
\]
Thus, $\min_{j \neq i} \| \bbeta_j - \bthetastar_i \| - \| \bbeta_i - \bthetastar_i \| > \frac{15\Delta}{16}$ for all $i \in [3]$. 
Observe that the premise of Lemma \ref{lem:accu_domination} is satisfied with $\delta = 15\Delta/ 16 \geq \max \left\{ \frac{4}{\sqrt{2\pi}}, \, 8 \sqrt{\log ( 2 \sqrt{2} \cdot \kfit \ktrue)} \right\}$. 
Therefore, it follows from Lemma \ref{lem:accu_domination} that for all $i \in [3]$,
\begin{align*}
    \| \bbeta_i - \bthetastar_i \| 
        &\leq 4 \kfit \ktrue \deltamax \cdot e^{- \frac{\delta^2}{64}}\\
        &= 72 \Delta \cdot e^{- \frac{225}{16384}\Delta^2}
\end{align*}
because $\deltamax = 2 \Delta$ and $\delta = \frac{15}{16}\Delta$.
\subsection{Proof of Corollary~\ref{cor:under}}\label{sec:proof_under}

\begin{proof}[Proof of Corollary~\ref{cor:under}]
The proof follows the same lines as in the proof of Theorem~\ref{thm:boost} in Section~\ref{sec:proof_main_boost}. 
By rescaling, we may assume unit variance $\std^{2}=1$. 
When $\ktrue=3$ and $\kfit=2$, the value $q$ and the sets $\left\{ S_{a}\right\} $ and $\left\{ S_{a}^{*}\right\} $ in Theorem~\ref{thm:master} can only have, up to permutation of component labels, the following possibilities: 
\begin{enumerate}
    \item \label{enu:underproof_1fit3}
    $(q, q_0, |\cA_0|) = (1,1,1)$. 
    \begin{enumerate}
        \item
        $S_{0}=\{2\}$; $S_{1}=\left\{ 1\right\} $, $S_{1}^{*}=\left\{ 1,2,3\right\} $.
    \end{enumerate}
    \item $(q, q_0, |\cA_0|)=(2,2,0)$.
        \begin{enumerate}
        \item \label{enu:underproof_1fit2_1fit1}$S_{1}=\left\{ 1\right\} ,S_{1}^{*}=\left\{ 1,2\right\} ;$
        $S_{2}=\left\{ 2\right\} ,S_{2}^{*}=\left\{ 3\right\} $; 
        \item \label{enu:underproof_1fit2_1fit1ext}$S_{1}=\left\{ 1\right\} ,S_{1}^{*}=\left\{ 1,3\right\} ;$
        $S_{2}=\left\{ 2\right\} ,S_{2}^{*}=\left\{ 2\right\} $. 
    \end{enumerate}
\end{enumerate}

We claim that Case \ref{enu:underproof_1fit2_1fit1ext} above, where $\bbeta_{1}$ fits two non-adjacent centers $\{\bthetastar_{1},\bthetastar_{3}\}$, is impossible. 
If we assume it is possible, then we must have $\bbeta_{1}\neq\bbeta_{2}$ by Theorem~\ref{thm:main}; say $\bbeta_{1}<\bbeta_{2}$. 
In this case, it holds that $\vor_{1}\subset(-\infty,\bbeta_{2})\subset(-\infty,\bthetastar_{3}]$,
where the last inclusion holds since $\left|\bbeta_{2}-\bthetastar_{2}\right|\le \Delta/192$ by \eqref{eqn:approx_bound.1dim.b} (cf. Theorem~\ref{thm:master}). 
It follows that $\P_{3}(\vor_{1})\le\P_{3}\left((-\infty,\bthetastar_{3})\right)=\frac{1}{2},$
contradicting the inequality $\P_{3}(\vor_{1})\ge1-1/(2^9 \cdot 3^3)$ in \eqref{eqn:weak_association.1dim}.

In Case~\ref{enu:underproof_1fit3} above, $\bbeta_{1}$ fits all three true centers and $\bbeta_{2}$ has near-empty association. 
This case is impossible by an argument similar to Case A in the proof of Theorem~\ref{thm:boost} (see Section~\ref{sec:1dim_proof}, Step 3 - Case A).

In Case~\ref{enu:underproof_1fit2_1fit1} above, $\bbeta_{1}$ fits $\{\bthetastar_{1},\bthetastar_{2}\}$ and $\bbeta_{2}$ fits $\bthetastar_{3}$. 
By an argument similar to Case B in the proof of Theorem~\ref{thm:boost} (see Section~\ref{sec:1dim_proof}, Step 3 - Case B), we find that the exponential error bounds in equation~(\ref{eq:under}) of Corollary~\ref{cor:under} must hold.
\end{proof}


\newpage
\bibliographystyle{plain}
\bibliography{ref}

\appendix
\section{Deferred Proof of Proposition \ref{prop:small_bdr1}}\label{sec:proof_proposition.2}

\subsection{Voronoi Cells and Their Geometry\label{sec:voronoi}}
In our analysis, the coefficients of association $\asso_{i}(\x), ~i \in [\kfit]$ defined in Definition \ref{defn:coeff_asso} play a key role in characterizing the gradient, the Hessian and optimality conditions of $L$. 
These quantities represent the strength of (soft-) associations between a data point $\x\in\real^{d}$ and the centers $\bbeta_{i}$, which is quantified by the relative magnitudes of the squared distances between $\x$ and the $\kfit$ centers; see \eqref{eq:asso}. 
To better understand the properties of $\asso_{i}(\x)$, it is useful to study the hard-association analogue thereof (i.e., $\asso_{i}(\x)$ in the limit $\sigma \to 0$), where a data point is associated only with the closest among the $\kfit$ centers. 
This hard association induces a partition of the space $\real^{d}$, which is the so-called Voronoi diagram of $\real^d$ generated by $\{ \bbeta_1, \dots, \bbeta_{\kfit}\}$.

In this section, we take a closer look at this Voronoi diagram and elucidate its relationship with the association coefficients $\asso_{i}(\x)$. 
Specifically, we recall the definition of Voronoi cells and introduce additional useful notions related to them. 
Thereafter, we state and prove Lemma \ref{lem:geometry} that will be used in the proof of Proposition \ref{prop:small_bdr1} in Section \ref{sec:complete_proof_proposition2}.

\paragraph{Useful notions related to Voronoi cells}
We begin by recalling the definition of Voronoi cells from Definition \ref{defn:voronoi}. 
Letting $\B = (\bbeta_i)_{i=1}^{\kfit}$, the $i$-th Voronoi cell associated with $\B$ has the following representations, cf. \eqref{eq:vor_def}:
\begin{align*}
    \vor_{i} 
        &= \vor_{i}(\B) \\
        & \coloneqq\left\{ \x\in\real^{d}:\left\Vert \x-\bbeta_{i}\right\Vert \le\left\Vert \x-\bbeta_{j}\right\Vert ,\forall j\in[\kfit]\right\} \\
        & =\left\{ \x \in \real^d :\asso_{i}(\x)\ge\asso_{j}(\x),~\forall j\in[\kfit]\right\}.
\end{align*}

For $i, j \in [\kfit]$ with $i \neq j$, we let 
\begin{equation}\label{eq:bbeta_bar}
    \bbetabar_{ij}\coloneqq \frac{\bbeta_{i}+\bbeta_{j}}{2}
\end{equation}
denote the mid point of $\bbeta_{i}$ and $\bbeta_{j}$. Then the Voronoi cell can be equivalently represented as
\begin{align*}
    \vor_i = \big\{ \x \in \real^d: \left\langle \x - \bbetabar_{ij}, \bbeta_i - \bbetabar_{ij} \right\rangle &\geq 0,
        \qquad \forall j \in [\kfit]\setminus\{i\} \big\}.
\end{align*}
This representation makes it clear that $\vor_{i}$ is a polyhedron generated by at most $\kfit - 1$ linear inequalities.

Next, we define the set of points that are equidistant from $\bbeta_{i}$ and $\bbeta_{j}$: for $i, j \in [\kfit]$ with $i \neq j$,
\begin{equation}\label{eq:bdr}
    \begin{aligned}
        \bdr_{ij} 
            & \coloneqq\left\{ \x \in \real^d :\left\Vert \x-\bbeta_{i}\right\Vert =\left\Vert \x-\bbeta_{j}\right\Vert \right\} \\
            & =\left\{ \x \in \real^d :\asso_{i}(\x)=\asso_{j}(\x)\right\} \\
            & =\left\{ \x \in \real^d :\left\langle \x-\bbetabar_{ij},\bbeta_{i}-\bbetabar_{ij}\right\rangle =0\right\}.
    \end{aligned}
\end{equation}
It is clear from the last expression of $\bdr_{ij}$ that $\bdr_{ij}$ is an affine subspace of codimension at most $1$; observe that $\bdr_{ij} = \real^d$ if and only if $\bbeta_i = \bbeta_j$. 
Note that $\bdr_{ij}$ is the affine hull of $\vor_i \cap \vor_j$.

\begin{remark}
    Note that if the $\bbeta_{i}$'s in $\B$ are distinct, then so are their associated Voronoi cells.
    In this case, the Voronoi cells $\{ \vor_i \}_{i=1}^{\kfit}$ form a partition of $\real^{d}$, up to the (Lebesgue) measure-zero boundaries $\vor_{i}\cap\vor_{j}\subseteq\bdr_{ij}$. 
    On the other hand, if $\bbeta_{i}=\bbeta_{j}$ for some pair $i,j\in[\kfit]$, then $\vor_{i}=\vor_{j}$ and $\bdr_{ij}=\real^{d}$.
\end{remark}

For a parameter $\alpha\ge0$, we define two parameterized families of sets 
\begin{equation}\label{eqn:alpha_enlarged}
    \begin{aligned}
        \softvor_{i}^{\alpha} 
            & \coloneqq\bigg\{ \x \in \real^d :\left\langle \x-\bbetabar_{ij},~\frac{\bbeta_{i}-\bbetabar_{ij}}{\|\bbeta_{i}-\bbetabar_{ij}\|}\right\rangle \ge-\alpha \sigma, \quad ~\forall j\in[\kfit]\setminus\{i\}\bigg\},\\
        \softbdr_{ij}^{\alpha}
            & \coloneqq \left\{ \x \in \real^d : \left| \left\langle \x-\bbetabar_{ij}, ~\frac{\bbeta_{i}-\bbetabar_{ij}}{\|\bbeta_{i}-\bbetabar_{ij}\|} \right\rangle \right| \le\alpha \sigma \right\} .
    \end{aligned}
\end{equation}
The sets $\softvor_{i}^{\alpha}$ and $\softbdr_{ij}^{\alpha}$ are the $\alpha\cdot \sigma$-enlargements of $\vor_{i}$ and $\bdr_{ij}$, respectively, with $\softvor_{i}^{0}=\vor_{i}$ and $\softbdr_{ij}^{0}=\bdr_{ij}$. 

\begin{remark}
    For any $\alpha \geq 0$, and any $i, j \in [\kfit]$ with $i \neq j$, we have $\softvor_{i}^{\alpha}\cap\softvor_{j}^{\alpha} \subseteq \softbdr_{ij}^{\alpha}$.
    Moreover, if $\bbeta_{i}=\bbeta_{j}$ for some pair $i,j\in[\kfit]$, then $\softvor_{i}^{\alpha}=\softvor_{j}^{\alpha}$ and $\softbdr_{ij}^{\alpha}=\real^{d}$.
\end{remark}

Finally, for a parameter $\delta > 0$ and $i, j \in [\kfit]$ with $i \neq j$, we let 
\begin{equation}\label{eq:goodset}
    \goodset_{ij}^{\delta}\coloneqq\left\{ \x \in \real^d :\asso_{i}(\x)\asso_{j}(\x)\ge \delta \right\},
\end{equation}
which denotes the set of points $\x \in \real^d$ that are simultaneously associated with both $\bbeta_i$ and $\bbeta_j$ (at level $\delta$).

\paragraph{A useful lemma}

The following lemma establishes the relationship among (1) the association coefficient $\asso_{i}(\x)$; (2) the $\alpha$-enlarged Voronoi cells $\softvor_{i}^{\alpha}$, $\softbdr_{ij}^{\alpha}$; and (3) the set $\goodset_{ij}^{\delta}$ for some choices of $\alpha \geq 0$ and $\delta > 0$.

\begin{lemma}[Soft Voronoi cells and boundaries]\label{lem:geometry}
    Let $\B \in \real^{d \times k}$ be an arbitrary ordered set of vectors. 
    For any $\alpha \in \real$, let
    \begin{align*}
        \alpha' &= \alpha'_{\alpha, \B, \sigma} 
            \coloneqq \alpha + \frac{\sigma}{\|\bbeta_i - \bbeta_j \|} \log \kfit,\\
        \delta &= \delta_{\alpha, \B, \sigma} 
            \coloneqq \frac{1}{\kfit^2} \exp \left( - 2\alpha \frac{ \| \bbeta_i - \bbeta_j \| }{\sigma} \right),\\
        \delta' &= \delta'_{\alpha, \B, \sigma} 
            \coloneqq \frac{1}{\kfit^2} \exp \left( - 3\alpha \frac{ \| \bbeta_i - \bbeta_j \| }{\sigma} \right).
    \end{align*}
    
    For each $i, j \in[\kfit]$ such that $i \neq j$, and for any $\alpha \in \real$, the following inclusion relations hold.
    \begin{enumerate}
    \item
    First, letting 
    \[
        \mathcal{S} \coloneqq \left\{ \x \in \real^d : \asso_{i}(\x)\ge  \frac{1}{\kfit} \exp \left( - \alpha \frac{ \| \bbeta_i - \bbeta_j \| }{\sigma} \right) \right\}, 
    \]
    we observe that 
    \begin{equation}\label{eq:geometry1}
        \softvor_{i}^{\alpha} 
            \subseteq \mathcal{S}
            \subseteq\softvor_{i}^{\alpha'};
    \end{equation}
    \item
    Second,
    \begin{equation}
        \softvor_{i}^{\alpha}\cap\softvor_{j}^{\alpha}
            \subseteq\goodset_{ij}^{\delta}
            \subseteq\softvor_{i}^{2\alpha'}\cap\softvor_{j}^{2\alpha'}.\label{eq:geometry2}
    \end{equation}
    \item
    Third,
    \begin{equation}\label{eq:geometry3}
        \softbdr_{ij}^{\alpha}\cap\softvor_{j}^{\alpha} \subseteq \goodset_{ij}^{\delta'}.
    \end{equation}
    \end{enumerate}
\end{lemma}

\begin{proof}[Proof of Lemma \ref{lem:geometry}]
    We begin this proof by making a preparatory observation. 
    Fix arbitrary $i, j\in[\kfit]$ with $i \neq j$, and recall from \eqref{eq:bbeta_bar} that $\bbetabar_{i j} =(\bbeta_{i}+\bbeta_{j})/2$. 
    Then we observe the equivalence of the following expressions: for any $\alpha \in \real$,
    \begin{align}
        &\left\langle \x-\bbetabar_{ij},~\frac{\bbeta_{i}-\bbetabar_{ij}}{\|\bbeta_{i}-\bbetabar_{ij}\|}\right\rangle \ge-\alpha \sigma    \nonumber\\
            &\Leftrightarrow~  
                \frac{1}{2 \| \bbeta_i - \bbeta_j \|} \left\{ \| \x - \bbeta_j \|^2 - \| \x - \bbeta_i \|^2 \right\} \ge -\alpha \sigma \nonumber\\
            &\Leftrightarrow~
                -\frac{1}{2} \| \x - \bbeta_i \|^2 \geq -\frac{1}{2} \| \x - \bbeta_j \|^2 - \alpha \sigma \| \bbeta_i - \bbeta_j \| \nonumber\\
            &\Leftrightarrow~
                \exp \left( -\frac{1}{2 \sigma^2} \| \x - \bbeta_i \|^2 \right) \nonumber\\
                &\quad\geq \exp \left( -\frac{1}{2 \sigma^2} \| \x - \bbeta_j \|^2 \right) 
                    \cdot \exp \left( - \alpha \frac{ \| \bbeta_i - \bbeta_j \| }{\sigma} \right) \nonumber\\
            &\Leftrightarrow~
                f_i(\x) \geq f_j(\x) \cdot \exp \left( - \alpha \frac{ \| \bbeta_i - \bbeta_j \| }{\sigma} \right) \nonumber\\
            &\Leftrightarrow~
                \asso_i(\x) \geq \asso_j(\x) \cdot \exp \left( - \alpha \frac{ \| \bbeta_i - \bbeta_j \| }{\sigma} \right).   \label{eq:geo_equiv}
    \end{align}
    In the remainder of this proof, we prove each of the three claims in the lemma.

    \bigskip
    \textit{\underline{(1) Proof of Claim 1.}}
    First of all, we observe by \eqref{eqn:alpha_enlarged} and \eqref{eq:geo_equiv} that for any $\alpha \in \real$,
    \begin{align*}
        \x \in \softvor_i^{\alpha}
            \qquad\implies\qquad   \asso_i(\x) \geq \asso_j(\x) \cdot \exp \left( - \alpha \frac{ \| \bbeta_i - \bbeta_j \| }{\sigma} \right), \quad\forall j \in [\kfit]\setminus\{i\}.
    \end{align*}
    Because $\sum_{i' \in [\kfit]} \asso_{i'}(\x) \equiv 1$, we obtain
    \begin{align*}
        \asso_i(\x) 
            &\geq \frac{1}{ 1 + (\kfit - 1) \cdot \exp \left( \alpha \frac{ \| \bbeta_i - \bbeta_j \| }{\sigma} \right) }\\
            &\geq \frac{1}{\kfit} \exp \left( - \alpha \frac{ \| \bbeta_i - \bbeta_j \| }{\sigma} \right).
    \end{align*}

    To prove the second inclusion, we observe that for any $\alpha'' \in \real_+$,
    \begin{align*}
        \asso_i(\x) \geq \alpha''
            ~~&\implies~~ \asso_i(\x) = \frac{f_i(\x)}{\sum_{j \in [\kfit]} f_j(\x) } \geq \alpha'' \\
            ~~&\implies~~ f_i(\x) \geq \alpha'' \sum_{j \in [\kfit]} f_j(\x) \\
            ~~&\implies~~ f_i(\x) \geq \alpha'' \max_{ \substack{j \in [\kfit] \\ j \neq i}} f_j(\x).
    \end{align*}
    Combining the last inequality with \eqref{eq:geo_equiv}, we obtain
    \begin{equation}\label{eqn:proof_lemma_soft.1}
    \begin{aligned}
        \asso_i(\x) \geq \alpha''
            &\qquad \implies \qquad
            \x \in \softvor_i^{\alpha'}  \quad\text{for}~~\alpha' = \frac{\sigma}{\|\bbeta_i - \bbeta_j \|} \cdot \log \left( \frac{1}{\alpha''} \right).
    \end{aligned}
    \end{equation}
    Choosing $\alpha'' = \frac{1}{\kfit} \exp \left( - \alpha \frac{ \| \bbeta_i - \bbeta_j \| }{\sigma} \right)$ yields $\alpha' = \alpha + \frac{\sigma}{\|\bbeta_i - \bbeta_j \|} \log \kfit$.

    \bigskip
    \textit{\underline{(2) Proof of Claim 2.}}
    By the first claim of this lemma, i.e., \eqref{eq:geometry1}, we observe that 
    \begin{align*}
        &\x \in \softvor_i^{\alpha} \cap \softvor_j^{\alpha}\\
            &\implies~
            \min\left\{ \asso_{i}(\x),\asso_{j}(\x)\right\} \geq \frac{1}{\kfit} \exp \left( - \alpha \frac{ \| \bbeta_i - \bbeta_j \| }{\sigma} \right)\\
            &\implies~\asso_{i}(\x)\asso_{j}(\x)\geq \frac{1}{\kfit^2} \exp \left( - 2\alpha \frac{ \| \bbeta_i - \bbeta_j \| }{\sigma} \right).
    \end{align*}
    Thus, it follows from the definition of the set $\goodset_{ij}^{\delta}$ in \eqref{eq:goodset} that
    \[
        \softvor_{i}^{\alpha}\cap\softvor_{j}^{\alpha}\subseteq\goodset_{ij}^{\delta} 
        \qquad\text{for} ~~ \delta = \frac{1}{\kfit^2} \exp \left( - 2\alpha \frac{ \| \bbeta_i - \bbeta_j \| }{\sigma} \right).
    \]
    As $\max\{\asso_i(\x), \asso_j(\x)\} \leq 1$, $\asso_i(\x)\asso_j(\x) \geq \delta$ implies $\min\{\asso_i(\x), \asso_j(\x) \} \geq \delta$. 
    Therefore, by the same argument as in \eqref{eqn:proof_lemma_soft.1}, we get
    \begin{align*}
        \x \in \goodset_{ij}^{\delta} 
            ~~\implies~~ \x \in \softvor_i^{\delta'} \cap \softvor_j^{\delta'}
    \end{align*}
    where
    \begin{align*}
        \delta' = \frac{\sigma}{\|\bbeta_i - \bbeta_j \|} \cdot \log \left( \frac{1}{\delta} \right).
    \end{align*}

    \bigskip
    \textit{\underline{(3) Proof of Claim 3.}}
    Finally, we note that
    \begin{align*}
        &\x\in\softbdr_{ij}^{\alpha}\cap\softvor_{j}^{\alpha}\\
            &\implies~ 
                \x \in \softvor_{j}^{\alpha}
                ~~\&~~ \left\langle \x-\bbetabar_{ij},~\frac{\bbeta_{i}-\bbetabar_{ij}}{\|\bbeta_{i}-\bbetabar_{ij}\|}\right\rangle \ge-\alpha\\
            &\implies~\asso_{j}(\x)\ge \frac{1}{\kfit} \exp \left( - \alpha \frac{ \| \bbeta_i - \bbeta_j \| }{\sigma} \right)
                \quad\&\quad \asso_i(\x) \geq \asso_j(\x) \cdot \exp \left( - \alpha \frac{ \| \bbeta_i - \bbeta_j \| }{\sigma} \right)
    \end{align*}
    by \eqref{eq:geometry1} and \eqref{eq:geo_equiv}. Therefore, 
    \begin{align*}
        \x\in\softbdr_{ij}^{\alpha}\cap\softvor_{j}^{\alpha}
            &\qquad\implies\qquad 
            \asso_i(\x) \asso_j(\x) \geq \frac{1}{k^2} \exp \left( - 3 \alpha \frac{ \| \bbeta_i - \bbeta_j \| }{\sigma} \right).
    \end{align*}
\end{proof}

\subsection{Helper Lemmas for the Proof of Proposition \ref{prop:small_bdr1}}\label{sec:prep_proof_proposition.2}
\paragraph{Gaussian lemmas}
Let $\phi_{\sigma}: \real \to \real$ denote the probability density function of the Gaussian distribution $\mathcal{N}(0,\sigma^2)$, i.e., $\phi_{\sigma}(x) = \frac{1}{\sqrt{2\pi \sigma^2}} e^{-x^2/(2\sigma^2)}$ for all $x \in \real$. 
We denote by $\phi = \phi_1$, omitting the subscript when $\sigma = 1$. 
We observe that $\phi_{\sigma}(x) = \frac{1}{\sigma} \phi(x/\sigma)$ for all $\sigma \in \real_+$ and all $x \in \real$.

Here we state and prove a simple technical lemma that will be used later in our analysis.
\begin{lemma}\label{lem:gaussian_property}
    For any $\sigma \in \real_+$, any $\tau \in \real_+$, and any $t_1, t_2 \in \real$ such that $0 \le t_1 \le t_2 \le \infty$,
    \begin{align*}
        \int_{t_1}^{t_2}\phi_{\sigma}(z)\ddup z  
            &\le \left( \sqrt{2\pi} + 1 \right) \cdot \max\left\{ 1,~\frac{\sigma}{\tau}\right\} 
            \int_{t_1}^{t_2}\indic\left\{ z - t_1 \le \tau\right\} \phi_{\sigma}(z) \,\ddup z.
    \end{align*}
\end{lemma}

\begin{proof}[Proof of Lemma \ref{lem:gaussian_property}]
    If $\tau \geq t_2 - t_2$, then the conclusion trivially follows. 
    In the rest of this proof, we assume $\tau < t_2 - t_1$ and consider two cases separately.
    
    \bigskip
    \textit{\underline{Case 1:} $\tau\ge \sigma$.}
    In this case, we observe that 
    \begin{align*}
        \int_{t_1 + \tau}^{t_2}\phi_{\sigma}(z)\ddup z 
            &= \int_{\frac{t_1 + \tau}{\sigma}}^{\frac{t_2}{\sigma}}\phi (z)\ddup z \\
            & \stackrel{(a)}{\le} \sqrt{2\pi}\cdot\phi \left(\frac{t_1 + \tau}{\sigma} \right) \\
            & \stackrel{(b)}{\le} \frac{\tau}{\sigma} \cdot \sqrt{2\pi} \cdot\phi \left(\frac{t_1 + \tau}{\sigma} \right) \\
            & \stackrel{(c)}{\le} \sqrt{2\pi} \int_{\frac{t_1}{\sigma}}^{\frac{t_1+\tau}{\sigma}}\phi(z)\ddup z \\
            &= \sqrt{2\pi} \int_{t_1}^{t_1+\tau} \phi_{\sigma} (z) \ddup z,
    \end{align*}
    where (a) follows from Lemma \ref{lem:gaussian_tail}, (b) is due to $\tau\ge \sigma$, and (c)  holds because $\phi$ is non-increasing on $\left[\frac{t_1}{\sigma}, \frac{t_1+\tau}{\sigma}\right]$.
    It follows that 
    \begin{align*}
        \int_{t_1}^{t_2}\phi(z)\ddup z 
            & =\int_{t_1}^{t_1+\tau}\phi(z)\ddup z+\int_{t_1+\tau}^{t_2}\phi(z)\ddup z\\
            & \le \left( \sqrt{2\pi} + 1 \right) \int_{t_1}^{t_1+\tau}\phi(z)\ddup z\\
            & =\left( \sqrt{2\pi} + 1 \right)  \int_{t_1}^{t_2}\indic\left\{ z - t_1 \le \tau\right\} \phi(z)\ddup z.
    \end{align*}

    \bigskip
    \textit{\underline{Case 2:} $\tau < \sigma$.} In this case, we follow a similar argument as above, observing that
    \begin{align*}
        \int_{t_1+\tau}^{t_2}\phi_{\sigma}(z)\ddup z 
            &= \int_{\frac{t_1+\tau}{\sigma}}^{\frac{t_2}{\sigma}}\phi(z)\ddup z\\
            & \le\sqrt{2\pi}\cdot\phi \left( \frac{t_1 + \tau}{\sigma} \right) \\
            & \le \left( \sqrt{2\pi} + 1 - \frac{\tau}{\sigma} \right) \cdot \phi \left( \frac{t_1 + \tau}{\sigma} \right) \\
            &\leq \left( \sqrt{2\pi} + 1 - \frac{\tau}{\sigma} \right) \cdot \frac{\sigma}{\tau} \cdot \int_{\frac{t_1}{\sigma}}^{\frac{t_1+\tau}{\sigma}}\phi(z)\ddup z\\ 
            &= \left( \big(\sqrt{2\pi} + 1\big) \frac{\sigma}{\tau} - 1 \right) \cdot \int_{t_1}^{t_1+\tau}\phi_{\sigma}(z)\ddup z.
    \end{align*}
    Consequently, it follows that 
    \begin{align*}
        \int_{t_1}^{t_2}\phi_{\sigma}(z)\ddup z 
            &=\int_{t_1}^{t_1+\tau}\phi_{\sigma}(z)\ddup z+\int_{t_1+\tau}^{t_2}\phi_{\sigma}(z)\ddup z\\
            &\le \frac{\sigma}{\tau} \cdot \left( \sqrt{2\pi} + 1 \right) \cdot \int_{t_1}^{t_1+\tau}\phi_{\sigma}(z)\ddup z.
    \end{align*}
\end{proof}

\paragraph{A useful geometric lemma}
Our proof of Proposition \ref{prop:small_bdr1} relies on the following geometric lemma.
\begin{lemma}[Controlling volume by intersection]\label{lem:vol_intersect}
    Let $\B \in \real^{d \times k}$ be an arbitrary ordered set of vectors. 
    For any $(s, i) \in [\ktrue] \times [\kfit]$ and any $\big(\alpha_j \in \real_+: j \in [\kfit]\setminus\{i\} \big)$, if $\bthetastar_{s}\notin\intr\vor_{i}$, then 
    \begin{equation}\label{eq:vol_intersect}
    \begin{aligned}
        \P_{s}\left(\vor_{i}\right) 
            & \le
                \left(\sqrt{2\pi} + 1 \right) \sum_{j\in [\kfit]\setminus\{i\}} \max\left\{ 1, ~\frac{1}{\alpha_j} \right\} \cdot  \P_{s}\left(\softvor_{i}^{\alpha_j}\cap\softbdr_{ij}^{\alpha_j}\right).
    \end{aligned}
    \end{equation}
\end{lemma}

\begin{proof}[Proof of Lemma \ref{lem:vol_intersect}]
    We present this proof in four steps. 
    In Step 1, we introduce some notation for the convenience of the later steps. 
    In Step 2, we derive an useful expression for the probability $\P_s(\vor_i)$; see \eqref{eq:vor_bound}. 
    In Step 3, we establish a lower bound for $\P_s \big( \softvor_i^{\alpha} \cap \softbdr_{ij}^{\alpha} \big)$ for $j \in [\kfit] \setminus \{i\}$,  cf. \eqref{eq:bdr_bound}. 
    Finally, in Step 4, we conclude the proof.

    \bigskip
    \textit{\underline{Step 1.}}
    We may assume that $\bthetastar_{s}=0$ without loss of generality, due to the Euclidean invariance; see Remark \ref{rem:coordinate}. 
    Since $\bthetastar_{s}\notin\intr\vor_{i}$ and $\vor_{}$ is a convex set, there exists a $\v\in\real^{d}$ such that $\left\langle \v,\x\right\rangle \ge0$ for all $\x\in\vor_{j}$ by the Separating Hyperplane Theorem. 
    Again by the Euclidean invariance (rotational invariance, in particular), we may assume that $\v=\e_{1}$. 
    For each point $\x\in\intr\vor_{i}$, the ray $\left\{ \x-b\e_{1} \in \real^d : b\ge0\right\} $ intersects a facet $F$ of the polyhedron $\vor_{i}$ at a unique point, which we denote by $\y = \y(\x)$, where $F\subseteq\bdr_{ij}$ for some $j\in[\kfit]$ with $\bbeta_{i}\neq\bbeta_{j}$; 
    we let $j_{*} = j_{*}(\x)$ denote such $j$ (if there exist multiple such $j$'s, we pick the smallest).
    
    It is clear that $\y$ and $j_*$ are independent of the first coordinate of $\x$, hence, we can write $\y=\y(\x_{2}^{d})$ and $j_*=j_*(\x_{2}^{d})$, where $\x_{2}^{d}\coloneqq(x_{2},\ldots,x_{d})\in\real^{d-1}$. 
    As a result, every $\x\in\intr\vor_{i}$ can be uniquely expressed as $\x=\y(\x_{2}^{d})+b \cdot \e_{1}$ for some $b = b(\x)\ge0$. 
    Note that for all $\x \in \vor_i$,
    \[
        x_{1}\ge y_{1}(\x_{2}^{d})=\left\langle \e_{1},\y(\x_{2}^{d})\right\rangle \ge0
    \]
    by construction of $\y$ and by the separating hyperplane theorem. 
    For each $i \in [\kfit]$, we let 
    \[
        \mathcal{L}_{i}\coloneqq\left\{ j\in[\kfit]\setminus\{i\}:j=j_*(\x_{2}^{d})\text{ for some \ensuremath{\x\in\intr\vor_{i}}}\right\}.
    \]
    See Figure~\ref{fig:vol_intersect} for an illustration of these notations.

    \begin{figure}[t]
    \begin{centering}
    \includegraphics[width=0.6\linewidth]{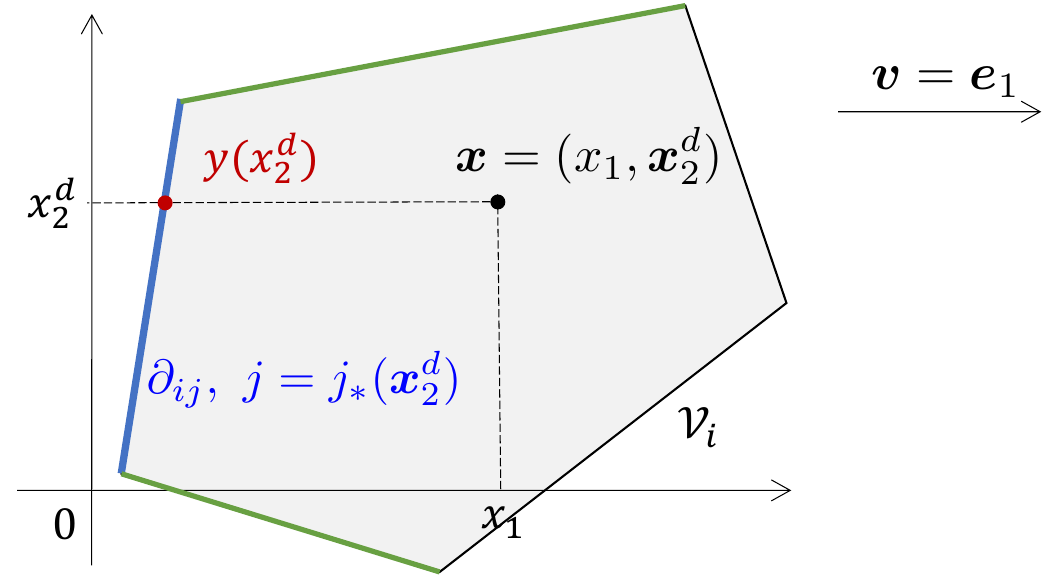}
    \par\end{centering}
        \caption{Illustration for the proof of Lemma~\ref{lem:vol_intersect}.
        The figure shows the polyhedral Voronoi cell $\protect\vor_{i}$ and
        the direction $\protect\v=\protect\e_{1}$ that defines the separating
        hyperplane between $\protect\vor_{i}$ and $\protect\bzero$. For
        each point $\protect\x=(x_{1},\protect\x_{2}^{d})^{\top}\in\protect\intr\protect\vor_{i}$,
        the ray $\left\{ \protect\x-b\protect\e_{1},b\ge0\right\} $
        intersects a facet $F$ of $\protect\vor_{i}$ at a unique point $\protect\y=\protect\y(\protect\x_{2}^{d})$,
        where $F\subseteq\protect\bdr_{ij}$ for some $j=j_*(\protect\x_{2}^{d})\in[\protect\kfit]$.
        The set $\mathcal{L}_{i}$ indexes the Voronoi boundaries colored in blue and green.}
        \label{fig:vol_intersect}
    \end{figure}

    \bigskip
    \textit{\underline{Step 2.}}
    We overload notation and let $\phi_{\sigma}$ denote the probability density function of $\mathcal{N}(0, \sigma^2 \Id_d)$ for any $d \in \NN_+$.
    As we assumed $\bthetastar_s = 0$, we can write 
    \begin{align}
        \P_{s}(\vor_{i})
            & =\int_{\real^{d}}\indic\left\{ \x\in\vor_{i}\right\} \cdot \phi_{\sigma}(\x)~\ddup\x \nonumber\\
            & =\sum_{j\in\mathcal{L}_{i}}\int_{\real^{d}}\indic\left\{ \x\in\vor_{i}~\&~ j_*(\x_{2}^{d})=j \right\} \cdot \phi_{\sigma}(\x)~\ddup\x  \nonumber\\
            & =\sum_{j\in\mathcal{L}_{i}}\int_{\real^{d-1}}\left[\int_{y_{1}\left(\x_{2}^{d}\right)}^{\infty}\indic\left\{ \x\in\vor_{i}\right\} \cdot \phi_{\sigma}(x_{1})~\ddup x_{1}\right] \cdot \indic\left\{ j_*(\x_{2}^{d})=j\right\}\cdot \phi_{\sigma}(\x_{2}^{d})~\ddup\x_{2}^{d}    \label{eqn:vor_expansion}
    \end{align}
    by the definition of $j_*(\cdot)$ and $\cL_i$, as well as the Fubini's theorem. 

    Define the quantity $u_i^*(\x_{2}^{d})\coloneqq\sup\left\{ x_{1} \in \real :(x_{1},\x_{2}^{d})^{\top}\in\vor_{i}\right\}$ with the convention that $y_{1}(\x_{2}^{d})=u^*_i(\x_{2}^{d})=0$ when $\left\{ (x_{1},\x_{2}^{d})^{\top}:x_{1}\in\real\right\} \cap\vor_{i}=\emptyset$. 
    Then we observe that $0 \leq y_1(\x_2^d) \leq u_i^*(\x_2^d) \leq \infty$. 
    Continuing from the equation \eqref{eqn:vor_expansion}, we have
    \begin{align}
        \P_{s}(\vor_{i}) 
            & =\sum_{j\in\mathcal{L}_{i}}\int_{\real^{d-1}}\left[\int_{y_{1}\left(\x_{2}^{d}\right)}^{u_i^*(\x_{2}^{d})}\phi_{\sigma}(x_{1})~\ddup x_{1}\right] \cdot \indic\left\{ j_*(\x_{2}^{d})=j\right\} \cdot \phi_{\sigma}(\x_{2}^{d})~\ddup\x_{2}^{d}.    \label{eq:vor_bound}
    \end{align}

    \bigskip
    \textit{\underline{Step 3.}}
    For each $j\in\mathcal{L}_{i}$, we may write using a similar argument as in Step 2 that for any $\alpha \geq 0$,
    \begin{align}
        \P_{s}\left(\softvor_{i}^{\alpha}\cap\softbdr_{ij}^{\alpha}\right) 
            &\stackrel{(a)}{\ge}\P_{s}\left(\vor_{i}\cap\softbdr_{ij}^{\alpha}\right)   \nonumber\\
            &\stackrel{(b)}{=}\sum_{j'\in\mathcal{L}_{i}}\int_{\real^{d-1}}\left[\int_{y_{1}\left(\x_{2}^{d}\right)}^{\infty}\indic\left\{ \x\in\vor_{i} \cap \softbdr_{ij}^{\alpha} \right\} \cdot\phi_{\sigma}(x_{1})~\ddup x_{1}\right] \cdot \indic\left\{ j_*(\x_{2}^{d})=j'\right\} \cdot \phi_{\sigma}(\x_{2}^{d})~\ddup\x_{2}^{d}   \nonumber\\
            &=\sum_{j'\in\mathcal{L}_{i}}\int_{\real^{d-1}}\left[\int_{y_{1}\left(\x_{2}^{d}\right)}^{u_i^*(\x_{2}^{d})}\indic\left\{ \x\in\softbdr_{ij}^{\alpha}\right\} \cdot \phi_{\sigma}(x_{1})~\ddup x_{1}\right] \cdot \indic\left\{ j_*(\x_{2}^{d})=j' \right\} \cdot \phi_{\sigma}(\x_{2}^{d})~\ddup\x_{2}^{d}
                    \label{eqn:intersection.1}
    \end{align}
    where the inequality (a) holds because $\vor_i \subseteq \softvor_i^{\alpha}$ and (b) is due to \eqref{eqn:vor_expansion} and \eqref{eq:vor_bound}.

    Next, we observe that for each $j\in\mathcal{L}_{i}$, the following holds: 
    \begin{equation}\label{eq:soft_bdr_suff_cond}
    \begin{aligned}
        j_*(\x_{2}^{d})=j\;\;\text{and}\;\;x_{1}-y_{1}\left(\x_{2}^{d}\right)\le\alpha \sigma
        \qquad
        \implies\qquad
            \x\in\softbdr_{ij}^{\alpha}.
    \end{aligned}
    \end{equation}
    \begin{quote}
        \textit{Proof of Claim \eqref{eq:soft_bdr_suff_cond}}: 
        Fix an $\x \in \cV_i$ with $j_*(\x_{2}^{d})=j$ and $x_{1}-y_{1}\left(\x_{2}^{d}\right)\le \alpha \sigma$.
        Since $\y(\x_{2}^{d})\in\bdr_{ij}$, we have $\left\langle \y(\x_{2}^{d})-\bbetabar_{ij},\bbeta_{i}-\bbetabar_{ij}\right\rangle =0$ by definition of $\bdr_{ij}$; see \eqref{eq:bdr}. 
        Therefore, 
        \begin{align*}
            \left|\left\langle \x-\bbetabar_{ij},\bbeta_{i}-\bbetabar_{ij}\right\rangle \right| 
                & =\left|\left\langle \x-\y(\x_{2}^{d}),\bbeta_{i}-\bbetabar_{ij}\right\rangle \right|\\
                & \le\left\Vert \x-\y(\x_{2}^{d})\right\Vert \cdot\left\Vert \bbeta_{i}-\bbetabar_{ij}\right\Vert \\
                & =\left|x_{1}-y_{1}(\x_{2}^{d})\right|\cdot\left\Vert \bbeta_{i}-\bbetabar_{ij}\right\Vert \\
                & \le \alpha \sigma \cdot\left\Vert \bbeta_{i}-\bbetabar_{ij}\right\Vert.
        \end{align*}
        The above inequality implies that $\x\in\softbdr_{ij}^{\alpha}$ by definition of $\softbdr_{ij}^{\alpha}$, cf. \eqref{eqn:alpha_enlarged}. 
    \end{quote}

    By the logical implication \eqref{eq:soft_bdr_suff_cond}, we have the following inequality for all $\alpha \geq 0$:
    \begin{equation}\label{eqn:intersection.2}
    \begin{aligned}
        \indic\left\{ \x\in\softbdr_{ij}^{\alpha}\right\}
            &\ge
            \indic\left\{ x_{1} - y_{1}\left(\x_{2}^{d}\right) \le \alpha \sigma \right\} \cdot \indic\left\{ j_*(\x_{2}^{d})=j\right\}.
    \end{aligned}
    \end{equation}
    Combining \eqref{eqn:intersection.1} and \eqref{eqn:intersection.2}, we obtain
    \begin{equation}\label{eq:bdr_bound}
    \begin{aligned}
        \P_{s}\left(\softvor_{i}^{\alpha}\cap\softbdr_{ij}^{\alpha}\right)
            &\ge     \int_{\real^{d-1}}\left[\int_{y_{1}\left(\x_{2}^{d}\right)}^{u_i^*(\x_{2}^{d})}\indic\left\{ x_{1} - y_{1}\left(\x_{2}^{d}\right) \le \alpha \sigma \right\} \cdot \phi_{\sigma}(x_{1})~\ddup x_{1}\right] \cdot \indic\left\{ j_*(\x_{2}^{d})=j \right\} 
            \cdot \phi_{\sigma}(\x_{2}^{d})~\ddup\x_{2}^{d}.
    \end{aligned}
    \end{equation}

    \bigskip
    \textit{\underline{Step 4.}}
    By Lemma \ref{lem:gaussian_property}, we observe that for all $\alpha > 0$,
    \begin{align*}
        \int_{y_{1}\left(\x_{2}^{d}\right)}^{u_i^*(\x_{2}^{d})} \phi_{\sigma}(x_{1})~\ddup x_{1}
            &\leq \left(\sqrt{2\pi} + 1 \right) \cdot \max\left\{ 1, ~\frac{1}{\alpha} \right\} \cdot \int_{y_{1}\left(\x_{2}^{d}\right)}^{u_i^*(\x_{2}^{d})}\indic\left\{ x_{1} - y_{1}\left(\x_{2}^{d}\right) \le \alpha \sigma \right\} \cdot \phi_{\sigma}(x_{1})~\ddup x_{1}.
    \end{align*}
    Combining this inequality with \eqref{eq:vor_bound} and the inequality \eqref{eq:bdr_bound}, we have the following inequality hold for any choice of $\alpha_j \in \real_+$ for $j \in [\kfit]\setminus\{i\}$:
    \begin{align*}
        \frac{1}{\sqrt{2\pi} + 1 } \cdot \P_{s}(\vor_{i})
            &\leq \sum_{j\in\mathcal{L}_{i}} \max\left\{ 1, ~\frac{1}{\alpha_j} \right\} \cdot  \P_{s}\left(\softvor_{i}^{\alpha_j}\cap\softbdr_{ij}^{\alpha_j}\right) \\
            &\leq \sum_{j\in [\kfit]\setminus\{i\}} \max\left\{ 1, ~\frac{1}{\alpha_j} \right\} \cdot  \P_{s}\left(\softvor_{i}^{\alpha_j}\cap\softbdr_{ij}^{\alpha_j}\right).
    \end{align*}
\end{proof}

\subsection{Completing the Proof of Proposition \ref{prop:small_bdr1}}\label{sec:complete_proof_proposition2}

\begin{proof}[Proof of Proposition \ref{prop:small_bdr1}]
    We begin by proving the first claim, and then use it to prove the second claim.
    
    \bigskip
    \textit{\underline{(1) Proof of Claim 1.}}
    We choose an arbitrary sequence $\big( \alpha_j \in \real_+: j \in [\kfit]\setminus\{i\} \big)$, and let $\delta_{ij}^{\alpha} \coloneqq \frac{1}{\kfit^2} \exp \left( - 3\alpha_j \frac{ \| \bbeta_i - \bbeta_j \| }{\sigma} \right)$ for all $j \in [\kfit]\setminus\{i\}$. Then we obtain the following upper bound
    \begin{align*}
        \frac{1}{\sqrt{2\pi} + 1 } \cdot \P_s\left( \vor_i \right)
            &\stackrel{(a)}{\leq} \sum_{j\in [\kfit]\setminus\{i\}} \max\left\{ 1, ~\frac{1}{\alpha_j} \right\} \cdot \P_{s}\left(\softvor_{i}^{\alpha_j}\cap\softbdr_{ij}^{\alpha_j}\right)      \\
            &= \sum_{j\in [\kfit]\setminus\{i\}} \max\left\{ 1, ~\frac{1}{\alpha_j} \right\} \cdot \frac{1}{\delta_{ij}^{\alpha}} 
                \cdot \E_{s}\left[ \delta_{ij}^{\alpha} \cdot \indic\left\{ \softvor_{i}^{\alpha}\cap\softbdr_{ij}^{\alpha} \right\} \right]\\
            &\leq\sum_{j\in [\kfit]\setminus\{i\}} \max\left\{ 1, ~\frac{1}{\alpha_j} \right\} \cdot \frac{1}{\delta_{ij}^{\alpha}} \cdot \E_{s}\left[ \Asso_i \Asso_j  \right]     \\
            &=\kfit^2 \sum_{j\in [\kfit]\setminus\{i\}} \max\left\{ 1, ~\frac{1}{\alpha_j} \right\} \cdot \E_{s}\left[ \Asso_i \Asso_j  \right] 
                \cdot  \exp \left( 3\alpha_j \frac{ \| \bbeta_i - \bbeta_j \| }{\sigma} \right).
    \end{align*}
    where (a) is obtained by Lemma \ref{lem:vol_intersect}, and (b) is due to Lemma \ref{lem:geometry}, Claim 3.

    \bigskip
    \textit{\underline{(2) Proof of Claim 2.}}
    Summing $\E_s\left[ \Asso_i \Asso_j \right]$ over $j \in [\kfit] \setminus \{i\}$, we obtain that
    \begin{align*}
        \sum_{j \in [\kfit]\setminus\{i\}} \E_s\left[ \Asso_i \Asso_j \right]
            & = \E_s\left[ \Asso_i (1 - \Asso_i) \right]\\
            & = \E_{s}\left[\Asso_{i} \cdot \indic\left\{ \vor_{i}\right\} \right]-\E_{s}\left[\Asso_{i}^{2} \cdot \indic\left\{ \vor_{i}\right\} \right]
                 +\E_{s}\left[\Asso_{i}(1-\Asso_{i}) \cdot \indic\left\{ \vor_{i}^{\complement}\right\} \right]\\ 
            &\overset{\text{(a)}}{\geq} \frac{1}{2}\E_{s}\left[\Asso_{i} \cdot \indic\left\{ \vor_{i}\right\} \right]-\P_{s}\left(\vor_{i}\right)+\frac{1}{2}\E_{s}\left[\Asso_{i} \cdot \indic\left\{ \vor_{i}^{\complement}\right\} \right]\\
            & =\frac{1}{2}\E_{s}\left[\Asso_{i}\right]-\P_{s}\left(\vor_{i}\right)
    \end{align*}
    where the inequality (a) holds because $\x\in\vor_{i}^{\complement}$ implies $\asso_{i}(\x)<\frac{1}{2}$. 
    Then it follows that
    \begin{align*}
        \E_{s}\left[\Asso_{i}\right]
            &\leq 2 \left\{ \sum_{j \in [\kfit]\setminus\{i\}} \E_s\left[ \Asso_i \Asso_j \right] + \P_{s}\left(\vor_{i}\right)  \right\}\\
            &\leq 3 \left(\sqrt{2\pi} + 1 \right) \cdot \kfit^2 \sum_{j\in [\kfit]\setminus\{i\}} \max\left\{ 1, ~\frac{1}{\alpha_j} \right\} \cdot \E_{s}\left[ \Asso_i \Asso_j  \right] 
                \cdot \exp \left( 3\alpha_j \frac{ \| \bbeta_i - \bbeta_j \| }{\sigma} \right).
    \end{align*}
    The last inequality follows from Claim 1 of this proposition.
\end{proof}

\section{Deferred Proof of Proposition \ref{prop:exclusive}}\label{sec:proof_proposition}
\begin{proof}[Proof of Proposition \ref{prop:exclusive}]
    In this proof, we prove the three claims one by one. 

    \bigskip
    \textit{\underline{(1) Proof of Claim 1.}}
    Fix an arbitrary $i \in [\kfit]$ that satisfies $\setA_i^{\delta} \neq \emptyset$. 
    Let $\mathcal{S}_i := \left\{ j \in [\kfit]: \bbeta_j = \bbeta_i \right\}$. 
    Then we may write
    \begin{equation}\label{eq:asso_i_expression}
        \E_{s}\left[\Asso_{i}\right]
            =\frac{1}{\left| \mathcal{S}_i \right|}\sum_{j \in \mathcal{S}_i}\E_{s}\left[\Asso_{j}\right]
            =\frac{1}{\left| \mathcal{S}_i \right|}\left(1-\sum_{j\in [\kfit] \setminus \mathcal{S}_i }\E_{s}\left[\Asso_{j}\right]\right).
    \end{equation}
    
    By definition of the set $\setA_i^{\delta}$, we observe that for all pair $(s,j) \in \setA_i^{\delta} \times \big( [\kfit] \setminus \mathcal{S}_i \big)$,
    \begin{equation}\label{eq:small_bdr2_assmp}
    \begin{aligned}
        \bthetastar_{s}\notin\intr\vor_{j}
            \quad\text{and}\quad
        \max_{j'\in[\kfit]\setminus\{j\}}  \frac{ \| \bbeta_j - \bbeta_{j'} \| }{\sigma} \cdot \E_{s}\left[\Asso_{j}\Asso_{j'}\right]<\delta.
    \end{aligned}
    \end{equation}
    It follows from Corollary~\ref{cor:small_bdr} (claim 2) and \eqref{eq:small_bdr2_assmp} that for all $(s,j) \in \setA_i^{\delta} \times \big( [\kfit] \setminus \mathcal{S}_i \big)$,
    \begin{align}
        \E_{s}\left[\Asso_{j}\right] 
            &\leq 
                9 \left(\sqrt{2\pi} + 1 \right) \cdot \kfit^2 \cdot \sum_{j'\in [\kfit]\setminus\{j\}} \frac{ \| \bbeta_i - \bbeta_j \| }{\sigma} \cdot \E_{s}\left[ \Asso_i \Asso_j  \right]
                    \nonumber\\
            &\leq
                9 \left(\sqrt{2\pi} + 1 \right) \cdot \kfit^2 (\kfit - 1) \cdot \delta
                \nonumber\\
            &\leq  
                9 \left(\sqrt{2\pi} + 1 \right) \cdot \kfit^3 \cdot \delta.  \label{eq:asso_j_upper_bound}
    \end{align}
    Plugging \eqref{eq:asso_j_upper_bound} into \eqref{eq:asso_i_expression}, we obtain that for all $s \in \setA_i^{\delta}$,
    \begin{equation}\label{eq:asso_i_bound}
        \E_{s}\left[\Asso_{i}\right]
            \geq \frac{1}{\left| \mathcal{S}_i \right|} \left\{ 1 - 9 \left(\sqrt{2\pi} + 1 \right) \cdot \kfit^4 \cdot \delta  \right\}.
    \end{equation}

    Now we assume there exists $i' \in \mathcal{S}_i $ such that $i' \neq i$. Then we have 
    \begin{align}
        \E_{s}[\Asso_{i}\Asso_{i'}]
            &=\E_{s}[\Asso_{i}^{2}]
            \ge \E_{s}[\Asso_{i}]^{2} \nonumber\\
            &\ge \frac{1}{\left| \mathcal{S}_i \right|^2} \Big\{ 1 - 9 \left(\sqrt{2\pi} + 1 \right) \cdot  \kfit^4 \cdot \delta \Big\}^2    \nonumber\\
            &\ge \frac{1}{\kfit^2 } \Big\{ 1 - 9 \left(\sqrt{2\pi} + 1 \right) \cdot  \kfit^4 \cdot \delta \Big\}^2.
                \label{eqn:product_lower_bound}
    \end{align}
    Letting $\varphi_* := 9 \left(\sqrt{2\pi} + 1 \right) $ for a shorthand, we observe that if $\delta \leq \frac{1}{2 \varphi_* \kfit^4}$, then it follows from \eqref{eqn:product_lower_bound} that
    \begin{align*}
        \E_{s}[\Asso_{i}\Asso_{i'}]
            &\geq  \frac{1}{\kfit^2 } \Big\{ 1 - \varphi_* \cdot  \kfit^4 \cdot \delta \Big\}^2
            \geq \frac{3}{4} \frac{1}{\kfit^2}
            \geq \frac{1}{2 \varphi_* \kfit^4}
            \geq \delta,
    \end{align*}
    which contradicts the assumption that $s\in \setA_i^{\delta}$. 
    Consequently, $\mathcal{S}_i = \{ i \}$, and the claim is proved.

    \bigskip
    \textit{\underline{(2) Proof of Claim 2.}}
    Now that $\mathcal{S}_i = \{ i \}$, it immediately follows from \eqref{eq:asso_i_bound} that for all $s \in \setA_i^{\delta}$,
    \[
        1 - \E_{s}\left[\Asso_{i}\right]
            \leq 9 \left(\sqrt{2\pi} + 1 \right) \cdot \kfit^4 \cdot \delta.
    \]

    \bigskip
    \textit{\underline{(3) Proof of Claim 3.}}
    It follows from Corollary~\ref{cor:small_bdr} (claim 1) and \eqref{eq:small_bdr2_assmp} that for all $s \in \setA_i^{\delta}$ and all $j \in [\kfit] \setminus \{ i\}$ (note that $\mathcal{S}_i = \{i\}$ by claim 1 proved above),
    \[
        \P_s(\vor_j) \leq 3\left(\sqrt{2\pi} + 1 \right) \cdot \kfit^4 \cdot \delta
    \]
    by the same argument as in \eqref{eq:asso_j_upper_bound}. Therefore, for all $s \in \setA_i^{\delta}$,
    \[
        \P_s \big( \vor_i^c \big)
            = \sum_{j \in [\kfit]\setminus\{i\}} \P_s \big( \vor_j \big)
            \leq  3\left(\sqrt{2\pi} + 1 \right) \cdot \kfit^4 \cdot \delta.
    \]
\end{proof}
\section{Deferred Proof of Proposition \ref{prop:proposition1}}\label{sec:proof_proposition.1}

\subsection{A Helper lemma for the Proof of Proposition \ref{prop:proposition1}}\label{sec:prep_proof_proposition.1}
Recall from Section \ref{sec:characterization} that any local minimum $\B$ of $L$ must satisfy the first-order stationary condition \eqref{eq:station_cond} and the second-order optimality condition $\nabla^{2}L(\bbeta)\succeq0$, where the Hessian $\nabla^{2}L$ can be computed using the expression \eqref{eq:hessian} as stated in Lemma \ref{lem:hessian_NLL}.
Based on this fact, we state a lemma that is useful in our proof of Proposition \ref{prop:proposition1}.

\begin{lemma}\label{lem:2nd_implication}
    Let $\B \in \real^{d \times \kfit}$ be a local minimum of $L$. 
    For any $i, j \in [\kfit]$ such that $i \neq j$, and any $\v_1, \v_2 \in \real^d$, 
    \begin{equation}\label{eq:2nd_ord_cond}
    \begin{aligned}
        \E_{*}\left[\Asso_{i}\Asso_{j}\big(\left\langle \bbeta_{i}-{\sf x},\v_1\right\rangle -\left\langle \bbeta_{j}-{\sf x},\v_2\right\rangle \big)^{2}\right]
            &\leq  \sigma^2 \cdot \left\Vert \v_1\right\Vert ^{2} \cdot \E_{*}\left[\Asso_{i}\right] +  \sigma^2 \cdot \left\Vert \v_2\right\Vert ^{2} \cdot \E_{*}\left[\Asso_{j}\right].
    \end{aligned}
    \end{equation}
\end{lemma}

\begin{proof}[Proof of Lemma \ref{lem:2nd_implication}]
    First of all, $\nabla^2 L(\B) \succeq 0$ because $\B$ is a local minimum of $L$. 
    Thus, $\v^{\top} \nabla^2 L(\B) \v \geq 0$ for all $\v \in \real^{d \kfit}$. 
    We choose $\v$ to be the flattened vector of $\V \in \real^{d \times \kfit}$ such that $\V_i = \v_1$ and $\V_j = \v_2$, all other columns being $0$. 
    Then we obtain by applying Lemma \ref{lem:hessian_NLL} that
    \begin{align}
        \sigma^4 \cdot \v^{\top} \nabla^2 L(\B) \v 
            &= \sigma^4 \cdot \v_{i}^{\top}\left[\frac{\partial^{2}}{\partial\bbeta_{i}\partial\bbeta_{i}}L(\B)\right]\v_{i} 
                + \sigma^4 \cdot \v_{j}^{\top}\left[\frac{\partial^{2}}{\partial\bbeta_{j}\partial\bbeta_{j}}L(\B)\right]\v_{j} 
                +2 \sigma^4 \cdot \v_{i}^{\top}\left[\frac{\partial^{2}}{\partial\bbeta_{i}\partial\bbeta_{j}}L(\B)\right]\v_{j}    \nonumber\\
            &= \E_{*}\left[(\Asso_{i}-1)\Asso_{i}\left\langle \bbeta_{i}-{\sf x},\v_1\right\rangle ^{2} + \sigma^2 \cdot \Asso_{i}\left\Vert \v_1\right\Vert^{2}\right]
                +\E_{*}\left[(\Asso_{j}-1)\Asso_{j}\left\langle \bbeta_{j}-{\sf x},\v_2\right\rangle ^{2}+  \sigma^2 \cdot \Asso_{j}\left\Vert \v_2\right\Vert ^{2}\right]   \nonumber\\
            &\quad+2\E_{*}\left[\Asso_{i}\Asso_{j}\left\langle \bbeta_{i}-{\sf x},\v_1\right\rangle \left\langle \bbeta_{j}-{\sf x},\v_2\right\rangle \right].        \label{eqn:hessian_intermediate}
    \end{align}
    Noticing that $\sum_{i'\in[\kfit]}\Asso_{i'}=1$ and $\Asso_{i'}\ge 0$ for all $i' \in [\kfit]$, we observe that $\Asso_{i} + \Asso_{j} \leq 1$. 
    Then it follows from \eqref{eqn:hessian_intermediate} that
    \begin{align*}
        \sigma^4 \cdot \v^{\top} \nabla^2 L(\B) \v
            &\leq - \E_{*} \left[ \Asso_i \Asso_j \cdot \big\{ \left\langle \bbeta_{i}-{\sf x},\v_1\right\rangle ^{2}  + \left\langle \bbeta_{j}-{\sf x},\v_2\right\rangle ^{2} \big\} \right]
                 + 2 \E_{*} \left[ \Asso_i \Asso_j \cdot \left\langle \bbeta_{i}-{\sf x},\v_1\right\rangle \left\langle \bbeta_{j}-{\sf x},\v_2\right\rangle   \Big\} \right] \\
                &\quad
                + \sigma^2 \cdot \E_{*} \left[ \Asso_{i}\left\Vert \v_1\right\Vert^{2} + \Asso_{j}\left\Vert \v_2\right\Vert ^{2} \right]\\
            &= - \E_{*} \left[ \Asso_i \Asso_j \cdot \big( \left\langle \bbeta_{i}-{\sf x},\v_1\right\rangle - \left\langle \bbeta_{j}-{\sf x},\v_2\right\rangle \big)^2 \right]
                 +  \sigma^2 \cdot \left\Vert \v_1\right\Vert^{2} \cdot \E_{*} \left[ \Asso_{i} \right]
                +  \sigma^2 \cdot \left\Vert \v_2\right\Vert ^{2} \cdot \E_{*} \left[ \Asso_{j} \right].
    \end{align*}
    To complete the proof, it suffices to recall $\v^{\top} \nabla^2 L(\B) \v \geq 0$.
\end{proof}

\subsection{Completing the Proof of Proposition \ref{prop:proposition1}}

\begin{proof}[Proof of Proposition \ref{prop:proposition1}]
    In this proof, we prove the two claims separately.
    
    \bigskip
    \textit{\underline{(1) Proof of Claim 1.}}
    Fix arbitrary $i, j \in [\kfit]$ such that $i \neq j$ and let $\v_{ij}:= \frac{\bbeta_i - \bbeta_j}{\| \bbeta_i - \bbeta_j\|}$.
    Applying Lemma \ref{lem:2nd_implication} with $\v_1 = \v_2 = \v_{ij}$, we obtain
    \begin{align*}
         \sigma^2 \cdot \E_{*}\left[\Asso_{i}+\Asso_{j}\right]
            &\ge\E_{*}\left[\Asso_{i}\Asso_{j}\left\langle \bbeta_{i}-\bbeta_{j},\v_{ij}\right\rangle ^{2}\right]\\
            &=\left\Vert \bbeta_{i}-\bbeta_{j}\right\Vert ^{2}\cdot\E_{*}\left[\Asso_{i}\Asso_{j}\right].
    \end{align*}
    Rearranging the terms, we obtain the desired inequality.

    \bigskip
    \textit{\underline{(2) Proof of Claim 2.}}
    Choose an arbibrary pair $(s, i) \in [\ktrue] \times [\kfit]$ and let $\u_{s\to i}:=\frac{\bbeta_{i}-\bthetastar_{s}}{\left\Vert \bbeta_{i}-\bthetastar_{s}\right\Vert }$. 
    First, we observe that for any $\x\in\real^{d}$,
    \begin{align}
        \left\Vert \bbeta_{i}-\bthetastar_{s}\right\Vert ^{2}
            &=\left\langle \bbeta_{i}-\bthetastar_{s},\u_{s\to i}\right\rangle ^{2} \nonumber\\
            &=\Big( \left\langle \bbeta_{i}-\x, \u_{s\to i} \right\rangle + \left\langle \x - \bthetastar_{s} ,\u_{s\to i}\right\rangle \Big)^{2}
                \nonumber\\
            &\le2\left\langle \bbeta_{i}-\x,\u_{s\to i}\right\rangle ^{2}+2\left\langle \x-\bthetastar_{s},\u_{s\to i}\right\rangle ^{2}.   \label{eq:prop1_claim2.a}
    \end{align}
    Second, applying Lemma \ref{lem:2nd_implication} with $\v_1 = \u_{s \to i}$ and $\v_2 = 0$, we get for all $j\in[\kfit]\setminus\{i\}$,
    \begin{equation}\label{eq:prop1_claim2.b}
         \sigma^2 \cdot \E_{*}\left[\Asso_{i}\right]
            \ge\E_{*}\left[\Asso_{i}\Asso_{j}\left\langle \bbeta_{i}-{\sf x},\u_{s\to i}\right\rangle ^{2}\right].
    \end{equation}
    Then it follows that for all $j\in[\kfit]\setminus\{i\}$,
    \begin{align}
         \sigma^2 \cdot \E_{*}\left[\Asso_{i}\right]
            &\stackrel{(a)}{\geq} \E_{*}\left[\Asso_{i}\Asso_{j}\left\langle \bbeta_{i}-{\sf x},\u_{s\to i}\right\rangle ^{2}\right] 
                    \nonumber\\
            &\stackrel{(b)}{=} \frac{1}{\ktrue} \sum_{s \in [\ktrue]} \E_{s}\left[\Asso_{i}\Asso_{j}\left\langle \bbeta_{i}-{\sf x},\u_{s\to i}\right\rangle ^{2}\right] 
                    \nonumber\\
            &\stackrel{(c)}{\geq} \frac{1}{\ktrue} \sum_{s \in [\ktrue]} 
                \bigg\{ \frac{1}{2} \E_{s}\left[\Asso_{i}\Asso_{j} \cdot \| \bbeta_i - \bthetastar_s\|^2 \right] 
                     - \E_{s}\left[\Asso_{i}\Asso_{j} \cdot \left\langle {\sf x}-\bthetastar_{s},\u_{s\to i}\right\rangle ^{2} \right] \bigg\} 
                    \label{eq:prop1_claim2.c}
    \end{align}
    where (a) follows from \eqref{eq:prop1_claim2.b}; (b) is by definition; see \eqref{eq:true_mixture}; and (c) is due to \eqref{eq:prop1_claim2.a}.
    
    By our model assumption (Gaussian mixture model), the random variable ${\sf z}_{i} := \left\langle {\sf x}-\bthetastar_{s},\u_{s\to i}\right\rangle $ is a centered Gaussian random variable with variance $\sigma^2$. Therefore, for all $(s, i, j) \in [\ktrue] \times [\kfit] \times [\kfit]$,
    \begin{equation}\label{eq:prop1_claim2.d}
        \E_s\left[ \Asso_i \Asso_j \cdot {\sf z}_i^2 \right] \leq \E_s\left[ {\sf z}_i^2 \right] = \sigma^2.
    \end{equation}
    Plugging \eqref{eq:prop1_claim2.d} into \eqref{eq:prop1_claim2.c}, we get for all $j\in[\kfit]\setminus\{i\}$,
    \begin{align}
         \sigma^2 \cdot \E_{*}\left[\Asso_{i}\right] 
            &\geq \frac{1}{\ktrue} \sum_{s \in [\ktrue]} \left( \frac{1}{2} \E_{s}\left[\Asso_{i}\Asso_{j} \right] \cdot \| \bbeta_i - \bthetastar_s\|^2 - \sigma^2 \right) \label{eqn:lower_intermediate}\\
            &\geq \frac{1}{2} \E_{*}\left[ \Asso_i \Asso_j \right] \cdot \min_{s \in [\ktrue]} \| \bbeta_i - \bthetastar_s \|^2 - \sigma^2. \nonumber
    \end{align}
    Rearranging the terms, we complete the proof of Claim 2.

    \bigskip
    \textit{\underline{(3) Proof of Claim 3.}}
    Continuing on the inequality \eqref{eqn:lower_intermediate}, we observe that 
    \begin{align*}
         \sigma^2 \cdot \E_{*}\left[\Asso_{i}\right]
            &\geq \frac{1}{\ktrue} \sum_{s \in [\ktrue]} \left( \frac{1}{2} \E_{s}\left[\Asso_{i}\Asso_{j} \right] \cdot \| \bbeta_i - \bthetastar_s\|^2 - \sigma^2 \right)\\
            &\geq \frac{1}{2 \ktrue} \E_{s}\left[\Asso_{i}\Asso_{j} \right] \cdot \| \bbeta_i - \bthetastar_s\|^2 - \sigma^2
    \end{align*}
    for all $s \in [\ktrue]$ and all $j \in [\kfit] \setminus \{i\}$ because $\E_{s}\left[\Asso_{i}\Asso_{j} \right] \cdot \| \bbeta_i - \bthetastar_s\|^2 \geq 0$ for all $(s,i,j) \in [\ktrue] \times [\kfit] \times [\kfit]$.

\end{proof}

\section{Deferred Proof of Proposition \ref{prop:small_bdr2}}\label{sec:proof_proposition.3}

\subsection{Helper Lemmas}

\paragraph{Basic properties}
Recall the definition of the Gaussian Q-function $Q(t) = \int_t^{\infty} \frac{1}{\sqrt{2\pi}} e^{-\frac{1}{2} s^2} ds$, and let $Q^{-1}: (0,1) \to \real$ denote its inverse. 
That is, we write $z = Q^{-1}(t)$ if $t = Q(z)$. 
\begin{lemma}\label{lem:deviation_bd}
    Let $\sfx \sim \cN(0, \Id_d)$ be the $d$-variate standard Gaussian random variable, and $\psi: \real^d \to [0,1]$. 
    Then
    \[
        \left\| \E_{\sfx}\big[ \psi(\sfx) \cdot \sfx \big] \right\| \leq \phi(z_{\psi})
    \]
    where $z_{\psi} = Q^{-1} \big( \E[\psi(\sfx)] \big)$.
\end{lemma}

Lemma \ref{lem:deviation_bd} implies an upper bound on the mean displacement of a Gaussian distribution $\cN(0, \Id_d)$ when reweighted using $\frac{\psi(\sfx)}{\E_x[\psi(\sfx)]}$:
\[
    \left\| \, \E_{\sfx}\left[ \frac{\psi(\sfx)}{\E_{\sfx}\big[ \psi(\sfx) \big]} \cdot \sfx \right] \, \right\| \leq \frac{\phi(z_{\psi})}{\E_{\sfx}\big[ \psi(\sfx) \big]}.
\]

\begin{proof}[Proof of Lemma \ref{lem:deviation_bd}]
    It suffices to observe that
    \begin{align*}
        \left\| \E_{\sfx}\big[ \psi(\sfx) \cdot \sfx \big] \right\|
            &= \sup_{u \in \real^d: \|u\|=1} \E_{\sfx} \left[ \psi(\sfx) \cdot \langle u, \sfx \rangle \right]\\
            &\leq \int_{z_\psi}^{\infty} t \cdot \phi(t) ~dt
            = \phi(z_\psi).
    \end{align*}
\end{proof}

\begin{lemma}\label{lem:convex_discr}
    Let $\w, \w'\in \{ \tilde{\w} \in \real^{\ktrue}: \tilde{w}_s \geq 0, \forall s \in [\ktrue], ~\sum_{s=1}^{\ktrue} \tilde{w}_s = 1 \}$, and 
    $\{ \v_1, \dots, \v_{\ktrue} \}, \{ \v'_1, \dots, \v'_{\ktrue} \} \subset \real^d$. 
    Then
    \begin{align*}
        \left\| \sum_{s=1}^{\ktrue} w_s \v_s - \sum_{s=1}^{\ktrue} w'_s \v'_s \right\| 
            &\leq \| \w - \w' \|_1 \cdot \min_{s \in [\ktrue]} \max_{s' \in [\ktrue]} \| \v_s - \v_{s'} \| 
                + \sum_{s=1}^{\ktrue} \left\|  w'_s (\v_s - \v'_s) \right\|.
    \end{align*}
\end{lemma}
\begin{proof}[Proof of Lemma \ref{lem:convex_discr}]
    Applying triangle inequality, we obtain
    \begin{align*}
        \left\| \sum_{s=1}^{\ktrue} w_s \v_s - \sum_{s=1}^{\ktrue} w'_s \v'_s \right\|
            &\leq \left\| \sum_{s=1}^{\ktrue} ( w_s - w'_s)  \v_s \right\|
                + \left\| \sum_{s=1}^{\ktrue} w'_s (\v_s - \v'_s) \right\|\\
            &= \left\| \sum_{s=2}^{\ktrue} ( w_s - w'_s) (\v_s - \v_1) \right\|
                + \left\| \sum_{s=1}^{\ktrue} w'_s (\v_s - \v'_s) \right\|\\
            &\leq \left( \sum_{s=2}^{\ktrue} |w_s - w'_s| \right) \cdot \max_{s \in [\ktrue]} \| \v_s - \v_1 \| 
                + \sum_{s=1}^{\ktrue} \left\|  w'_s (\v_s - \v'_s) \right\|\\
            &\leq \| \w - \w' \|_1 \cdot \max_{s \in [\ktrue]} \| \v_s - \v_1 \| 
                + \sum_{s=1}^{\ktrue} \left\|  w'_s (\v_s - \v'_s) \right\|.
    \end{align*}
    Since the choice of $\bv_1$ is arbitrary, we can replace it with any $\bv_{s'}$, $s' \in [\ktrue]$.
\end{proof}

\paragraph{Properties of the coefficients of association}
Next, we present two useful properties of the coefficients of association, which we use later in the proof of Proposition \ref{prop:small_bdr2}.

Recall the definition of set $\setA_i^{\delta}$ from \eqref{eq:setA}: for any $i \in [\kfit]$ and for any $\delta \geq 0$,
\begin{align*}
    \setA_i^{\delta} \coloneqq \bigg\{ s \in [\ktrue] ~\bigg|~ \iota_{\B}(\bthetastar_s) = i ~~~\text{and}
        ~~~\max_{j, j' \in [\kfit]}  \frac{ \| \bbeta_j - \bbeta_{j'} \| }{\sigma} \cdot \E_s \left[ \Asso_j \Asso_{j'} \right] < \delta \bigg\}.
\end{align*}
We additionally define $\setB_i^{\delta}$ as follows: for any $i \in [\kfit]$ and for any $\delta \geq 0$,
\begin{equation}\label{eq:setB}
\begin{aligned}
    \setB_i^{\delta} \coloneqq \bigg\{ s \in [\ktrue] ~\bigg|~ &\iota_{\B}(\bthetastar_s) \neq i ~~~\text{and}
        ~~~\max_{j \in [\kfit]\setminus\{i\}}  \frac{ \| \bbeta_i - \bbeta_j \| }{\sigma} \cdot  \E_s \left[ \Asso_i \Asso_{j} \right] < \delta \bigg\}.
\end{aligned}
\end{equation}

\begin{lemma}\label{lem:A_or_B}
    Let $i \in [\kfit]$ and $\delta > 0$. 
    If $\setA_{i}^{\delta} \neq \emptyset$ and $\delta > \frac{4 \ktrue \cdot \sigma }{\deltamin} $, 
    then $\left| \setA_i^{\delta} \cup \setB_i^{\delta} \right| \geq \ktrue - 1$. 
\end{lemma}

\begin{proof}[Proof of Lemma \ref{lem:A_or_B}]
    Observe that if $s \in [\ktrue] \setminus \big( \setA_i^{\delta} \cup \setB_i^{\delta} \big)$, then either of the following must be true by definition of the sets $\setA_i^{\delta}$ and $\setB_i^{\delta}$:
    \begin{enumerate}[label=(\alph*)]
        \item
        $\iota_{\B}(\bthetastar_s) = i$ and $\max_{j \in [\kfit]} \frac{ \| \bbeta_i - \bbeta_{j} \| }{\sigma} \cdot \E_s \left[ \Asso_i \Asso_{j} \right] \geq \delta$; or
        \item
        $\iota_{\B}(\bthetastar_s) \neq i$ and $\max_{j \in [\kfit]} \frac{ \| \bbeta_i - \bbeta_j \| }{\sigma} \cdot \E_s \left[ \Asso_i \Asso_{j} \right] \geq \delta$.
    \end{enumerate}
    In either cases, there must exists $j \in [\kfit] \setminus \{i\}$ such that $\frac{ \| \bbeta_{i} - \bbeta_{j} \| }{\sigma} \cdot \E_s\left[ \Asso_{i} \Asso_{j} \right] \geq \delta$. 
    That is, $i, j \in \cE_s^{\delta}$ by definition of the set $\cE_s^{\delta}$ presented in \eqref{eqn:entangled}.
    By Corollary \ref{cor:proximity}, 
    \begin{align}\label{eqn:prop3_proximity}
        \frac{\| \bbeta_{i} - \bthetastar_s\|}{\sigma} 
            &\leq \frac{ 2 \ktrue}{\delta}. 
    \end{align}
    
    Now we assume that $\left| \setA_i^{\delta} \cup \setB_i^{\delta} \right| < \ktrue - 1$, i.e., $\left| [\ktrue] \setminus \big( \setA_i^{\delta} \cup \setB_i^{\delta} \big) \right| \geq 2$. 
    We choose $s_1, s_2 \in [\ktrue] \setminus \big( \setA_i^{\delta} \cup \setB_i^{\delta} \big)$ such that $s_1 \neq s_2$. 
    Observe that $i \in \cE^{\delta}_{s_1} \cap \cE^{\delta}_{s_2}$, and therefore, it follows from \eqref{eqn:prop3_proximity} that
    \begin{align*}
        \| \bthetastar_{s_1} - \bthetastar_{s_2} \|
            &\leq \| \bbeta_{l_1} - \bthetastar_{s_1} \| + \| \bbeta_{l_1} - \bthetastar_{s_2} \|\\
            &\leq  \frac{4 \ktrue \cdot \sigma}{\delta}.
    \end{align*}
    If $\delta > \frac{4 \ktrue \cdot \sigma}{\deltamin}$, then $ \frac{4 \ktrue \cdot \sigma}{\delta} < \deltamin$, contradicting the definition $\deltamin\coloneqq\min_{\substack{s,s'\in[\ktrue]\\s\neq s} }\left\Vert \bthetastar_{s}-\bthetastar_{s'}\right\Vert$. 
    Therefore we conclude that $\left| \setA_i^{\delta} \cup \setB_i^{\delta} \right| \geq \ktrue - 1$.
\end{proof}

\begin{lemma}\label{lem:association_bound}
    Let $i \in [\kfit]$ and $\delta > 0$. 
    If $\setA_{i}^{\delta} \neq \emptyset$, then 
    \begin{align*}
        \max_{s \in \setB_{i}^{\delta}} \big\| \bthetastar_s - \bbeta_i \big\| \cdot \E_s \big[ \Asso_i \big] 
            &\leq 45 \left(\sqrt{2\pi} + 1 \right) \cdot \kfit^3 \cdot \delta \cdot \deltacell(\B) + 4 \std
    \end{align*}
    where $\deltacell(\B) \coloneqq \max_{j \in [\kfit]} \max_{s \in [\ktrue]: \bthetastar_s \in \vor_j} \| \bthetastar_s - \bbeta_j \|$. 
\end{lemma}

\begin{proof}[Proof of Lemma \ref{lem:association_bound}]
    Choose an arbitrary $s \in \setB_{i}^{\delta}$ and let $j = \iota_{\B}(\bthetastar_s)$. 
    Notice that $j \neq i$, and we recall from \eqref{eq:bbeta_bar} that $\bbetabar_{ij} \coloneqq \frac{\bbeta_{i}+\bbeta_{j}}{2}$. 
    Then we define three quantities:
    \begin{align*}
        d_{ij} &\coloneqq \big\| \bbeta_i - \bbeta_j \|,\\
        t &\coloneqq \left\langle \bthetastar_s - \bbetabar_{ij}, \frac{\bbeta_j - \bbeta_i}{ \| \bbeta_j - \bbeta_i \| } \right\rangle,\\
        n &\coloneqq 
            \sqrt{ \| \bthetastar_s - \bbeta_j \|^2 - \left\langle \bthetastar_s - \bbeta_j, \frac{\bbeta_j - \bbeta_i}{\|\bbeta_j - \bbeta_i\|} \right\rangle }.
    \end{align*}
    In what follows, we prove the lemma by considering two cases: (1) $d_{ij} \leq \deltacell(\B)$ or $\big| t - \frac{d_{ij}}{2} \big| \geq \frac{d_{ij}}{4}$ or $n \geq \frac{d_{ij}}{4}$; and (2) $d_{ij} > \deltacell(\B)$ and $\big| t - \frac{d_{ij}}{2} \big| < \frac{d_{ij}}{4}$ and $n < \frac{d_{ij}}{4}$.
    
    \paragraph{Case 1.} 
    We observe that
    \begin{align}
        \frac{1}{9 \left(\sqrt{2\pi} + 1 \right)} \E_{s}\left[\Asso_{i}\right]
            &\stackrel{(a)}{\leq}
            \kfit^2 \sum_{j\in [\kfit]\setminus\{i\}} \frac{ \| \bbeta_i - \bbeta_j \| }{\sigma} \cdot \E_{s}\left[ \Asso_i \Asso_j  \right]
                \nonumber\\
            &\stackrel{(b)}{\leq}
            \kfit^2 \sum_{j\in [\kfit]\setminus\{i\}} \delta
                \nonumber\\
            &\leq \kfit^3 \cdot \delta.   \label{eqn:asso_i_upper}
    \end{align}
    where (a) is due to Corollary \ref{cor:small_bdr}, Claim 2, and (b) is by definition of $\setB_i^{\delta}$, cf. \eqref{eq:setB}. 
    Thereafter, we consider the three subcases individually.

    \bigskip
    \textit{Case 1-A.} 
    First of all, suppose that $d_{ij} \leq \deltacell(\B)$. 
    Observe that 
    \begin{align*}
        \| \bthetastar_s - \bbeta_i \| 
            &\leq \| \bthetastar_s - \bbeta_j \| + \| \bbeta_j - \bbeta_i \|\\
            &\leq \deltacell(\B) + d_{ij}\\
            &\leq 2 \deltacell(\B).
    \end{align*}
    This, combined with \eqref{eqn:asso_i_upper}, yields
    \begin{equation}\label{eqn:case1_A}
        \big\| \bthetastar_s - \bbeta_i \big\| \cdot \E_s \big[ \Asso_i \big] 
            \leq 18 \left(\sqrt{2\pi} + 1 \right) \cdot \kfit^3 \cdot \delta \cdot \deltacell(\B).
    \end{equation}

    \bigskip
    \textit{Case 1-B.} 
    Next, suppose that $\big| t - \frac{d_{ij}}{2} \big| \geq \frac{d_{ij}}{4}$. 
    It is easy to observe that 
    \[
        \| \bthetastar_s - \bbeta_i \| 
            \leq 5 \| \bthetastar_s - \bbeta_j \| 
            \leq 5 \deltacell(\B).
    \]
    Likewise, it follows from \eqref{eqn:asso_i_upper} that
    \begin{equation}\label{eqn:case1_B}
        \big\| \bthetastar_s - \bbeta_i \big\| \cdot \E_s \big[ \Asso_i \big] 
            \leq 45 \left(\sqrt{2\pi} + 1 \right) \cdot \kfit^3 \cdot \delta \cdot \deltacell(\B).
    \end{equation}
    
    \bigskip
    \textit{Case 1-C.} 
    Lastly, suppose that $n \geq \frac{d_{ij}}{4}$. 
    Similarly, we observe that 
    \begin{align*}
        \| \bthetastar_s - \bbeta_i \| 
            &\leq \left( \frac{(\sqrt{5}+2)^2 + 1}{(\sqrt{5}-2)^2 + 1} \right)^{1/2} \| \bthetastar_s - \bbeta_j \| \\
            &\leq 5 \deltacell(\B),
    \end{align*}
    and therefore, 
    \begin{equation}\label{eqn:case1_C}
        \big\| \bthetastar_s - \bbeta_i \big\| \cdot \E_s \big[ \Asso_i \big] 
            \leq 45 \left(\sqrt{2\pi} + 1 \right) \cdot \kfit^3 \cdot \delta \cdot \deltacell(\B).
    \end{equation}

    \paragraph{Case 2.} 
    Now we suppose that (i) $d_{ij} > \deltacell(\B)$, (ii) $\big| t - \frac{d_{ij}}{2} \big| < \frac{d_{ij}}{4}$, and (iii) $n < \frac{d_{ij}}{4}$. 
    Observe that
    \begin{equation}\label{eqn:case2_term.A}
        \| \bthetastar_s - \bbeta_i \|^2 
            = \left( t + \frac{d_{ij}}{2} \right)^2 + n^2
            < \frac{13}{8} d_{ij}^2.
    \end{equation}

    Next, we fix a coordinate system: without loss of generality, we may let $\bthetastar_s = 0$ and let $\e_1 = \frac{\bbeta_i - \bbeta_j}{\|\bbeta_i - \bbeta_j\|}$; see Figure \ref{fig:illustration_integral}. 
    Writing $\x = (x_1, x_2^d)$, we observe that for any $\alpha \in \big(0, \frac{1}{2}\big)$,
    \begin{align*}
        \E_s\big[ \Asso_i \big] 
            &= \E_s\left[ \Asso_i \cdot \indic\Big\{ \sfx_1 \leq \alpha \cdot d_{ij} \Big\} \right] 
                + \E_s\left[ \Asso_i \cdot \indic\Big\{ \sfx_1 > \alpha \cdot d_{ij} \Big\} \right].
    \end{align*}
    It suffices to establish upper bounds for the two terms on the right hand side.
    \begin{itemize}
        \item 
        If $x_1 \leq \alpha \cdot d_{ij}$, then
        \begin{align*}
            \asso_i(\x) 
                &= \frac{e^{-\frac{\|\x-\bbeta_i\|^2}{2 \std^2}}}{\sum_{j'=1}^{\kfit}e^{-\frac{\|\x-\bbeta_{j'}\|^2}{2  \std^2}}}\\
                &\leq \frac{e^{-\frac{\|\x-\bbeta_i\|^2}{2 \std^2}}}{e^{-\frac{\|\x-\bbeta_j\|^2}{2 \std^2}}}\\
                &= e^{ \frac{1}{\std^2} \left\langle \bbeta_i - \bbeta_j, \x - \bbetabar_{ij} \right\rangle }\\
                &\leq e^{- \big( \frac{1}{2} - \alpha \big) \frac{d_{ij}^2}{\std^2}}.
        \end{align*}
        Thus, $\E_s\left[ \Asso_i \cdot \indic\Big\{ \sfx_1 \leq \alpha \cdot d_{ij} \Big\} \right] \leq e^{- \big( \frac{1}{2} - \alpha \big) \frac{d_{ij}^2}{\std^2}}$.
        \item 
        Since $\asso_i(\x) \leq 1$, we have
        \begin{align*}
            \E_s\left[ \Asso_i \cdot \indic\Big\{ \sfx_1 > \alpha \cdot d_{ij} \Big\} \right]
                &\leq \E_s\left[ \indic\Big\{ \sfx_1 > \alpha \cdot d_{ij} \Big\} \right]\\
                &= \P_s\left( \sfx_1 > \alpha \cdot d_{ij} \right)\\
                &= Q \left( \frac{\alpha \cdot d_{ij}}{\std}  \right)\\
                &\stackrel{(a)}{\leq} \sqrt{2\pi} \phi\left( \frac{\alpha \cdot d_{ij}}{\std} \right)
                    \\
                &= e^{- \frac{\alpha^2}{2} \frac{d_{ij}^2}{\std^2}},
        \end{align*}
        where (a) follows from Lemma \ref{lem:gaussian_tail}.
    \end{itemize}
    We optimize the value of $\alpha \in (0, 1/2)$ to balance these two upper bounds by solving $\frac{1}{2}-\alpha = \frac{\alpha^2}{2}$, and obtain $\alpha^{\star} = \sqrt{2} - 1$. Therefore,
    \begin{equation}\label{eqn:case2_term.B}
    \begin{aligned}
        \E_s\big[ \Asso_i \big] 
            &= \E_s\left[ \Asso_i \cdot \indic\Big\{ \sfx_1 \leq \alpha^{\star} \cdot d_{ij} \Big\} \right] 
                + \E_s\left[ \Asso_i \cdot \indic\Big\{ \sfx_1 > \alpha^{\star} \cdot d_{ij} \Big\} \right]\\
            &\leq 2 e^{- \big(\frac{3}{2} - \sqrt{2} \big) \frac{d_{ij}^2}{\std^2}}.
    \end{aligned}
    \end{equation}

    Combining \eqref{eqn:case2_term.A} and \eqref{eqn:case2_term.B}, we obtain
    \begin{align*}
        \big\| \bthetastar_s - \bbeta_i \big\| \cdot \E_s \big[ \Asso_i \big]
            \leq \std \cdot \sqrt{\frac{13}{2}} \frac{d_{ij}}{\std} \cdot e^{- \big(\frac{3}{2} - \sqrt{2} \big) \frac{d_{ij}^2}{\std^2}}
            \leq 4 \std,
    \end{align*}
    where the last inequality follows from observing $\max_{z \geq 0} \Big\{ \sqrt{13/2} \cdot z \cdot e^{- \big(\frac{3}{2} - \sqrt{2} \big) z^2 } \Big\} < 4$.
    
    \begin{figure}[t]
    \begin{centering}
    \includegraphics[width=0.6 \linewidth]{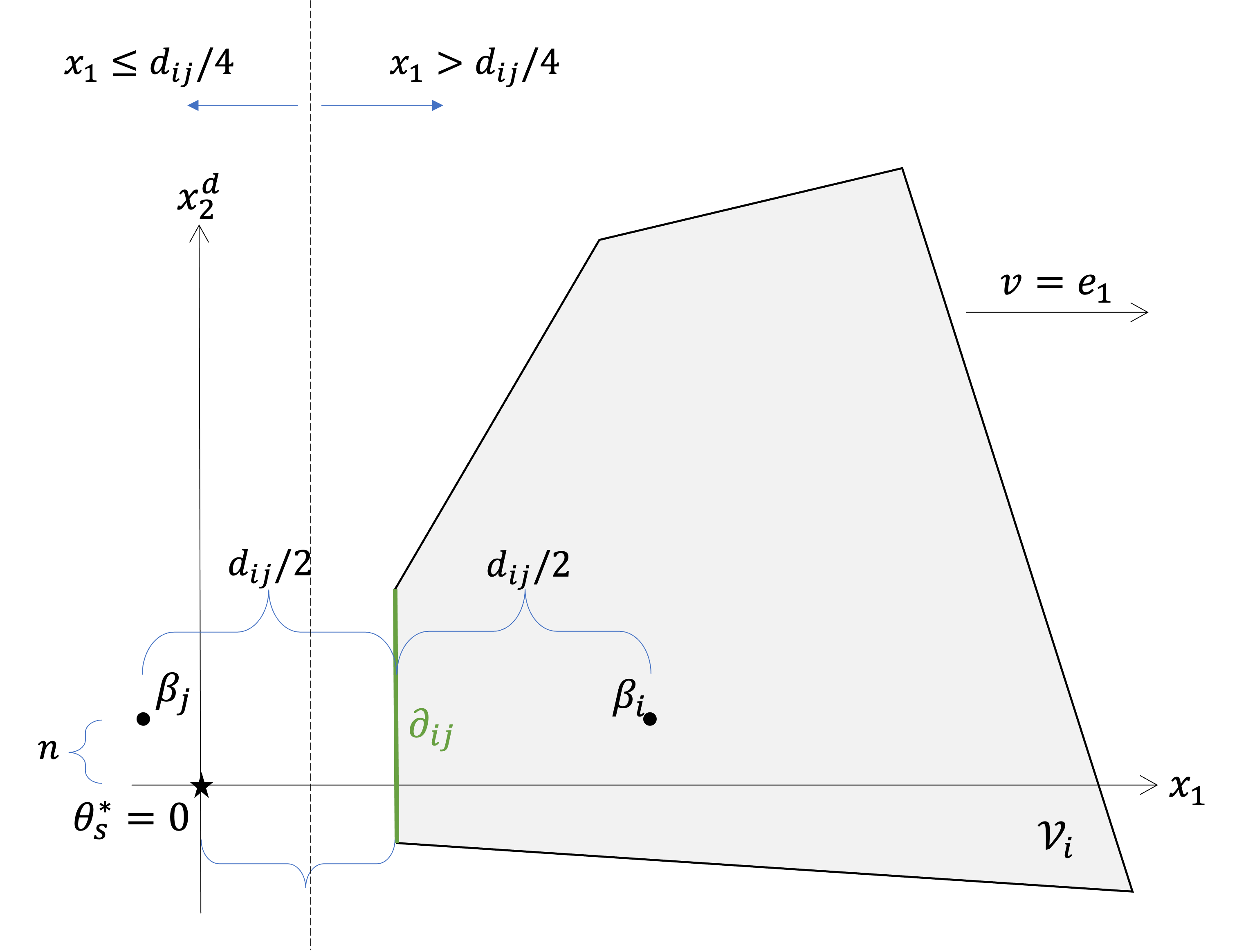}
    \par\end{centering}
        \caption{Illustration for the proof of Lemma~\ref{lem:association_bound}, specifically Case 2.}
        \label{fig:illustration_integral}
    \end{figure}
\end{proof}

    


\subsection{Proof of Proposition \ref{prop:small_bdr2}}
\begin{proof}[Proof of Proposition \ref{prop:small_bdr2}]
We present a proof of Proposition \ref{prop:small_bdr2} in three steps.

\paragraph{Step 1. Preliminary observations}
Fix $\delta > 0$ and choose an $i \in [\kfit]$ such that $\setA_i^{\delta} \neq \emptyset$. 
Observe that
\begin{align*}
    \bbeta_i
        &= \frac{\E_*[ \Asso_i \sfx]}{\E_*[ \Asso_i ]}\\
        &= \sum_{s \in [\ktrue]} \underbrace{\frac{ \E_s[\Asso_i]}{ \sum_{s \in [\ktrue]} \E_s[\Asso_i]}}_{\eqqcolon w_s} \cdot \underbrace{\E_s \left[ \frac{\Asso_i}{\E_s[\Asso_i]} \cdot \sfx \right]}_{\eqqcolon\bv_s}.
\end{align*}
by Lemma \ref{lem:gradient_NLL}, and more specifically, by \eqref{eq:station_cond}.
We let for each $s \in [\ktrue]$,
\[
    \hw_s \coloneqq \frac{ \indic\{s \in \setA_i^{\delta}\} }{ \sum_{s \in [\ktrue]} \indic\{s \in \setA_i^{\delta}\} }
        = \begin{cases} \frac{1}{|\setA_i^{\delta}|} & \text{if }s \in \setA_i^{\delta},\\ 0 & \text{otherwise.} \end{cases}
\]
Then we define
\begin{align*}
    \hbbeta_i
        &\coloneqq \sum_{s \in [\ktrue]} \hw_s \cdot \bthetastar_s
        =\frac{1}{|\setA_i^{\delta}|} \sum_{s \in \setA_i^{\delta}} \bthetastar_s.
\end{align*}
Further, we define two additional points in $\real^d$ that serve as proxies of $\bbeta_i$ to facilitate our analysis:
\begin{align}
    \bbeta'_i &\coloneqq \sum_{s \in \setB_i^{\delta} } w_s \cdot \bbeta_i + \sum_{s \in [\ktrue] \setminus \setB_i^{\delta} } w_s \cdot \bv_s,
        \label{eqn:beta_p}\\
    \bbeta''_i &\coloneqq \sum_{s \in \setA_i^{\delta} } \frac{w_s}{\sum_{s \in \setA_i^{\delta}} w_s} \cdot \bv_s.
        \label{eqn:beta_pp}
\end{align}

Observe that
\begin{align*}
    \bbeta'_i - \bbeta_i 
        &\stackrel{(a)}{=} \left( \sum_{s \in \setB_i^{\delta} } w_s \cdot \bbeta_i + \sum_{s \in [\ktrue] \setminus \setB_i^{\delta} } w_s \cdot \bv_s \right) - \bbeta_i
            \\
        &= \sum_{s \in \setA_i^{\delta}} w_s \cdot \big( \bv_s - \bbeta_i \big) + \sum_{s \in \setC_i^{\delta}} w_s \cdot \big( \bv_s - \bbeta_i \big)
            \\
        &\stackrel{(b)}{=} \left( \sum_{s \in \setA_i^{\delta}} w_s \right) \cdot \big( \bbeta''_i - \bbeta_i \big)
            + \sum_{s \in \setC_i^{\delta}} w_s \cdot \big( \bv_s - \bbeta_i \big),
\end{align*}
where (a) is due to \eqref{eqn:beta_p} and (b) is due to \eqref{eqn:beta_pp}.
Therefore,
\begin{equation}\label{eqn:beta_prime}
\begin{aligned}
    \big\| \bbeta''_i - \bbeta_i \big\|
        &\leq \frac{1}{\sum_{s \in \setA_i^{\delta}} w_s} \left( \big\|\bbeta'_i - \bbeta_i\big\| + \sum_{s \in \setC_i^{\delta}} w_s \cdot \big\| \bv_s - \bbeta_i \big\| \right).
\end{aligned}
\end{equation}
By triangle inequality and \eqref{eqn:beta_prime}, we obtain
\begin{align}
    \big\| \hbbeta_i - \bbeta_i \big\|
        &\leq \big\| \hbbeta_i - \bbeta''_i \big\| + \big\| \bbeta''_i - \bbeta_i \big\|
            \nonumber\\
        &\leq \big\| \hbbeta_i - \bbeta''_i \big\| 
            + \frac{1}{\sum_{s \in \setA_i^{\delta}} w_s} \big\|\bbeta'_i - \bbeta_i\big\|
            + \frac{1}{\sum_{s \in \setA_i^{\delta}} w_s} \sum_{s \in \setC_i^{\delta}} w_s \cdot \big\| \bv_s - \bbeta_i \big\|. 
            \label{eqn:prop4_step1}
\end{align}

\paragraph{Step 2. Establishing separate upper bounds}
Next, we establish upper bounds for the three terms on the right hand side of \eqref{eqn:prop4_step1}.

\bigskip\noindent
\textit{Step 2-A. An upper bound on the first term in \eqref{eqn:prop4_step1}.} 
Recall that $\hbbeta_i = \sum_{s \in \setA_i^{\delta}} \hw_s \cdot \hbv_s$ and $\bbeta''_i = \sum_{s \in \setA_i^{\delta}} w''_s \cdot \bv_s$ where 
\[
    w''_s = \frac{w_s}{\sum_{s \in \setA_i^{\delta}} w_s} = \frac{\E_s[\Asso_i]}{\sum_{s \in \setA_i^{\delta}} \E_s[\Asso_i]}.
\]
Therefore, by Lemma \ref{lem:convex_discr}, we obtain
\begin{align}\label{eqn:prop4_step2A.1}
    \| \hbbeta_i - \bbeta''_i \|
        &\leq \left( \sum_{s \in \setA_i^{\delta}} | \hw_s - w''_s | \right) \cdot \min_{s \in \setA_i^{\delta}} \max_{s' \in \setA_i^{\delta}} \| \bthetastar_s - \bthetastar_{s'}  \| 
            + \sum_{s \in \setA_i^{\delta}} \left\| w''_s \cdot \big( \bthetastar_s - \bv_s \big) \right\|.
\end{align}

Let $\bdw \coloneqq 9 \left(\sqrt{2\pi} + 1 \right) \cdot \kfit^3 \cdot \delta$, to avoid cluttered notation. 
Recall from Corollary \ref{cor:small_bdr} and Proposition \ref{prop:exclusive} that 
\begin{align}
    1 - \E_s[ \Asso_i ] &\leq 9 \left(\sqrt{2\pi} + 1 \right) \cdot \kfit^4 \cdot \delta
        = \kfit \cdot\bdw,
        &&\forall s \in \setA_i^{\delta},
            \label{eqn:bdw_A}\\
    \E_s[ \Asso_i ] &\leq 9 \left(\sqrt{2\pi} + 1 \right) \cdot \kfit^3 \cdot \delta 
        = \bdw,
        &&\forall s \in \setB_i^{\delta}.
            \label{eqn:bdw_B}
\end{align}

It follows from \eqref{eqn:bdw_A} that for all $s \in \setA_i^{\delta}$,
\begin{align}
    \big| \hw_s - w''_s \big| 
        &\leq \max \left\{ \frac{1}{|\setA_i^{\delta}|} - \frac{1 - \kfit \cdot \bdw}{|\setA_i^{\delta}|}, \frac{1}{|\setA_i^{\delta}|\big( 1 - \kfit \cdot \bdw \big) } - \frac{1}{|\setA_i^{\delta}|} \right\}    \nonumber\\
        &= \frac{1}{|\setA_i^{\delta}|} \cdot \frac{\kfit \cdot \bdw}{1 - \kfit \cdot \bdw}  \label{eqn:prop4_step2A.2}
\end{align}

Let $z_{s,i} \coloneqq Q^{-1} \big( \E_s[\Asso_i] \big)$ for all $s \in \setA_i^{\delta}$, where $Q$ is the Gaussian Q-funciton. 
Observe that
\begin{align}
    \sum_{s \in \setA_i^{\delta}} \left\| w''_s \cdot \big( \bthetastar_s - \bv_s \big) \right\| 
        &= \frac{1}{\sum_{s \in \setA_i^{\delta}} \E_s[\Asso_i]} \sum_{s \in \setA_i^{\delta}} \left\| \E_s[\Asso_i] \cdot \big( \bthetastar_s - \bv_s \big) \right\|
            \nonumber\\
        &\stackrel{(a)}{=} \frac{1}{\sum_{s \in \setA_i^{\delta}} \E_s[\Asso_i]} \sum_{s \in \setA_i^{\delta}} \left\| \E_s \big[\Asso_i \cdot \big( \sfx - \bthetastar_s \big) \big] \right\|
            \nonumber\\
        &\stackrel{(b)}{\leq} \frac{1}{\sum_{s \in \setA_i^{\delta}} \E_s[\Asso_i]} \sum_{s \in \setA_i^{\delta}} \phi( z_{s, i} ) \cdot \std
            \nonumber\\
        &\stackrel{(c)}{\leq} \frac{3 \kfit \cdot \bdw}{1 - \kfit \cdot \bdw} \cdot \std 
            \label{eqn:prop4_step2A.3}
\end{align}
where (a) holds because $\bv_s = \E_s \left[ \frac{\Asso_i}{\E_s[\Asso_i]} \cdot \sfx \right]$; (b) is by Lemma \ref{lem:deviation_bd}; and (c) follows from the observation: for all $s \in \setA_i^{\delta}$,
\begin{align*}
    &1 - \E_s[\Asso_i] \leq \kfit \cdot \bdw \leq \frac{1}{4}\\
    &\implies\quad
    z_{s,i} \leq -0.5\\
    &\implies\quad
    \phi( z_{\delta} ) = \phi( -z_{\delta} ) \leq \frac{z_{\delta}^2 + 1}{-z_{\delta}} Q(-z_{\bdw}) \leq 3 \kfit \cdot \bdw.
\end{align*}

Combining the inequalities \eqref{eqn:prop4_step2A.1}, \eqref{eqn:prop4_step2A.2}, \eqref{eqn:prop4_step2A.3}, we obtain
\begin{align}\label{eqn:prop4_step2A}
    \| \hbbeta_i - \bbeta''_i \| 
        &\leq \frac{\kfit \cdot \bdw}{1 - \kfit \cdot \bdw} \cdot \left( \min_{s \in \setA_i^{\delta}} \max_{s' \in \setA_i^{\delta}} \| \bthetastar_s - \bthetastar_{s'} \|
            + 3 \cdot \std \right).
\end{align}

\bigskip\noindent
\textit{Step 2-B. An upper bound on the second term in \eqref{eqn:prop4_step1}.} 
Recall the definition of $\setB_i^{\delta}$ from \eqref{eq:setB} and that $\bbeta'_i = \sum_{s \in \setB_i^{\delta} } w_s \cdot \bbeta_i + \sum_{s \in [\ktrue] \setminus \setB_i^{\delta} } w_s \cdot \bv_s$, cf. \eqref{eqn:beta_p}. 
Thus, we observe that
\begin{align}\label{eqn:residual_term}
    \frac{1}{\sum_{s \in \setA_i^{\delta}} w_s} \cdot \big\|\bbeta'_i - \bbeta_i\big\| 
        &= \frac{1}{\sum_{s \in \setA_i^{\delta}} w_s} \cdot \left\| \sum_{s \in \setB_i^{\delta}} w_s \cdot \big( \bv_s - \bbeta_i \big) \right\|
            \nonumber\\
        &\leq \frac{1}{\sum_{s \in \setA_i^{\delta}} \E_s[\Asso_i]} \cdot \sum_{s \in \setB_i^{\delta}} \left\|  \E_s[\Asso_i] \cdot \big( \bv_s - \bbeta_i \big) \right\|.
\end{align}
Recall that $\bv_s = \E_s \left[ \frac{\Asso_i}{\E_s[\Asso_i]} \cdot \sfx \right]$, and thus,
\begin{align*}
    \E_s[\Asso_i] \cdot \big( \bv_s - \bbeta_i \big)
        &= \E_s[\Asso_i \cdot \sfx ] - \E_s[\Asso_i] \cdot \bbeta_i \\
        &= \E_s \big[\Asso_i \cdot (\sfx - \bthetastar_s) \big] + \E_s[\Asso_i] \cdot \big( \bthetastar_s - \bbeta_i\big).
\end{align*}
Observe that 
\begin{equation}\label{eqn:simple_upper_norm}
\begin{aligned}
    \big\| \E_s \big[\Asso_i \cdot (\sfx - \bthetastar_s) \big] \big\|
        &\leq \E_s \big[\Asso_i \cdot \| \sfx - \bthetastar_s\| \big] \\
        &\leq E_s \| \sfx - \bthetastar_s\| \\
        &\leq \std \sqrt{d}.
\end{aligned}
\end{equation}
By Lemma \ref{lem:association_bound} and \eqref{eqn:simple_upper_norm}, we have
\begin{equation}\label{eqn:simple_upper_norm.2}
\begin{aligned}
    \big\| \E_s[\Asso_i] \cdot \big( \bv_s - \bbeta_i \big) \big\|
        &\leq 45 \left(\sqrt{2\pi} + 1 \right) \cdot \kfit^3 \cdot \delta \cdot \deltacell(\B) 
            + \std \big( \sqrt{d} + 4 \big)\\
        &= 5 \bdw \cdot \deltacell(\B) + \std \big( \sqrt{d} + 4 \big). 
\end{aligned}
\end{equation}
Combining \eqref{eqn:residual_term} and \eqref{eqn:simple_upper_norm.2}, we obtain
\begin{equation}\label{eqn:prop4_step2B}
\begin{aligned}
    \frac{1}{\sum_{s \in \setA_i^{\delta}} w_s} \cdot \big\|\bbeta'_i - \bbeta_i\big\|
        \leq \frac{|\setB_i^{\delta}|}{|\setA_i^{\delta}|} \cdot \frac{ 5 \bdw \cdot \deltacell(\B) + \std \big( \sqrt{d} + 4 \big)}{ 1 - \kfit \cdot \bdw }.
\end{aligned}
\end{equation}

\bigskip\noindent
\textit{Step 2-C. An upper bound on the third term in \eqref{eqn:prop4_step1}.} 
Recall that $\setC_i^{\delta}$ is either an empty set or a singleton. 
If $\setC_i^{\delta} = \emptyset$, there is nothing to prove. 
Thus, we assume $\setC_i^{\delta} = \{ s_0 \}$. 
Let $z_{s_0} \coloneqq Q^{-1} \left( \E_{s_0}\big[ \Asso_i \big] \right)$, and observe that 
\begin{align}
   \frac{ w_{s_0} }{\sum_{s \in \setA_i^{\delta}} w_s} \cdot \left\| \bv_{s_0} - \bbeta_i \right\| 
        &\leq \frac{ \E_{s_0}[\Asso_i] }{ \sum_{s \in \setA_i^{\delta}} \E_s[\Asso_i] } \cdot  \left( \left\| \bv_{s_0} - \bthetastar_{s_0} \right\| + \left\| \bthetastar_{s_0} - \bbeta_i \right\| \right) 
            \nonumber\\
        &\leq \frac{1}{|\setA_i^{\delta}|\cdot ( 1 - \kfit \cdot \bdw )} \cdot \Big( \E_{s_0} \big[ \Asso_i \cdot \big( \sfx - \bthetastar_{s_0} \big) \big] + \left\| \bthetastar_{s_0} - \bbeta_i \right\| \Big)
            \nonumber\\
        &\stackrel{(a)}{\leq} \frac{1}{|\setA_i^{\delta}|\cdot ( 1 - \kfit \cdot \bdw )} \cdot \Big( \phi(z_{s_0}) \cdot \std + \frac{2 \ktrue}{\delta} \cdot \std \Big)
            \nonumber\\
        &\leq \frac{1}{ 1 - \kfit \cdot \bdw } \cdot \frac{ \std }{ |\setA_i^{\delta} | } \left( \frac{1}{\sqrt{2\pi}} + \frac{2\ktrue}{\delta} \right),
            \label{eqn:prop4_step2C}
\end{align}
where (a) follows from Lemma \ref{lem:deviation_bd} and Corollary \ref{cor:proximity}.

\paragraph{Step 3. Concluding the proof} 
To conclude the proof, we insert the upper bounds \eqref{eqn:prop4_step2A}, \eqref{eqn:prop4_step2B}, and \eqref{eqn:prop4_step2C} into \eqref{eqn:prop4_step1} and observe that $\bdw \coloneqq 9 \left(\sqrt{2\pi} + 1 \right) \cdot \kfit^3 \cdot \delta \leq \frac{1}{2 \kfit}$ due to the assumption $\delta \leq \frac{1}{ 18 (\sqrt{2\pi} + 1 ) \cdot  \kfit^4 }$:
\begin{align*}
    \big\| \hbbeta_i - \bbeta_i \big\|
        &\leq 
            \frac{\kfit \cdot \bdw}{1 - \kfit \cdot \bdw} \cdot \left( \min_{s \in \setA_i^{\delta}} \max_{s' \in \setA_i^{\delta}} \| \bthetastar_s - \bthetastar_{s'} \| + 3 \cdot \std \right) 
            + \frac{|\setB_i^{\delta}|}{|\setA_i^{\delta}|} \cdot \frac{ 5 \bdw \cdot \deltacell(\B) + \std \big( \sqrt{d} + 4 \big)}{ 1 - \kfit \cdot \bdw }\\
            &\quad
            + \frac{1}{ 1 - \kfit \cdot \bdw } \cdot \frac{ \std }{ |\setA_i^{\delta} | } \left( \frac{1}{\sqrt{2\pi}} + \frac{2\ktrue}{\delta} \right)\\
        &\leq 
            \left( 2 \kfit \cdot \bdw \cdot \tdeltacell^{i, \delta}(\B) + 3 \cdot \std \right)
            + \frac{2 |\setB_i^{\delta}|}{|\setA_i^{\delta}|} \cdot \left( 5 \bdw \cdot \deltacell(\B) + \std \big( \sqrt{d} + 4 \big) \right)
            + \frac{ 2 \std }{ |\setA_i^{\delta} | } \left( \frac{1}{\sqrt{2\pi}} + \frac{2\ktrue}{\delta} \right)\\
        &\leq 
            18 \left(\sqrt{2\pi} + 1 \right) \cdot \kfit^3 \cdot \left( \frac{5\ktrue}{|\setA_i^{\delta}|} \cdot \deltacell(\B)
            + \kfit \cdot\tdeltacell^{i,\delta}(\B) \right) \cdot \delta 
            + \frac{  4\ktrue \cdot \std }{ |\setA_i^{\delta} | }  \cdot \frac{1}{\delta} \\
            &\quad    + \std \left\{ \frac{1}{|\setA_i^{\delta}|} \cdot \left( 2\ktrue \cdot (\sqrt{d}+4) + \sqrt{\frac{2}{\pi}} \right) + 3 \right\}.
\end{align*}
\end{proof}

\section{Deferred Proof of Proposition \ref{prop:cover_properties}}\label{sec:proof_covering}

\begin{proof}[Proof of Proposition \ref{prop:cover_properties}]
 In this proof, we prove the four claims one by one.

\bigskip
\textit{\underline{(1) Proof of Claim 1 (non-emptiness).}} 
We begin by showing the non-emptiness of $\cS^{\delta}_a$. 
First, it is clear from line 5 of Algorithm \ref{alg:partition} that $\cS^{\delta}_a \neq \emptyset$ for all $a \in [q_0]$. 
Second, to prove $\cS^{\delta}_a \neq \emptyset$ for all $a \in [q]\setminus[q_0]$, we assume $\exists a \in [q]\setminus[q_0]$ such that $\cE_{s_a}^{\delta} = \emptyset$ where $s_a \in [\ktrue]$ denotes the unique element in $\cT^{\delta}_a$. 
By this assumption, $\max_{i \in [\kfit]} \max_{j \in [\kfit]\setminus\{i\}} \frac{ \| \bbeta_i - \bbeta_j \| }{\sigma} \cdot \E_{s_a}\left[ \Asso_i \Asso_j \right] < \delta$. 
In addition, we observe that $s_a \not\in \setA_i^{\delta}$ for all $i \in [\kfit]$ because $s_a \in \cR^{\delta}$. 
These two observations together imply that $\iota_{\B}(\bthetastar_{s_a}) \neq i$ for all $i \in [\kfit]$, which is a contradiction. 
As a result, we conclude that $\cS^{\delta}_a \neq \emptyset$ for all $a \in [q]\setminus[q_0]$.
Next, it is easy to observe that $\cT^{\delta}_a \neq \emptyset$ for all $a \in [q]$ by construction; see line 6 and line 15 of Algorithm \ref{alg:partition}.

Moreover, it is clear from the construction (see lines 5 and 15 in Algorithm \ref{alg:partition}) that $|\cS^{\delta}_a|=1$ for all $a \in [q_0]$, and $|\cT^{\delta}_a|=1$ for all $a \in [q] \setminus [q_0]$. 
Lastly, we can verify that $|\cS^{\delta}_a|\geq2$ for all $a \in [q] \setminus [q_0]$ because $|\cE_s^{\delta}| \geq 2$ unless $\cE_s^{\delta} = \emptyset$ by definition, cf. \eqref{eqn:entangled}.

\bigskip
\textit{\underline{(2) Proof of Claim 2 (covering).}} 
It is clear from line 19 of Algorithm \ref{alg:partition} that $\bigcup_{a=0}^{\kfit} \cS^{\delta}_a = [\kfit]$. 

Next, we show that for all $s \in [\ktrue]$, there exists $a \in [q]$ such that $s \in \cT^{\delta}_a$. 
To this end, we choose an arbitrary $s \in [\ktrue]$ and consider two possibilities: (i) $\cE_s^{\delta} = \emptyset$; and (ii) $\cE_s^{\delta} \neq \emptyset$. 
If (i) is the case, then there exists $a \in [q_0]$ such that $s \in \cT^{\delta}_a = \setA^{\delta}_{i}$ where $i_a$ is the unique element of $\cS^{\delta}_a$; see lines 5 and 6 of Algorithm \ref{alg:partition}. 
If (ii) is the case, then there exists $a \in [q] \setminus [q_0]$ such that $\cT^{\delta}_a = \{ s \}$ by construction.

\bigskip
\textit{\underline{(3) Proof of Claim 3 (inclusion of Voronoi centers).}} 
For $a \in [q_0]$, if $s \in \cT^{\delta}_a$, then $s \in \setA^{\delta}_{i_a}$ where $i_a = \iota_{\B}(\bthetastar_s)$ is the unique element in $\cS^{\delta}_a$. 
Thus, it is clearly true by construction

Next, we let $a \in [q] \setminus [q_0]$ and let $s_a$ denote the unique element in $\cT^{\delta}_a$. 
Then it suffices to show that $i = \iota_{\B}(\bthetastar_{s_a}) \in \cE^{\delta}_{s_a}$. 
If we assume otherwise, then there must exist $b \in [q] \setminus [q_0]$ such that $b \neq a$ and $i \in \cS^{\delta}_b$.  
Then it follows from Corollary \ref{cor:proximity} that 
\[
    \frac{\| \bbeta_i - \bthetastar_{s_b} \|}{\sigma} 
        \leq  \frac{2\ktrue}{\delta}
\]
where $s_b$ is the unique element in $\cT^{\delta}_b$. 
Also, we get 
\[
    \frac{\| \bbeta_i - \bthetastar_{s_a} \|}{\sigma}
        \leq \min_{j \in \cS^{\delta}_a}\frac{\| \bbeta_j - \bthetastar_{s_a} \|}{\sigma} 
        \leq  \frac{2\ktrue}{\delta}
\]
because $\bthetastar_{s_a} \in \vor_i$. 
Then we get
\begin{align*}
    \left\| \bthetastar_{s_a} - \bthetastar_{s_b} \right\|
        &\leq \left\| \bbeta_i - \bthetastar_{s_a} \right\| + \left\| \bbeta_i - \bthetastar_{s_b} \right\| \\
        &\leq \frac{ 4 \ktrue \cdot \sigma }{ \delta } \\
        &< \deltamin
\end{align*}
where the last inequality follows from the premise of the theorem that $\delta > \frac{ 4 \ktrue \cdot \sigma }{ \deltamin }$. 
This is a contradiction due to the definition of $\deltamin$ in \eqref{eq:separation}.

\bigskip
\textit{\underline{(4) Proof of Claim 4 (disjointness of $\bbT^{\delta}$).}} 
\begin{enumerate}[label=(\alph*)]
    \item
    Let $a \in [q_0]$.
    \begin{itemize}
        \item
        Suppose that $b \in [q_0] \setminus \{a\}$. 
        Assume that $\cT^{\delta}_a \cap \cT^{\delta}_b \neq \emptyset$, i.e., $\exists s \in \cT^{\delta}_a \cap \cT^{\delta}_b \neq \emptyset$. 
        Letting $i_a$ and $i_b$ denote the unique elements of $\cS^{\delta}_a$ and $\cS^{\delta}_b$, respectively, we observe that $s \in \setA^{\delta}_{i_a} \cap \setA^{\delta}_{i_b}$ by construction; see lines 5 and 6 of Algorithm \ref{alg:partition}. 
        Then it follows that 
        \begin{align*}
            \E_{s}\left[ \Asso_{i_a} \right]
                &\stackrel{(a)}{\geq}  1 - 9 \big(\sqrt{2\pi} + 1 \big) \cdot \kfit^4 \cdot \delta
                    \\
                &\stackrel{(b)}{>} \frac{1}{2},
        \end{align*}
        where (a) is by Proposition \ref{prop:exclusive} (Claim 2); and (b) is by the premise that $\delta < \frac{1}{ 18 (\sqrt{2\pi} + 1 ) \cdot \kfit^4 }$. 
        Likewise, we obtain $\E_{s}\left[ \Asso_{i_b} \right] > \frac{1}{2}$, thereby, $\E_{s} \left[ \Asso_{i_a} + \Asso_{i_b} \right] > 1$, which is a contradiction. 
        Therefore, $\cT^{\delta}_a \cap \cT^{\delta}_b = \emptyset$ for all $b \in [q_0] \setminus\{ a\}$. 

        \item
        Suppose that $b \in [q] \setminus [q_0]$. 
        We observe that $\cS_b^{\delta} \subseteq \cR^{\delta} = [\ktrue] \setminus \big(\bigcup_{a'=1}^{q_0} \cT^{\delta}_{a'} \big)$ for any $b \in [q] \setminus [q_0]$, by construction (see line 10 and line 15 in Algorithm \ref{alg:partition}). 
        Thus, $\cT^{\delta}_a \cap \cT^{\delta}_b = \emptyset$.
    \end{itemize}

    \item
    Let $a \in [q]\setminus [q_0]$ and $b \in [q] \setminus ( [q_0] \cup \{i\})$.
    It is obvious from construction that $\cT^{\delta}_a \cap \cT^{\delta}_b = \emptyset$.
\end{enumerate}

\bigskip
\textit{\underline{(5) Proof of Claim 5 (partial disjointness of $\bbS^{\delta}$).}} 
\begin{enumerate}[label=(\alph*)]
    \item
    Let $a = 0$. It is obvious that $\cS^{\delta}_a \cap \cS^{\delta}_b = \emptyset$ for all $b \in [q]$ by definition; see line 19 of Algorithm \ref{alg:partition}.

    \item
    Let $a \in [q_0]$. 
    \begin{itemize}
        \item
        Suppose that $b \in [q_0] \setminus \{a\}$. 
        It is clear that $\cS^{\delta}_a \cap \cS^{\delta}_b = \emptyset$ for all $b \in [q_0] \setminus\{a\}$ from line 5 of Algorithm \ref{alg:partition}.  
        
        \item
        For the case $b \in [q] \setminus [q_0]$, we do not claim disjointness for $\bbS^{\delta}$.
    \end{itemize}

    \item
    Let $a \in [q]\setminus [q_0]$ and $b \in [q] \setminus ( [q_0] \cup \{i\})$. 
    We assume there exists $i \in [\kfit]$ such that $i \in \cS^{\delta}_a \cap \cS^{\delta}_b$.
    Letting $s_a$ and $s_b$ denote the unique elements in $\cT^{\delta}_a$ and $\cT^{\delta}_b$, respectively, we have
    \begin{align*}
        &\max_{j \in [\kfit]\setminus \{i\}} \frac{\| \beta_i - \beta_j \|}{\sigma} \cdot \E_{s_a}\left[ \Asso_i \Asso_j \right] \geq \delta
        \quad\text{and}\\
        &\max_{j \in [\kfit]\setminus \{i\}} \frac{\| \beta_i - \beta_j \|}{\sigma} \cdot \E_{s_b}\left[ \Asso_i \Asso_j \right] \geq \delta.
    \end{align*}
    Then it follows from Corollary \ref{cor:proximity} that 
    \begin{align*}
        \left\| \bthetastar_{s_a} - \bthetastar_{s_b} \right\|
            &\leq \left\| \bbeta_i - \bthetastar_{s_a} \right\| + \left\| \bbeta_i - \bthetastar_{s_b} \right\| \\
            &\leq \frac{ 4 \ktrue \cdot \sigma }{ \delta }\\
            &< \deltamin
    \end{align*}
    where the last inequality follows from the premise of the theorem that $\delta > \frac{ 4 \ktrue \cdot \sigma }{ \deltamin }$.  
    This is a contradiction due to the definition of $\deltamin$ in \eqref{eq:separation}: $\left\| \bthetastar_{s_a} - \bthetastar_{s_b} \right\| \geq \deltamin$. Therefore, $\cS^{\delta}_a \cap \cS^{\delta}_b = \emptyset$.
\end{enumerate}
\end{proof}
\section{Deferred Proof of Proposition \ref{prop:distillation}}\label{sec:proof_distillation}

\begin{proof}[Proof of Proposition \ref{prop:distillation}]
    Our proof is based on the induction argument. 
    We let $q^{(t)}, q_0^{(t)}, \bbS^{(t)}, \bbT^{(t)}$ denote the values and the collections of sets after the $t$-th iteration of the outermost loop in Algorithm \ref{alg:distillation}. 
    Assuming the conclusions hold at the end of the $t$-th iteration, we consider the two possible scenarios at the $(t+1$)-th iteration.

    \medskip
    \underline{Scenario A:} $\cS_a \setminus \bigcup_{b \in \setconf_a} \cS_b \neq \emptyset$, i.e., Line 3 of Algortihm \ref{alg:distillation} is executed.
    \begin{enumerate}
        \item 
        Claim 1:
        \begin{enumerate}
            \item 
            Comparing $\bbS^{(t+1)}$ to $\bbS^{(t)}$, we only have made one $\cS_{a}$ smaller (but $\cS_{a}$ is still non-empty). 
            To show that the new $\bbS^{(t+1)} \setminus \{ \cS_0 \}$ is still a $\big( \{ \cS_a \}_{a=1}^{q_0}, \{ \cS_a \}_{a=q_0+1}^{q} \big)$-quasi-partition of $[\kfit] \setminus \cS_0$, it suffices to verify that $\bigcup_{\cS \in \bbS\setminus\{\cS_{0}\}} \cS = [\kfit]\setminus\cS_{0}$. 
            This is true because $\bigcup_{b \in \setconf_a} \cS_b$ still contains the elements that are removed from $\cS_{b}$.
            \item 
            $\bbT^{(t+1)}$ is a partition because $\bbT^{(t)}$ is a partition and $\bbT^{(t+1)} = \bbT^{(t)}$.
            \item 
            Claims 1-(c) and 1-(d) of Theorem \ref{thm:master} hold as the sets $\cS_{a'},~a'\in[q]\setminus\{a\}$ and $\cT_{b'},~b'\in[q]$ are unaffected.
        \end{enumerate}
        \item
        Claim 2 continues to hold as the sets $\cS_{a'},~a'\in[q]\setminus\{a\}$ and $\cT_{b'},~b'\in[q]$ are unaffected.
    
        \item
        Claim 3:
        \begin{enumerate}
            \item 
            Equation \eqref{eqn:approx_bound.1} still holds because the sets $\cS_{a},\cT_{a}$ remain unchanged for all $a\in[q_{0}]$.
            \item 
            Equation \eqref{eqn:approx_bound.2} still holds because 
            \[
                \max_{i \in \cS_{a}\setminus\bigcup_{b\in \setconf_a}\cS_{b} }\left\| \bbeta_i - \bthetastar_{s_a} \right\|
                \leq \max_{i \in \cS_{a} }\left\| \bbeta_i - \bthetastar_{s_a} \right\|.
            \]
        \end{enumerate}

    \end{enumerate}

    \medskip
    \underline{Scenario B:} $\cS_a \setminus \bigcup_{b \in \setconf_a} \cS_b = \emptyset$, i.e., Lines 5--12 of Algorithm \ref{alg:distillation} are executed.
    \begin{enumerate}
        \item 
        Claim 1:
        \begin{enumerate}
            \item 
            The only change to $\bbS$ was removing $\cS_{b}$, so we only need
            to check the sets in the new $\bbS\setminus\{\cS_{0}\}$ still cover
            $[\kfit]\setminus\cS_{0}$. This is true since the condition $\cS_{b}\setminus\bigcup_{a\in A_{b}}\cS_{a}=\emptyset$
            implies that all the elements in $\cS_{b}$ are still contained in
            other $\cS_{a}$'s.
            \item 
            $\bbT^{(t)}$ is a partition of $[\ktrue]$ and we only merged two sets in $\bbT^{(t)}$. 
            Thus, $\bbT^{(t+1)}$ is still a partition.
            \item 
            It is trivial to see that Claims 1-(c) and 1-(d) of Theorem \ref{thm:master} continue to hold.
        \end{enumerate}
        \item
        It is straightforward that Claim 2 continues to hold.
    
        \item
        Claim 3:
        \begin{enumerate}
            \item 
            Recall from Lines 5--6 of Algorithm \ref{alg:partition} that $\cS_{b_0}=\{i\}$ where $i = \arg\min \cS_a$. 
            By the premise for $\bbS^{(t)}$, we have 
            \begin{itemize}
                \item
                $\frac{1}{\sigma}\big\| \bbeta_{i} - \frac{1}{\left|\cT_{b_0}\right|}\sum_{s\in\cT_{b_0}}\bthetastar_{s}\big\| \leq \varepsilon_1$ where
                $\varepsilon_1 = \left(2^6 \cdot 3^{3} \cdot (\sqrt{2\pi} + 1) \cdot \ktrue \cdot \kfit^3 \cdot (\kfit + \ktrue) \cdot \frac{\deltamax}{\sigma} \right)^{1/2} + 2 \deff^{1/2}$,
                and
                \item
                $\frac{1}{\sigma} \big\| \bbeta_{i} - \bthetastar_{s_{a}} \big\|$, where $\varepsilon_2 = \left( 2^2 \cdot 3^3 \cdot (\sqrt{2\pi} + 1) \cdot \ktrue \cdot \kfit^{3} \cdot(\kfit + \ktrue) \cdot \frac{\deltamax}{\sigma}\right)^{1/2}$ and $s_{a}$ denotes the unique element in $\cT_{a}$. 
            \end{itemize}
            Therefore, it follows that 
            \begin{align*}
                \frac{1}{\sigma} \left\| \bbeta_{i} - \frac{1}{\left|\cT_{a}\cup\cT_{b_0}\right|}\sum_{s\in\cT_{a}\cup\cT_{b_0}}\bthetastar_{s} \right\|
                    &\leq \frac{|\cT_a|}{\left|\cT_{a}\cup\cT_{b_0}\right|} \cdot \frac{1}{\sigma} \left\| \bbeta_{i} - \frac{1}{\left|\cT_{a}\right|}\sum_{s\in\cT_{a}}\bthetastar_{s} \right\| \\
                        &\quad
                        + \frac{|\cT_{b_0}|}{\left|\cT_{a}\cup\cT_{b_0}\right|} \cdot \frac{1}{\sigma} \left\| \bbeta_{i} - \frac{1}{\left|\cT_{b_0}\right|}\sum_{s\in\cT_{b_0}}\bthetastar_{s} \right\|\\
                    &= \frac{1}{\left|\cT_{b_0}\right| + 1} \cdot \frac{1}{\sigma} \left\| \bbeta_{i} - \bthetastar_{s_a} \right\|
                        + \frac{|\cT_{b_0}|}{\left|\cT_{b_0}\right| + 1} \cdot \frac{1}{\sigma} \left\| \bbeta_{i} - \frac{1}{\left|\cT_{b_0}\right|}\sum_{s\in\cT_{b_0}}\bthetastar_{s} \right\|\\
                    &\leq \frac{1}{\left|\cT_{b_0}\right| + 1} \cdot  \left( |\cT_{\b_0}| \cdot \varepsilon_1 + \varepsilon_2 \right)\\
                    &\leq \max\{ \varepsilon_1, \varepsilon_2 \} = \varepsilon_1.
            \end{align*}
            \item 
            Equation \eqref{eqn:approx_bound.2} holds because it is only $\cS_{a},\cT_{a}$ that are removed during the $(t+1)$-th iteration, and all the other sets $\cS_{a'},\cT_{a'}$, $a'\in[q]\setminus[q_{0}]\setminus\{a\}$ remain the same.
        \end{enumerate}
    \end{enumerate}
\end{proof}
\section{Deferred Proofs of Technical Lemmas in Section \ref{sec:useful_lemmas}}

\subsection{Proof of Lemma \ref{lem:separation}}\label{sec:proof_lemma.tech1}
\begin{proof}[Proof of Lemma \ref{lem:separation}]
    In this proof we let $\E_0$ denote $\E_{\sfx \sim \cN(0,1)}$. 
    By Stein's identity (Lemma \ref{lem:stein}) and the premise that $\beta \geq |\alpha|$, we have
    \begin{align}
        \E_0 \left[ \psiab(\sfx) \cdot \left( \beta - \sfx \right) \right] 
            &= \beta \cdot \E_0 \left[ \psiab(\sfx)^2 \right] + \alpha \cdot \E_0 \left[ ( 1 - \psiab(\sfx)) \cdot \psiab(\sfx) \right]   \nonumber\\
            &\geq \beta \cdot \E_0 \left[ \psiab(\sfx) \cdot \big( 2 \psiab(\sfx) - 1 \big) \right].  \label{eqn:separation.eq1}
    \end{align}

    Letting $c \coloneqq \frac{\alpha + \beta}{2}$ and $\delta \coloneqq \frac{\beta - \alpha}{2}$, we may write
    \[
        \psiab(x) \cdot \left( 2\psiab(x) - 1 \right)
            = e^{-\frac{(x-\beta)^2}{2}} \cdot g(x)
    \]
    where
    \begin{equation}\label{eqn:g_function}
    \begin{aligned}
        g(x) &= \frac{ e^{(x-c)\delta} - e^{-(x-c)\delta} }{ \left( e^{(x-c)\delta} + e^{-(x-c)\delta} \right) }
            \cdot \frac{1}{\left( e^{-(x-c-\delta)^2/2} + e^{-(x-c+\delta)^2/2} \right) }.
    \end{aligned}
    \end{equation}
    Observe that $g( c + z ) = - g( c - z )$. By a change of variable $z = x - c$, we obtain
    \begin{align}
        \E_0 \left[ \psiab(\sfx) \cdot \big( 2\psiab(\sfx) - 1 \big) \right]  
            &= \frac{1}{\sqrt{2\pi}} \int_{-\infty}^{\infty} e^{-(x-\beta)^2/2} \cdot g(x) \cdot e^{-x^2/2} ~\ddup x      \nonumber\\
            &= \frac{1}{\sqrt{2\pi}} \int_{-\infty}^{\infty} g(z + c) \cdot e^{-(z + c -\beta)^2/2} \cdot e^{-(z+c)^2/2} ~\ddup z     \nonumber\\
            &= \frac{1}{\sqrt{2\pi}} \int_{0}^{\infty} g(z + c) \cdot \underbrace{ e^{-(z + c -\beta)^2/2} \cdot e^{-(z+c)^2/2} }_{\eqqcolon h_1(z)} \ddup z \nonumber\\
                &\quad- \frac{1}{\sqrt{2\pi}} \int_{0}^{\infty} g(z + c) \cdot \underbrace{ e^{-(-z + c -\beta)^2/2} \cdot e^{-(-z+c)^2/2} }_{\eqqcolon h_2(z)} ~\ddup z.     \label{eqn:lem_separation_int}
    \end{align}
    Note that $\frac{h_1(z)}{h_2(z)} = e^{-2\alpha z} \geq 1$, $\forall z \geq 0$ because $\alpha < 0$ by assumption. 
    Because $g(z + c ) \geq 0$ for all $z \geq 0$, we have 
    \begin{align*}
        \E_0 \left[ \psiab(\sfx) \cdot \big( 2\psiab(\sfx) - 1 \big) \right]
            &\geq \frac{1}{\sqrt{2\pi}} \int_{-1/\alpha}^{\infty} g(z + c) \cdot \big[ h_1(z) - h_2(z) \big] ~\ddup z.
    \end{align*}
    Then we observe that by definition, cf. \eqref{eqn:g_function}, for all $z \geq -1/\alpha > 1/\delta$, 
    \begin{align*}
        g( z + c )
            &= \frac{ e^{z\delta} - e^{-z\delta} }{ \left( e^{z\delta} + e^{-z\delta} \right) \cdot \left( e^{-(z-\delta)^2/2} + e^{-(z+\delta)^2/2} \right) }\\
            &\geq \frac{ e - e^{-1}}{e + e^{-1}} \cdot \frac{1}{2 e^{-(z-\delta)^2/2}}\\
            &\geq \frac{1}{4} e^{(z-\delta)^2/2}.
    \end{align*}
    Moreover, if $z \geq -1/\alpha$, then $\frac{h_1(z)}{h_2(z)} = e^{-2\alpha z} = e^2 \geq 2$, and therefore,
    \begin{align}
        \E_0 \left[ \psiab(\sfx) \cdot \big( 2\psiab(\sfx) - 1 \big) \right]
            &\geq \frac{1}{\sqrt{2\pi}} \int_{-1/\alpha}^{\infty} \frac{1}{4} e^{(z-\delta)^2/2} \cdot \frac{1}{2} h_1(z) ~\ddup z  \nonumber\\
            &= \frac{1}{8\sqrt{2\pi}} \int_{-1/\alpha}^{\infty} e^{(z-\delta)^2/2} \cdot e^{-(z + c -\beta)^2/2} \cdot e^{-(z+c)^2/2}  ~\ddup z \nonumber\\
            &= \frac{1}{8\sqrt{2\pi}} \int_{-1/\alpha}^{\infty} e^{-(z-c)^2/2} ~\ddup z \nonumber\\
            &= \frac{1}{8} \cdot Q\left( c - \frac{1}{\alpha} \right).  \label{eqn:separation.eq2}
    \end{align}
    Combining \eqref{eqn:separation.eq1} and \eqref{eqn:separation.eq2} completes the proof.    
\end{proof}

\subsection{Proof of Lemma \ref{lem:var_bound}}
\begin{proof}[Proof of Lemma \ref{lem:var_bound}]
    Recall the definition of $\psiab$ from \eqref{eqn:psiab} and observe that
    \begin{equation}\label{eqn:psiab_alt}
    \begin{aligned}
        \psiab(x) 
            &= \frac{e^{-(x-\beta)^2/2}}{e^{-(x-\alpha)^2/2} + e^{-(x-\beta)^2/2}}\\
            &= \frac{1}{1 + e^{ -(\beta - \alpha)\cdot \left( x - \frac{\alpha+\beta}{2}\right) }}.
    \end{aligned}
    \end{equation}
    In this proof we let $\E_0$ denote $\E_{\sfx \sim \cN(0,1)}$ and $\var_{0}$ denote $\var_{\sfx \sim \cN(0,1)}$. 
    Letting $c \coloneqq \frac{\alpha + \beta}{2}$, $\delta \coloneqq \beta - \alpha$, and $\mu = \E_0 \left[ \psiab(\sfx) \right]$, we may assume $c \geq 0$. 
    Also, we may assume $\delta > 0$; if $\delta < 0$, we consider $\psiab'(x) = 1 - \psiab(x)$ instead of $\psiab(x)$ as $\var_0\left[ \psiab(\sfx) \right] = \var_0\left[ \psiab'(\sfx) \right]$. 
    We have
    \begin{align*}
        \sqrt{2\pi} \cdot \var_{0}\left(\psiab(\sfx) \right)
            &= \sqrt{2 \pi} \cdot \E_0 \left[ \left(\psiab(\sfx) - \mu \right)^2 \right]\\
            &\stackrel{(a)}{=} \int_{-\infty}^{\infty}  \left( \frac{1}{1 + e^{ - \delta \cdot \left( x - c \right) }} - \mu \right)^2 \cdot e^{-x^2/2} ~\ddup x    \\
            &=  \int_{-\infty}^{c}  \left( \frac{1}{1 + e^{ - \delta \cdot \left( x - c \right) }} - \mu \right)^2 \cdot e^{-x^2/2} ~\ddup x
                +  \int_{c}^{\infty}  \left( \frac{1}{1 + e^{ - \delta \cdot \left( x - c \right) }} - \mu \right)^2 \cdot e^{-x^2/2} ~\ddup x\\
            &=  \int_{-c}^{\infty}  \left( \frac{1}{1 + e^{ \delta \cdot \left( x + c \right) }} - \mu \right)^2 \cdot e^{-x^2/2} ~\ddup x
                +  \int_{c}^{\infty}  \left( \frac{1}{1 + e^{ - \delta \cdot \left( x - c \right) }} - \mu \right)^2 \cdot e^{-x^2/2} ~\ddup x\\
            &\stackrel{(b)}{\geq} \int_{c}^{\infty} \Bigg[ \left( \frac{1}{1 + e^{ \delta \cdot \left( x + c \right) }} - \mu \right)^2 
                + \left( \frac{1}{1 + e^{ - \delta \cdot \left( x - c \right) }} - \mu \right)^2 \Bigg] \cdot e^{-x^2/2} ~\ddup x,
    \end{align*}
    where (a) follows from \eqref{eqn:psiab_alt}, and (b) follows from that $c \geq 0$ and $\left( \frac{1}{1 + e^{ \delta \cdot \left( x + c \right) }} - \mu \right)^2 \geq 0$ for all $x \in [-c, c]$. 
    Because $u^2 + v^2 \geq \frac{1}{2} (u-v)^2$ for all $u, v \in \real$, it follows that
    \begin{align}
        \sqrt{2\pi} \cdot \var_{0}\left(\psiab(\sfx) \right) 
            &\geq \frac{1}{2} \int_{c}^{\infty} \left( \frac{1}{1 + e^{ - \delta \cdot \left( x - c \right) }} -  \frac{1}{1 + e^{ \delta \cdot \left( x + c \right) }}  \right)^2 \cdot e^{-x^2/2} ~\ddup x
                \nonumber\\
            &= \frac{1}{2} \int_{c}^{\infty} \frac{ \left( e^{ \delta \cdot \left( x + c \right) } - e^{ - \delta \cdot \left( x - c \right)} \right)^2 }{ \left( 1 + e^{ - \delta \cdot \left( x - c \right)} \right)^2 \left( 1 + e^{ \delta \cdot \left( x + c \right) } \right)^2}  \cdot e^{-x^2/2} ~\ddup x
                \nonumber\\
            &\stackrel{(a)}{\geq} \frac{1}{8} \int_{c}^{\infty} \frac{ \left( e^{ \delta \cdot \left( x + c \right) } - 1 \right)^2 }{ \left( 1 + e^{ \delta \cdot \left( x + c \right) } \right)^2}  \cdot e^{-x^2/2} ~\ddup x
                \nonumber\\
            &\geq \frac{1}{8} \int_{c+\epsilon_1}^{c + \epsilon_2} \left( \frac{  e^{ \delta \cdot \left( x + c \right) } - 1  }{ e^{ \delta \cdot \left( x + c \right) } + 1} \right)^2  \cdot e^{-x^2/2} ~\ddup x,    \label{eqn:var_lower.1}
    \end{align}
    for any $\epsilon_1, \epsilon_2 \in \real$ such that $0 \leq \epsilon_1 \leq \epsilon_2$. 
    Note that we used $e^{ -\delta \cdot \left( x - c \right) } \leq 1 \leq e^{ \delta \cdot \left( x + c \right) }, ~\forall x \geq c$ to get the inequality (a). 
    Since the function $x \mapsto \frac{  e^{ \delta \cdot \left( x + c \right) } - 1  }{ e^{ \delta \cdot \left( x + c \right) } + 1}$ is non-decreasing, we have
    \begin{align}
        \int_{c+\epsilon_1}^{c + \epsilon_2} \left( \frac{  e^{ \delta \cdot \left( x + c \right) } - 1  }{ e^{ \delta \cdot \left( x + c \right) } + 1} \right)^2  \cdot e^{-x^2/2} ~\ddup x  
            & \geq (\epsilon_2 - \epsilon_1) \cdot \left( \frac{  e^{ \delta \cdot \left( 2c + \epsilon_1 \right) } - 1  }{ e^{ \delta \cdot \left( 2c + \epsilon_1 \right) } + 1} \right)^2 \cdot e^{-(c + \epsilon_2)^2/2}
                \nonumber\\
            & \geq (\epsilon_2 - \epsilon_1) \cdot \left( \frac{ \delta \cdot \left( 2c + \epsilon_1 \right)  }{ 2 e^{ \delta \cdot \left( 2c + \epsilon_1 \right)} } \right)^2 \cdot e^{-(c + \epsilon_2)^2/2}
                \nonumber\\
            & \stackrel{(a)}{=} \left( c + \delta \right) \cdot \left( \frac{ \delta \cdot \left( 2c + \delta \right)  }{ 2 e^{ \delta \cdot \left( 2c + \delta \right)} } \right)^2 \cdot e^{-2(c + \delta)^2}
                \nonumber\\
            & \geq \frac{1}{4} \delta^2 (c + \delta)^3 \cdot e^{-4 (c + \delta)^2},
                \label{eqn:var_lower.2}
    \end{align}
    where (a) is attained by choosing $\epsilon_1 = \delta,~\epsilon_2 = c + 2\delta$. 
    Inserting the lower bound \eqref{eqn:var_lower.2} to \eqref{eqn:var_lower.1}, we obtain
    \[
        \var_{0}\left(\psiab(\sfx) \right)
            \geq \frac{1}{32 \sqrt{2\pi}} \delta^2 (c + \delta)^3 \cdot e^{-4 (c + \delta)^2}.
    \]
\end{proof}

\subsection{Proof of Lemma \ref{lem:Psi_domination}}
\begin{proof}[Proof of Lemma \ref{lem:Psi_domination}]
    Let $\alpha \geq 0$ be a parameter whose value will be determined later in this proof. 
    Then we write
    \begin{equation}\label{eqn:asso_partition}
    \begin{aligned}
        \E_{s}\left[\Asso_{j}\right]
            &= \E_{s}\left[\Asso_{j} \cdot \indic\left\{ \left| {\sf x} - \bbeta_j \right| - \left| {\sf x} - \bbeta_{i_s} \right| \geq \alpha \right\}\right]
                + \E_{s}\left[\Asso_{j} \cdot \indic\left\{ \left| {\sf x} - \bbeta_j \right| - \left| {\sf x} - \bbeta_{i_s} \right| < \alpha \right\}\right].
    \end{aligned}
    \end{equation}
    We establish upper bounds for the two terms on the right-hand side of \eqref{eqn:asso_partition} separately.

    First, for $x$ such that $\left| x - \bbeta_j \right| - \left| x - \bbeta_{i_s} \right| \geq \alpha$, we have 
    $\left| x - \bbeta_j \right|^2 \geq \left| x - \bbeta_{i_s} \right|^2 + \alpha^2$, and thus,
    \begin{align*}
        \asso_j(x) 
            &= \frac{ e^{-|x - \bbeta_j|^2/2} }{ \sum_{i' \in [\kfit]} e^{-|x - \bbeta_{i'}|^2/2} }\\
            &\leq \frac{ e^{-|x - \bbeta_j|^2/2} }{ e^{-|x - \bbeta_j|^2/2} + e^{-|x - \bbeta_{i_s}|^2/2} }\\
            &\leq \frac{1}{1 + e^{\alpha^2/2}}.
    \end{align*}
    Therefore,
    \begin{align}
        \E_{s}\left[\Asso_{j} \cdot \indic\left\{ \left| {\sf x} - \bbeta_j \right| - \left| {\sf x} - \bbeta_{i_s} \right| \geq \alpha \right\}\right]
            &\leq \E_{s}\left[ \frac{1}{1 + e^{\alpha^2/2}} \cdot \indic\left\{ \left| {\sf x} - \bbeta_j \right| - \left| {\sf x} - \bbeta_{i_s} \right| \geq \alpha \right\} \right] \nonumber\\
            &\qquad\leq  \frac{1}{1 + e^{\alpha^2/2}}.    \label{eqn:asso_partition.1}
    \end{align}

    Second, we observe that 
    \begin{align*}
        \left\{ x \in \real: \left| x - \bbeta_j \right| - \left| x - \bbeta_{i_s} \right| < \alpha \right\}
            &\subseteq \left\{ x \in \real: \left\langle x, \frac{\bbeta_j - \bthetastar_s}{|\bbeta_j - \bthetastar_s|} \right\rangle \geq \frac{\delta_j}{2} - \alpha \right\} \eqqcolon S_{\alpha}.
    \end{align*}
    Therefore, 
    \begin{align}
        \E_{s}\left[\Asso_{j} \cdot \indic\left\{ \left| {\sf x} - \bbeta_j \right| - \left| {\sf x} - \bbeta_{i_s} \right| < \alpha \right\}\right] 
            &\leq \E_{s}\left[\Asso_{j} \cdot \indic\left\{ {\sf x} \in S_{\alpha} \right\}\right]  \nonumber\\
            &\leq \E_{s}\left[ \indic\left\{ {\sf x} \in S_{\alpha} \right\}\right] \nonumber\\
            &= Q\left( \frac{\delta_j}{2} - \alpha \right)   \label{eqn:asso_partition.2}
    \end{align}
    where $Q$ is the Gaussian Q-function, cf. \eqref{eqn:Gaussian_Q}.

    Plugging in \eqref{eqn:asso_partition.1} and \eqref{eqn:asso_partition.2} to \eqref{eqn:asso_partition} and using the $Q$-function upper bound in \eqref{eqn:Q_bounds}, we obtain
    \[
        \E_{s}\left[\Asso_{j}\right]
            \leq \frac{1}{1 + e^{\alpha^2/2}} + \frac{1}{\frac{\delta_j}{2} - \alpha} \frac{1}{\sqrt{2\pi}} e^{-\frac{1}{2} ( \frac{\delta_j}{2} - \alpha )^2}.
    \]
    Choosing $\alpha = \frac{\delta_j}{4}$ and further simplifying this upper bound, we complete the proof.
    
\end{proof}

\subsection{Proof of Lemma \ref{lem:accu_domination}}
\begin{proof}[Proof of Lemma \ref{lem:accu_domination}]
    Recall that $\bbeta_i = \frac{\E_{*}\left[\Asso_{i}\sf x\right]}{\E_{*}\left[\Asso_{i}\right]}$, cf. \eqref{eq:station_cond}, for all $i \in [\kfit]$ and $\bthetastar_s = \E_{s}[\textsf{x}]$, for all $s \in [\ktrue]$. 
    Moreover, we may assume $\frac{1}{|\cT|} \sum_{s \in \cT} \bthetastar_s = \bzero$ without loss of generality (see Remark \ref{rem:coordinate}). 
    Thus, 
    \begin{equation}\label{eqn:accu_error.0}
        \bbeta_i - \frac{1}{|\cT|} \sum_{s \in \cT} \bthetastar_s
            = \frac{\sum_{s \in [\ktrue]}\E_{s}\left[\Asso_{i}\sf x\right]}{\sum_{s \in [\ktrue]} \E_{s}\left[\Asso_{i}\right]}.
    \end{equation}
    Then we observe that
    \begin{align}
        \sum_{s \in [\ktrue]}\E_{s}\left[\Asso_{i}\sf x\right] 
            & = \sum_{s \in \cT}\E_{s}\left[\Asso_{i}\sf x\right] + \sum_{s \in [\ktrue] \setminus \cT}\E_{s}\left[\Asso_{i}\sf x\right]  \nonumber\\
            &= \sum_{s \in \cT}\E_{s}[\sfx] 
                + \underbrace{\left( \sum_{s \in \cT} \sum_{j \in [\kfit]\setminus\{i\}} \E_{s}\left[\Asso_{j}\sf x\right] + \sum_{s \in [\ktrue] \setminus \cT}\E_{s}\left[\Asso_{i}\sf x\right] \right)}_{\eqqcolon \epsilon_1}    \label{eqn:accu_numerator}
    \end{align}
    because $\sum_{j \in [\kfit]} \Asso_j = 1$ with probability $1$ (w.r.t. all $s$). Likewise, we can see that 
    \begin{equation}\label{eqn:accu_denominator}
    \begin{aligned}
        \sum_{s \in [\ktrue]} \E_{s}\left[\Asso_{i}\right]
            &= \sum_{s \in \cT}\E_{s} 1 
                + \underbrace{\left( \sum_{s \in \cT} \sum_{j \in [\kfit]\setminus\{i\}} \E_{s}\left[\Asso_{j}\right] + \sum_{s \in [\ktrue] \setminus \cT}\E_{s}\left[\Asso_{i} \right] \right)}_{\eqqcolon\epsilon_2}.
    \end{aligned}
    \end{equation}
    Combining \eqref{eqn:accu_numerator} and \eqref{eqn:accu_denominator} with \eqref{eqn:accu_error.0}, we get
    \begin{equation}\label{eqn:accu_error}
        \bbeta_i - \frac{1}{|\cT|} \sum_{s \in \cT} \bthetastar_s
            = \frac{\sum_{s \in \cT}\E_{s}[\sfx] + \epsilon_1}{\sum_{s \in \cT}\E_{s} 1 + \epsilon_2}
            = \frac{\epsilon_1}{|\cT| + \epsilon_2}
    \end{equation}
    because $\sum_{s \in \cT}\E_{s}[\sfx] = \sum_{s \in \cT} \bthetastar_s = \bzero$.

    Next, we argue that the ``perturbation terms'' --- $\epsilon_1$ and $\epsilon_2$ in \eqref{eqn:accu_numerator} and \eqref{eqn:accu_denominator} --- have small norms. 
    To this end, we begin by observing that 
    \begin{align*}
        \left\| \E_s[ \Asso_j \sfx] \right\|
            &\leq \left( \E_s[ \Asso_j^2 ] \cdot \E_s \left[ \|\sfx\|^2 \right] \right)^{1/2}   &&\because\text{Cauchy-Schwarz}\\
            &\leq \left( \E_s[ \Asso_j ] \cdot \E_s \left[ \|\sfx\|^2 \right] \right)^{1/2}     &&\because 0 \leq \Asso_j \leq 1.
    \end{align*}
    It follows from the premise \eqref{eqn:accu_premise} and Lemma \ref{lem:Psi_domination} that for any $(s,j) \in \left( \cT \times [\kfit]\setminus\{i\} \right) \cup \left( [\ktrue] \setminus \cT \times \{i\} \right)$,
    \[
        \E_{s}\left[\Asso_{j}\right] 
            \leq \left( 1 + \frac{4}{\sqrt{2\pi} \cdot \delta } \right) e^{- \frac{\delta^2}{32}}.
    \]
    Also, we can easily observe that
    \begin{align*}
        \E_s \left[ \|\sfx\|^2 \right] 
            &\leq \E_s \left[ \|\sfx - \bthetastar_s \|^2 \right] + \| \bthetastar_s \|^2 \\
            &\leq \sigma^2 \cdot d + \deltamax^2 \\
            &= 1 + \deltamax^2.
    \end{align*}
    All in all, 
    \begin{align}
        \left| \epsilon_2 \right|
            &\leq \left( |\cT| \cdot (\kfit - 1) + (\ktrue - |\cT|) \right) \cdot
                \left( 1 + \frac{4}{\sqrt{2\pi} \cdot \delta } \right)^{1/2} e^{- \frac{\delta^2}{64}} \nonumber\\
            &\leq \kfit \ktrue \cdot 
                \left( 1 + \frac{4}{\sqrt{2\pi} \cdot \delta } \right)^{1/2} e^{- \frac{\delta^2}{64}},   \label{eqn:accU_eps2}\\
        \left\| \epsilon_1 \right\|
            &\leq \left( |\cT| \cdot (\kfit - 1) + (\ktrue - |\cT|) \right) \cdot \left( 1 + \deltamax^2 \right)^{1/2} \cdot \left( 1 + \frac{4}{\sqrt{2\pi} \cdot \delta } \right)^{1/2} e^{- \frac{\delta^2}{64}}  \nonumber\\
            &\leq \sqrt{2} \kfit \ktrue \deltamax \cdot
                \left( 1 + \frac{4}{\sqrt{2\pi} \cdot \delta } \right)^{1/2} e^{- \frac{\delta^2}{64}}.   \label{eqn:accU_eps1}
    \end{align}

    Observe that $|\epsilon_2| \leq 1/2$ because $\delta \geq \max \left\{ \frac{4}{\sqrt{2\pi}}, \, 8 \sqrt{\log ( 2 \sqrt{2} \cdot \kfit \ktrue)} \right\}$. 
    Combining the upper bounds \eqref{eqn:accU_eps2} and \eqref{eqn:accU_eps1} with \eqref{eqn:accu_error}, we obtain
    \begin{align*}
        \left\| \bbeta_i - \frac{1}{|\cT|} \sum_{s \in \cT} \bthetastar_s \right\|
            &= \frac{ \| \epsilon_1 \|}{ \big| |\cT| + \epsilon_2 \big|}\\
            &\leq \frac{\| \epsilon_1 \|}{|\cT| - |\epsilon_2|} \leq 2 \| \epsilon_1 \|\\
            &\leq 4 \kfit \ktrue \deltamax \cdot e^{- \frac{\delta^2}{64}}. 
    \end{align*}
\end{proof}

\end{document}